\pdfoutput=1
\documentclass[10pt,twocolumn,letterpaper]{article}

\usepackage{iccv}              

\definecolor{iccvblue}{rgb}{0.21,0.49,0.74}
\usepackage[pagebackref,breaklinks,colorlinks,allcolors=iccvblue]{hyperref}
\newcommand{\boldparagraph}[1]{\vspace{0.2cm}\noindent{\bf #1.}}
\usepackage{multirow}
\usepackage{tikz}
\usetikzlibrary{shapes}
\usepackage{graphicx}  
\usepackage{array}

\definecolor{myPurple}{RGB}{29,64,76}
\definecolor{myBlue}{RGB}{45,128,120}
\definecolor{myRed}{RGB}{204,79,66}
\definecolor{myYellow}{RGB}{222,158,70}
\definecolor{markGreen}{RGB}{67,160,71}
\usepackage{pifont}
\newcommand{\cmark}{\ding{51}}  
\newcommand{\xmark}{\ding{55}}  
\newcommand{\circlecheck}{%
  \tikz\draw[markGreen, fill=markGreen] (0,0) circle (0.45em)
    node[white]{\cmark};%
}

\newcommand{\circlecross}{%
  \tikz\draw[red, fill=red] (0,0) circle (0.45em)
    node[white]{ \xmark};%
}

\def\paperID{1411} 
\def\confName{ICCV}
\def\confYear{2025}

\title{LoftUp: Learning a Coordinate-Based Feature Upsampler for \\Vision Foundation Models}

\author{Haiwen Huang\textsuperscript{1,2}
\qquad
Anpei Chen\textsuperscript{1,2} \qquad Volodymyr Havrylov\textsuperscript{1} \\  Andreas Geiger\textsuperscript{1,2} \qquad Dan Zhang\textsuperscript{3} \\
\textsuperscript{1}University of Tübingen \qquad \textsuperscript{2} Tübingen AI Center \qquad \textsuperscript{3} Bosch Center for Artificial Intelligence
}

\begin{document}
\maketitle
\begin{abstract}
Vision foundation models (VFMs) such as DINOv2 and CLIP have achieved impressive results on various downstream tasks, but their limited feature resolution hampers performance in applications requiring pixel-level understanding. Feature upsampling offers a promising direction to address this challenge. In this work, we identify two critical factors for enhancing feature upsampling:  the upsampler architecture and the training objective. For the upsampler architecture, we introduce a coordinate-based cross-attention transformer that integrates the high-resolution images with coordinates and low-resolution VFM features to generate sharp, high-quality features. For the training objective, we propose constructing high-resolution pseudo-groundtruth features by leveraging class-agnostic masks and self-distillation. Our approach effectively captures fine-grained details and adapts flexibly to various input and feature resolutions. Through experiments, we demonstrate that our approach significantly outperforms existing feature upsampling techniques across various downstream tasks. Our code is released at \url{https://github.com/andrehuang/loftup}.
\end{abstract}
\vspace{-0.2cm}    
\section{Introduction}
\label{sec:intro}

High-quality pretrained representations from Vision Foundation Models (VFMs) have become standard for a wide range of computer vision tasks~\cite{el2024probing, radford2021learning, yu2023convolutions, huang2024renovating, oquab2024dinov, tumanyan2024dinotrack, am-radio, kirillov2023segment, ravi2024sam2, zhai2023sigmoid, tschannen2025siglip}. However, because of the patrification or aggressive pooling operations in VFMs, the output features are typically $16$ or even more times smaller in spatial resolution than the input images, limiting their utility for tasks that require fine-grained, pixel-level understanding.

To address the challenge of limited feature resolution in VFMs, one straightforward approach is to use larger image inputs to obtain higher-resolution features. However, processing high-resolution inputs incurs a quadratic increase in computational cost and can introduce severe artifacts if the VFMs are not trained for such resolutions. Moreover, training or fine-tuning VFMs on high-resolution images demands substantial computational resources and meticulous tuning~\cite{oquab2024dinov, am-radio, ravi2024sam2, tschannen2025siglip2}.
An alternative strategy is to train task-specific decoders that leverage multi-layer intermediate features to upsample VFM outputs~\cite{liu2023simpleclick, cheng2021mask2former, yu2023convolutions, huang2024renovating, li2022exploring}. Yet, this approach often comes with significant training costs and necessitates retraining the decoder for each new application. In addition, many downstream tasks lack sufficient data needed to fine-tune a high-quality decoder.

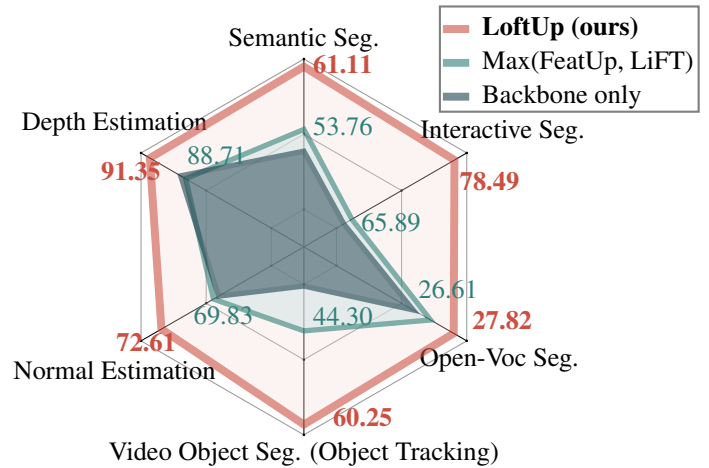
\begin{figure}[t]
\centering


\newcommand{\D}{6}

\newdimen\R
\R=5cm

\newcommand{\A}{360/\D}

\newcommand{\dimMax}[1]{%
  \ifcase#1
    0 
  \or 62 
  \or 92  
  \or 73.7  
  \or 62  
  \or 28.5  
  \or 80 
  \fi
}

\newcommand{\dimMin}[1]{%
  \ifcase#1
    0 
  \or 40   
  \or  80   
  \or 65   
  \or 30   
  \or 20   
  \or  60   
  \fi
}

\newcommand{\coord}[2]{%
  (\A * #1 : {((#2)-\dimMin{#1}) / (\dimMax{#1}-\dimMin{#1}) * \R})%
}


\begin{tikzpicture}[scale=0.5, rotate=30]

  \coordinate (O) at (0,0);

  \foreach \i in {1,...,\D} {
    \draw (\i*\A:0) -- (\i*\A:\R);
  }

\foreach \f in {0.2,0.6, 1.0} {
    \draw [opacity=0.4]
      \coord{1}{\dimMin{1} + \f*(\dimMax{1}-\dimMin{1})}
      \foreach \i in {2,...,\D} {
         -- \coord{\i}{\dimMin{\i} + \f*(\dimMax{\i}-\dimMin{\i})}
      } -- cycle;

    \foreach \i in {1,...,\D} {
       \fill \coord{\i}{\dimMin{\i} + \f*(\dimMax{\i}-\dimMin{\i})} circle (1pt);
    }
  }

  \path (1*\A:1.10\R) node {\normalsize Semantic Seg.};
  \path (2*\A-3:1.2\R) node {\normalsize Depth Estimation};
  \path (3*\A+3:1.2\R) node {\normalsize Normal Estimation};
  \path (4*\A:1.10\R) node {\normalsize Video Object Seg. (Object Tracking)};
  \path (5*\A:1.20\R) node {\normalsize Open-Voc Seg.};
  \path (6*\A:1.22\R) node {\normalsize Interactive Seg.};


  \draw [color=myRed, line width=3pt, opacity=0.6, fill=myRed!20, fill opacity=0.3]
    \coord{1}{61.11} --
    \coord{2}{91.35} --
    \coord{3}{72.61} --
    \coord{4}{60.25} --
    \coord{5}{27.82} --
    \coord{6}{78.49} -- cycle;

  \draw [color=myBlue, line width=2pt, fill=myBlue!20, opacity=0.6, fill opacity=0.6]
    \coord{1}{53.76} --
    \coord{2}{88.71} --
    \coord{3}{69.83} --
    \coord{4}{44.30} --
    \coord{5}{26.61} --
    \coord{6}{65.89} -- cycle;


  \draw [color=myPurple, line width=2pt, fill=myPurple, opacity=0.6, fill opacity=0.5]
    \coord{1}{51.21} --
    \coord{2}{89.08} --
    \coord{3}{69.56} --
    \coord{4}{36.66} --
    \coord{5}{25.70} --
    \coord{6}{64.90} -- cycle;

\node[font=\normalsize\bfseries, color=myRed, anchor=south west] at \coord{1}{59.2} {61.11};
\node[font=\normalsize\bfseries, color=myRed, anchor=south west] at \coord{2.25}{94} {91.35};
\node[font=\normalsize\bfseries, color=myRed, anchor=south west] at \coord{3}{75.51} {72.61};
\node[font=\normalsize\bfseries, color=myRed, anchor=south west] at \coord{4.1}{62.25} {60.25};
\node[font=\normalsize\bfseries, color=myRed, anchor=south west] at \coord{5}{28.32} {27.82};
\node[font=\normalsize\bfseries, color=myRed, anchor=north west] at \coord{6}{78} {78.49};

\node[font=\normalsize, color=myBlue, anchor=south west] at \coord{1}{52} {53.76};
\node[font=\normalsize, color=myBlue, anchor=south west] at \coord{2}{89.3} {88.71};
\node[font=\normalsize, color=myBlue, anchor=south west] at \coord{3.1}{71.83} {69.83};
\node[font=\normalsize, color=myBlue, anchor=south west] at \coord{4}{45.00} {44.30};
\node[font=\normalsize, color=myBlue, anchor=south west] at \coord{5}{25.5} {26.61};
\node[font=\normalsize, color=myBlue, anchor=north west] at \coord{6.2}{66.89} {65.89};


  \begin{scope}[xshift=6cm, yshift=3.0cm, rotate=-30]
  \draw[draw=gray, fill=none, thick] (-0.08,0.8) rectangle (6.8cm,-2cm);
    \draw [color=myRed,   line width=3pt, opacity=0.6] (0,0.2) -- (0.7,0.2);
    \node [anchor=west] at (0.8,0.2)   {\normalsize\bfseries LoftUp (ours)};

    \draw [color=myBlue,   line width=3pt, opacity=0.6] (0,-0.7) -- (0.7,-0.7);
    \node [anchor=west] at (0.8,-0.7)   {\normalsize Max(FeatUp, LiFT)};

    \draw [color=myPurple, line width=3pt, opacity=0.6] (0,-1.6) -- (0.7,-1.6);
    \node [anchor=west] at (0.8,-1.6) {\normalsize Backbone only};
  \end{scope}

\end{tikzpicture}

\caption{\textbf{LoftUp improves significantly across various tasks over the VFM backbone (DINOv2-S~\cite{oquab2024dinov}) and current SoTA feature upsampling performance (FeatUp~\cite{fu2024featup} and LiFT~\cite{suri2024lift}).} See experiment details in~\cref{sec:exp}. }
\label{fig:spider}
\vspace{-0.4cm}
\end{figure}

More recently, FeatUp~\cite{fu2024featup} and LiFT~\cite{suri2024lift} have independently introduced task-agnostic feature upsamplers trained with general reconstruction losses, demonstrating that such methods can substantially enhance VFM performance across a variety of tasks. By upsampling VFM features and pairing them with a lightweight task-specific decoder, these approaches provide a promising alternative that avoids heavy, task-specific training while achieving a more generalizable solution. Nonetheless, as shown in~\cref{fig:spider}, current feature upsamplers still fall short of reaching optimal performance.

\begin{figure*}[t]
    \centering
    \begin{subfigure}[b]{0.12\textwidth}
         \centering
         \includegraphics[width=\textwidth]{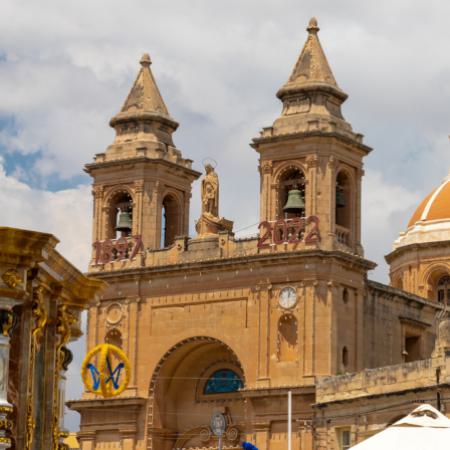}
         \label{fig:image}
     \end{subfigure}
     \hfill
     \begin{subfigure}[b]{0.12\textwidth}
         \centering
         \includegraphics[width=\textwidth]{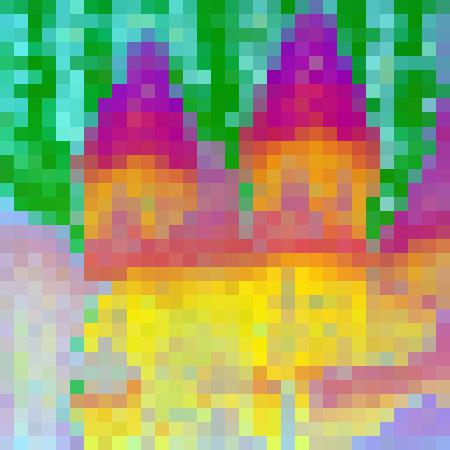}
         \label{fig:image}
     \end{subfigure}
     \hfill
     \begin{subfigure}[b]{0.12\textwidth}
         \centering
         \includegraphics[width=\textwidth]{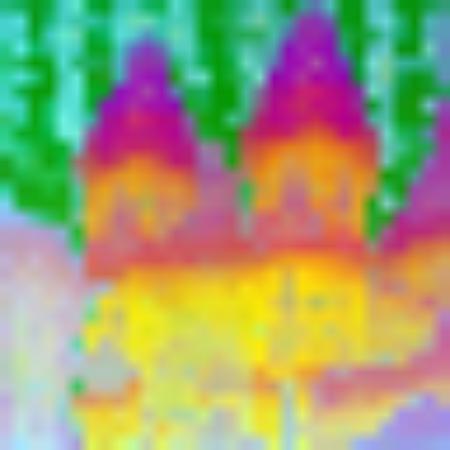}
         \label{fig:bilinear-openvoc}
     \end{subfigure}
     \hfill
     \begin{subfigure}[b]{0.12\textwidth}
         \centering
         \includegraphics[width=\textwidth]{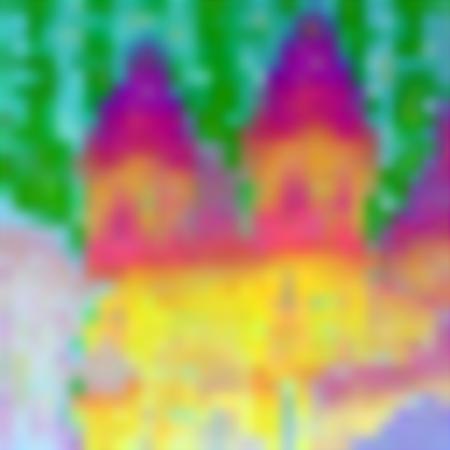}
         \label{fig:three sin x}
     \end{subfigure}
     \hfill
     \begin{subfigure}[b]{0.12\textwidth}
         \centering
         \includegraphics[width=\textwidth]{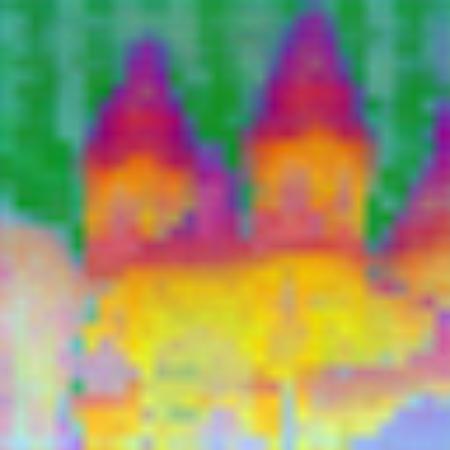}
         \label{fig:five over x}
     \end{subfigure}
     \hfill
     \begin{subfigure}[b]{0.12\textwidth}
         \centering
         \includegraphics[width=\textwidth]{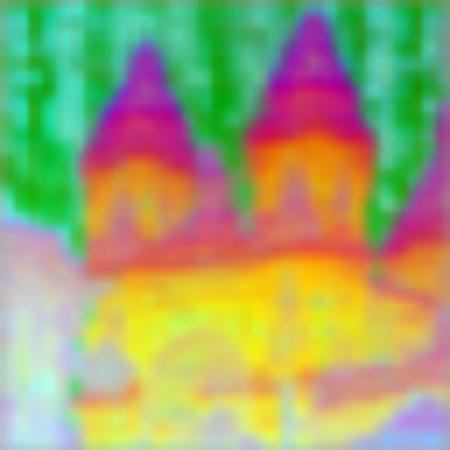}
         \label{fig:five over x}
     \end{subfigure}
     \hfill
     \begin{subfigure}[b]{0.12\textwidth}
         \centering
         \includegraphics[width=\textwidth]{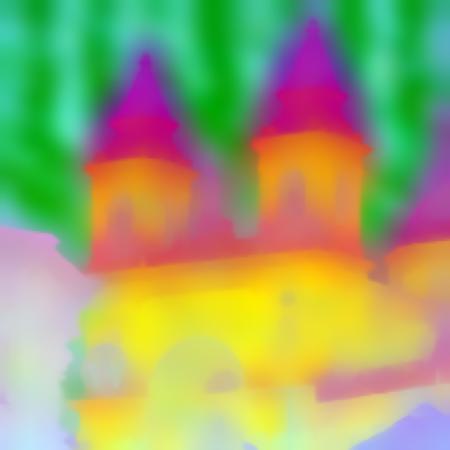}
         \label{fig:five over x}
     \end{subfigure}
     \hfill
     \begin{subfigure}[b]{0.12\textwidth}
         \centering
         \includegraphics[width=\textwidth]{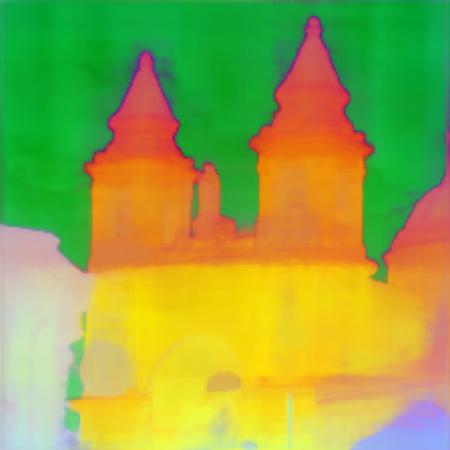}
         \label{fig:five over x}
     \end{subfigure} \\
     \vspace{-0.2cm}
      \begin{subfigure}[b]{0.12\textwidth}
         \centering
         \includegraphics[width=\textwidth]{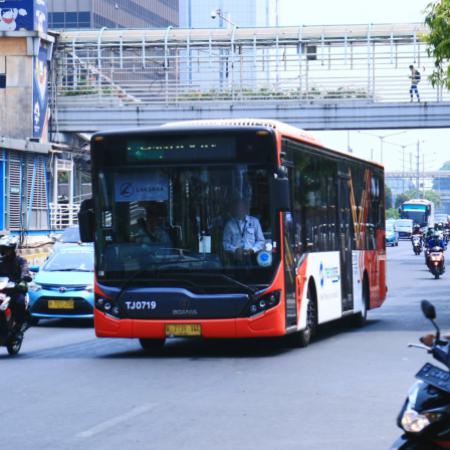}
         \label{fig:image}
     \end{subfigure}
     \hfill
     \begin{subfigure}[b]{0.12\textwidth}
         \centering
         \includegraphics[width=\textwidth]{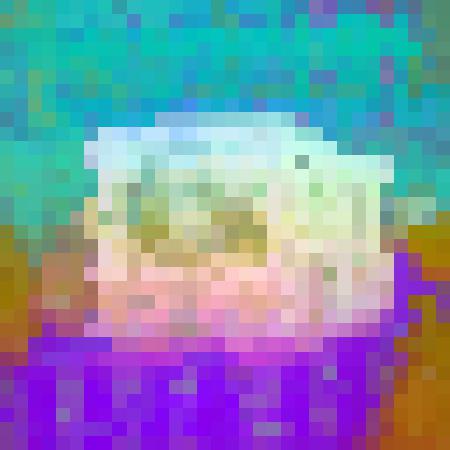}
         \label{fig:image}
     \end{subfigure}
     \hfill
     \begin{subfigure}[b]{0.12\textwidth}
         \centering
         \includegraphics[width=\textwidth]{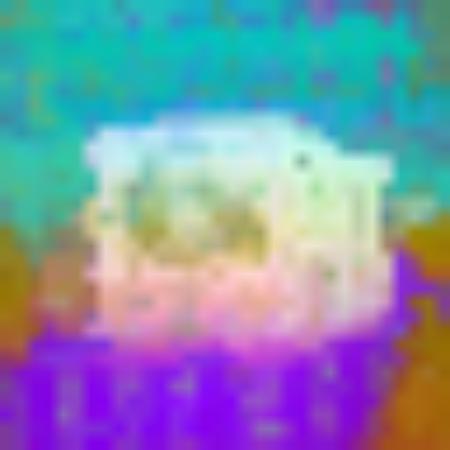}
         \label{fig:bilinear-openvoc}
     \end{subfigure}
     \hfill
     \begin{subfigure}[b]{0.12\textwidth}
         \centering
         \includegraphics[width=\textwidth]{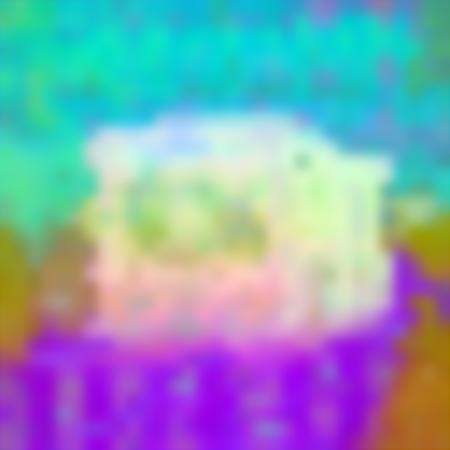}
         \label{fig:three sin x}
     \end{subfigure}
     \hfill
     \begin{subfigure}[b]{0.12\textwidth}
         \centering
         \includegraphics[width=\textwidth]{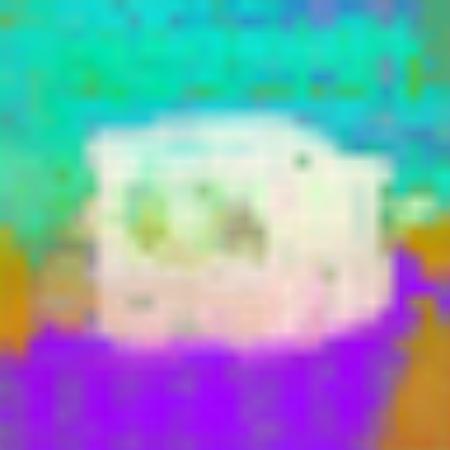}
         \label{fig:five over x}
     \end{subfigure}
     \hfill
     \begin{subfigure}[b]{0.12\textwidth}
         \centering
         \includegraphics[width=\textwidth]{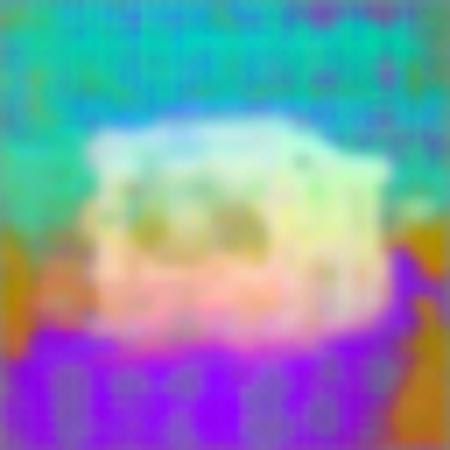}
         \label{fig:five over x}
     \end{subfigure}
     \hfill
     \begin{subfigure}[b]{0.12\textwidth}
         \centering
         \includegraphics[width=\textwidth]{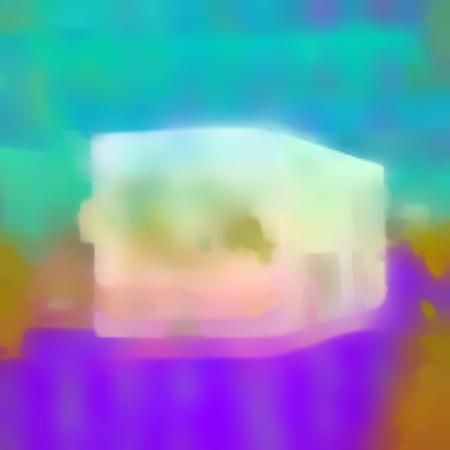}
         \label{fig:five over x}
     \end{subfigure}
     \hfill
     \begin{subfigure}[b]{0.12\textwidth}
         \centering
         \includegraphics[width=\textwidth]{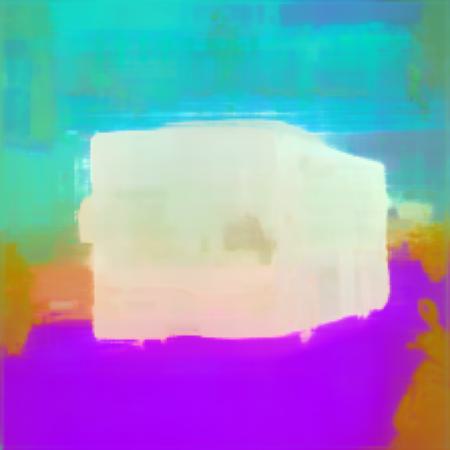}
         \label{fig:five over x}
     \end{subfigure} \\
     \vspace{-0.2cm}
         \begin{subfigure}[b]{0.12\textwidth}
         \centering
         \includegraphics[width=\textwidth]{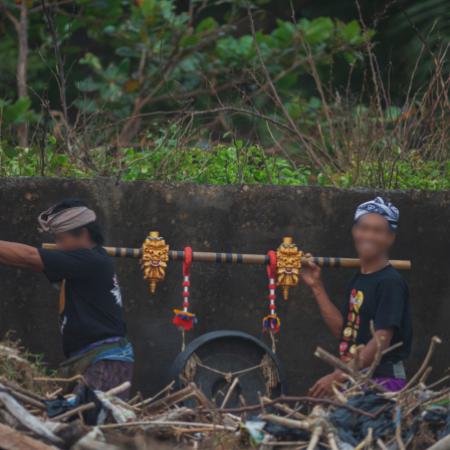}
         \caption{Original image}
         \label{fig:image}
     \end{subfigure}
     \hfill
     \begin{subfigure}[b]{0.12\textwidth}
         \centering
         \includegraphics[width=\textwidth]{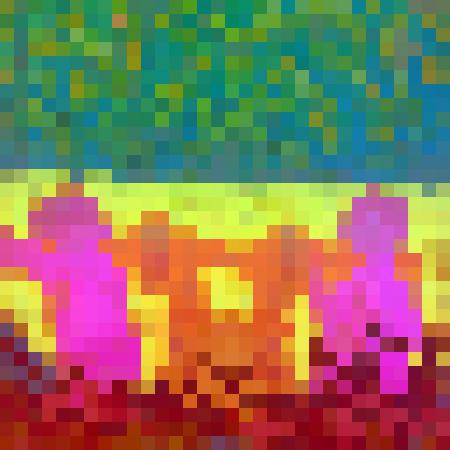}
         \caption{Low res}
         \label{fig:image}
     \end{subfigure}
     \hfill
     \begin{subfigure}[b]{0.12\textwidth}
         \centering
         \includegraphics[width=\textwidth]{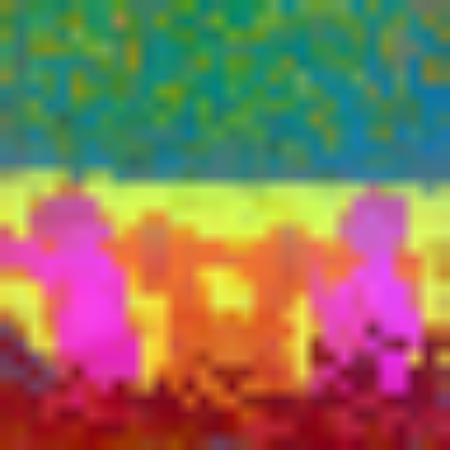}
         \caption{Bilinear}
         \label{fig:bilinear-openvoc}
     \end{subfigure}
     \hfill
     \begin{subfigure}[b]{0.12\textwidth}
         \centering
         \includegraphics[width=\textwidth]{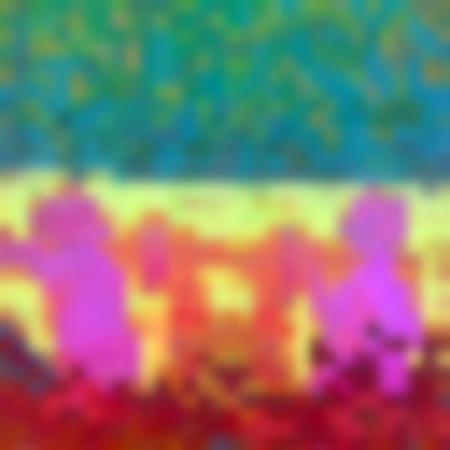}
         \caption{Resize-conv}
         \label{fig:three sin x}
     \end{subfigure}
     \hfill
     \begin{subfigure}[b]{0.12\textwidth}
         \centering
         \includegraphics[width=\textwidth]{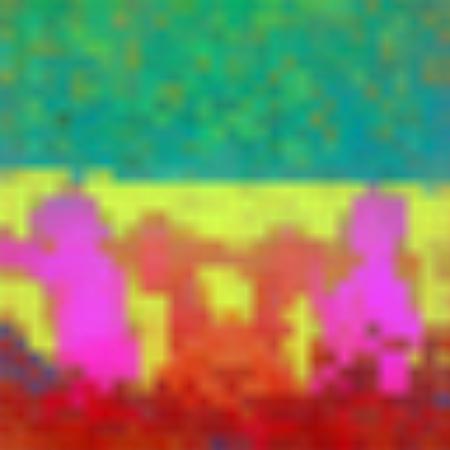}
         \caption{LIIF~\cite{chen2021liif}}
         \label{fig:five over x}
     \end{subfigure}
     \hfill
     \begin{subfigure}[b]{0.12\textwidth}
         \centering
         \includegraphics[width=\textwidth]{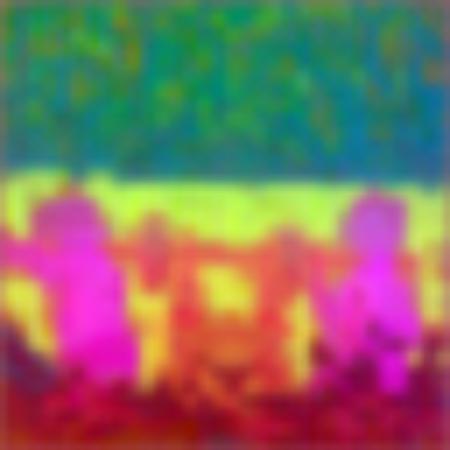}
         \caption{LiFT~\cite{suri2024lift}}
         \label{fig:five over x}
     \end{subfigure}
     \hfill
     \begin{subfigure}[b]{0.12\textwidth}
         \centering
         \includegraphics[width=\textwidth]{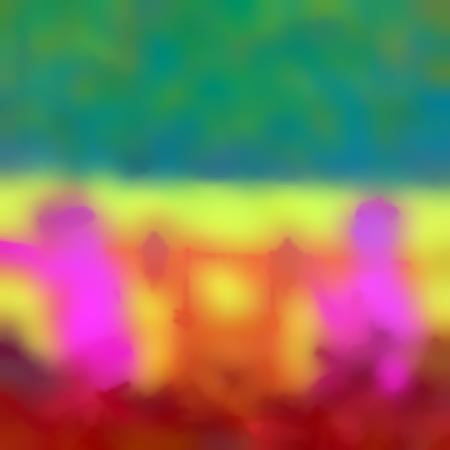}
         \caption{FeatUp~\cite{fu2024featup}}
         \label{fig:five over x}
     \end{subfigure}
     \hfill
     \begin{subfigure}[b]{0.12\textwidth}
         \centering
         \includegraphics[width=\textwidth]{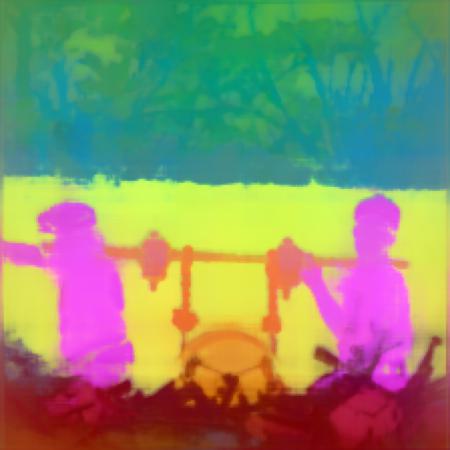}
         \caption{\textbf{LoftUp} (Ours)}
         \label{fig:five over x}
     \end{subfigure} \\
    \caption{\textbf{Comparison of features from upsamplers.} Backbone is DINOv2-S/14~\cite{oquab2024dinov}.}
    \vspace{-0.3cm}
    \label{fig:vis-feats}
\end{figure*}

In this work, we systematically explore the design space of feature upsamplers and identify two critical components: the upsampler architecture and the training objective. The architecture determines the capacity of the upsampler to learn effectively, while the training objective defines the upper performance limit. By optimizing both elements, our approach achieves substantially stronger results than previous state-of-the-art methods.

Regarding the \textbf{upsampler architecture}, to capture high-resolution details while avoiding the cumulative artifacts from multiple layers of interpolation or deconvolution, we propose a simple coordinate-based transformer that directly predicts high-resolution features for each pixel. Specifically, our model takes image coordinates and RGB values as inputs and performs cross-attention with the low-resolution VFM feature map. This  facilitates a fine-grained, content-aware mapping from coordinates to high-resolution features, effectively bypassing the constraints imposed by fixed local kernels and standard upsampling layers. Moreover, unlike the implicit approach in FeatUp~\cite{fu2024featup}, which requires test-time optimization, our method learns feature upsampling directly from the training dataset and generalizes  to diverse scenes without additional test-time adjustments.

For the \textbf{training objective}, the primary challenge is the absence of groundtruth high-resolution feature annotations. Although downstream task labels such as depth or masks can be used, this approach risks compromising the task-agnostic nature of the upsampler and hinders its ability to generalize to unseen tasks. 
Due to this challenge, previous task-agnostic feature upsampling works, FeatUp~\cite{fu2024featup} and LiFT~\cite{suri2024lift}, both compute the training loss at a low resolution. 
In this work, we address these limitations by constructing pseudo-groundtruth (pseudo-GT) features directly at the input image resolution. Specifically, we leverage class-agnostic masks generated by off-the-shelf segmentation foundation models~\cite{kirillov2023segment, ravi2024sam2} to ensure that the pseudo-GT accurately reflects the underlying geometry and delineates object boundaries. We further refine the pseudo-GT using a self-distillation strategy that reduces noise and artifacts. This high-resolution pseudo-GT enables loss computation at a high resolution, empowering the upsampler to learn fine-grained details.

In~\cref{fig:vis-feats}, we qualitatively show that LoftUp yields markedly sharper and more detailed features than alternative upsampling methods.
To further demonstrate the versatility and effectiveness of our approach, we evaluate it on a range of downstream tasks such as semantic segmentation, depth estimation, and video objectschannen2025siglip2. As shown in~\cref{fig:spider}, our approach leads to performance gains of 10–20\% over previous SoTA upsamplers for most tasks, and an impressive nearly 50\% improvement on video object segmentation~\cite{Pont-Tuset_arXiv_2017_davis}.
Furthermore, thanks to its coordinate-based design, our upsampler adapts seamlessly to various input and feature resolutions, catering to the diverse requirements of downstream applications. Overall, with less than a 20\% increase in parameters compared to the original foundation models, our feature upsampler offers a task-agnostic, lightweight, and plug-and-play enhancement that significantly boosts VFM backbones across multiple tasks.

\section{Related Work}
\label{sec:related_work}
Feature upsampling refers to increasing the spatial resolution of a feature map. In this work, our goal is to increase the feature resolution to the original image resolution, that is, full resolution.
Traditional non-learnable methods include various ways of interpolation~\cite{mckinley1998cubic, duchon1979lanczos} and image-adaptive filtering such as joint bilateral filtering (JBU)~\cite{kopf2007joint} and guided image filtering~\cite{he2012guided}.
In modern deep learning, previous work has proposed various architecture- and downstream-task-specific feature upsamplers. For example,  Index Networks~\cite{lu2020index} and A2U~\cite{dai2021learning} are effective on image matting, but fall short in other tasks. PointRend~\cite{kirillov2020pointrend} proposes a point-rendering method specifically for upsampling segmentation output. 
And CARAFE~\cite{wang2019carafe}, SAPA~\cite{lu2022sapa}, and FADE~\cite{lu2022fade} are proposed specifically for encoder-decoder architectures.
More recently, with the success of vision foundation models such as DINOv2~\cite{oquab2024dinov} and CLIP~\cite{radford2021learning}, there is a trend for feature upsamplers to be downstream-task-agnostic so that they can be used with the VFM backbone together in various applications~\cite{fu2024featup, suri2024lift}. Our work falls into this category. With this task-agnostic goal in mind, we further explain the architecture and training objective design as follows.

\boldparagraph{Architecture for Feature Upsamplers}
Traditional upsampler architectures rely on multiple layers of interpolation or deconvolution to transform low-resolution features into higher resolutions. Examples include JBU~\cite{kopf2007joint}, standard deconvolution~\cite{Noh_2015_deconv, shi2016deconvolution, dumoulin2016guide}, resize-convolution~\cite{odena2016deconvolution}  and U-Net-style upsampling modules~\cite{suri2024lift, ronneberger2015unet}.  However, the multi-layer design inevitably leads to error accumulation, resulting in increased blurriness as the resolution increases.

In this work, inspired by coordinate-based methods in 3D reconstruction~\cite{yu2020pixelnerf, Lin_2023_vit_nerf, Ye_2023_featurenerf, Szymanowicz_2024_splatterimage}, we adopt a coordinate-based approach and view feature upsampling as a mapping from high-resolution coordinates to high-resolution features.
This effectively bypasses the limitations of  standard upsampling layers.
Previously, FeatUp also proposed a coordinate-based network (MLP) for feature upsampling~\cite{fu2024featup}. However, their approach requires per-image optimization and is therefore not scalable. Another related work, LIIF~\cite{chen2021liif}, also employs an MLP to parameterize high-resolution outputs but is limited to local feature interactions.
In contrast, our method does not need test-time optimization and enables global interactions between image inputs and low-resolution features through a cross-attention mechanism, leading to stronger upsampling performance. 

\boldparagraph{Task-agnostic Training Objective for Feature Upsampling}
Due to the absence of groundtruth high-resolution features, it is a challenge to create high-quality training objective for task-agnostic feature upsampling.
FeatUp~\cite{fu2024featup} and LiFT~\cite{suri2024lift} were the first two works to propose a task-agnostic training pipeline for feature upsamplers. However, their training is at a low resolution, leaving the high-resolution features severely under-constrained.
We will explain them in more detail~\cref{sec:method}.
In this paper, we propose a self-distillation approach to generate full-resolution pseudo-GT, which is then used to supervise feature upsampling training at full resolution. This approach fully unlocks the potential of the feature upsampler to capture fine-grained details in high-resolution images. 




\begin{figure}
    \centering
    \includegraphics[width=\linewidth]{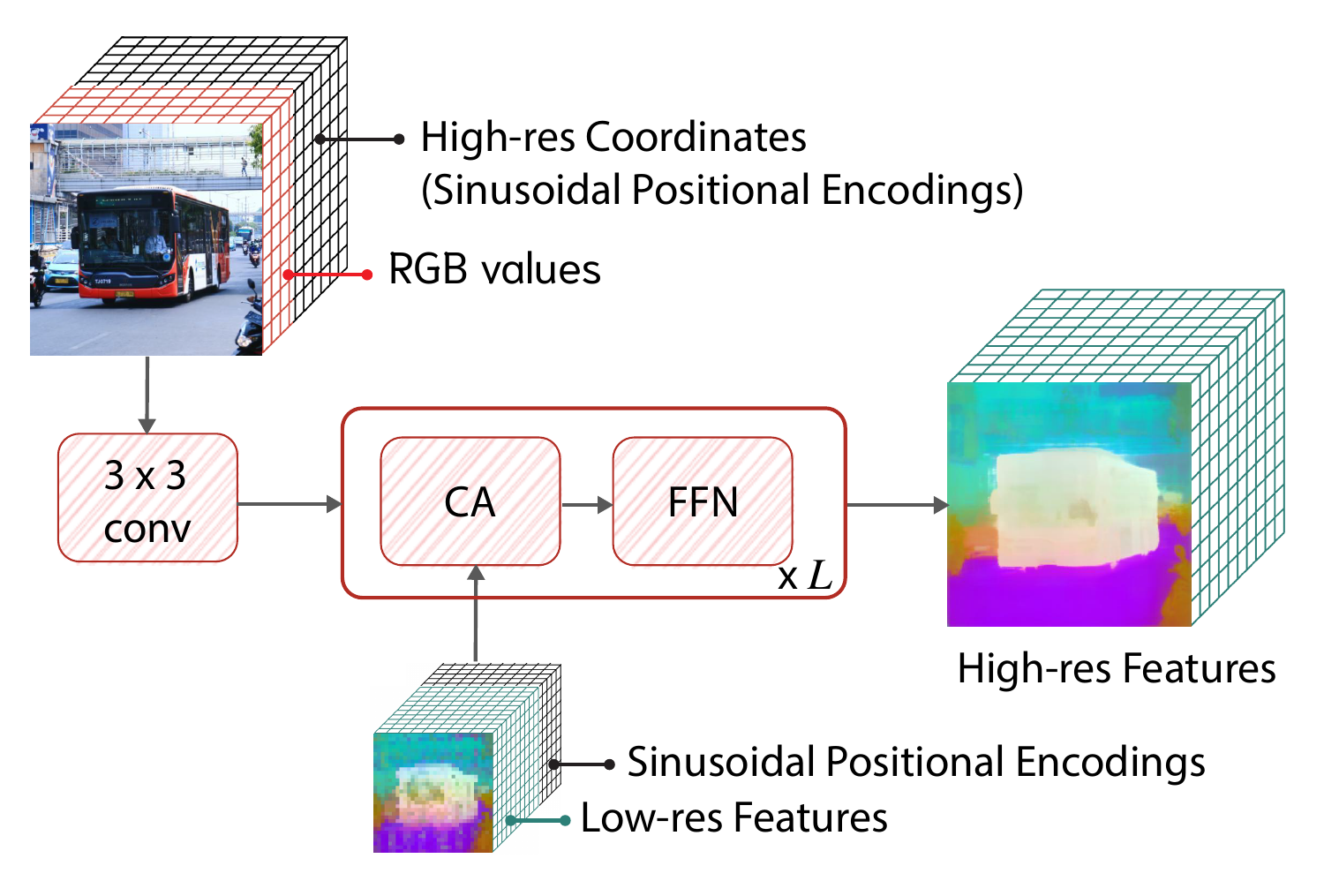}
    \caption{\textbf{Architecture of LoftUp.} Our coordinate-based network with cross-attention mechanism effectively integrates the fine-grained details from image RGB values and semantically-rich low-res features to produce high-resolution feature maps.}
    \label{fig:fit-arch}
    \vspace{-0.2cm}
\end{figure}

\begin{figure*}[t]
     \centering
         \includegraphics[width=\textwidth]{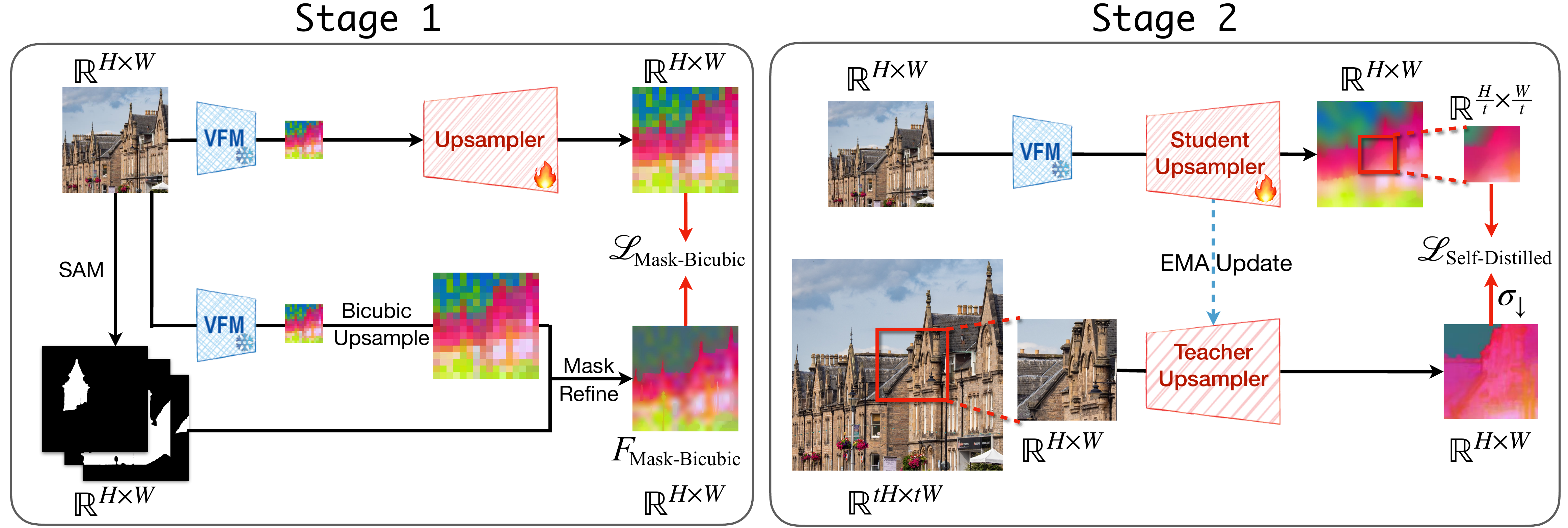}
         \vspace{-0.1cm}
    \caption{\textbf{Our two-stage LoftUp training approach.} Stage 1 trains an upsampler with class-agnostic masks to refine bicubic-upsampled features. Stage 2 employs self-distillation, initializing teacher and student upsamplers from Stage 1's pre-trained model. All VFM image inputs share the resolution ($H\times W$). For visual clarity, the VFM block is omitted from Stage 2’s teacher branch.}
    \label{fig:pipeline}
    \vspace{-0.2cm}
\end{figure*}


\section{Coordinate-Based Feature Representation}
Prior works typically address feature upsampling as a gradual resolution enhancement process, where features are progressively lifted to a higher resolution. 
In contrast, we adopt a coordinate-based representation of the high-resolution features and view feature upsampling as a mapping from a pixel coordinate $(x,y)$ to the high-resolution features at that pixel.
Specifically, we propose a cross-attention mechanism to effectively incorporate the low-resolution features and the high-resolution image inputs with coordinates.

As shown in~\cref{fig:fit-arch}, following prior work on 3D coordinate-based reconstruction~\cite{yu2020pixelnerf, mildenhall2021nerf}, our model encodes full-resolution coordinates with sinusoidal positional embeddings and concatenates them with RGB values. A convolutional layer then projects this combined input into the feature dimension. Next, an $L$-block cross-attention transformer uses these high-resolution features as queries and low-resolution VFM features as keys and values, producing the final high-resolution feature map.

Our cross-attention design enables high-frequency details to interact globally with semantically rich representations, facilitating global-content-aware upsampling. This overcomes the limitations of fixed or locally predicted kernels in prior feature upsamplers. As illsustrated in~\cref{fig:vis-feats}, multi-layer upsamplers such as resize-conv~\cite{odena2016deconvolution}, LiFT (U-Net)~\cite{suri2024lift}, and FeatUp (a modified JBU)~\cite{fu2024featup} tend to produce blurry outputs with artifacts. Additionally, while LIIF~\cite{chen2021liif} employs a coordinate-based design, its reliance on local interactions limits its upsampling quality. In contrast, our LoftUp accurately captures object boundaries and produces fine-grained feature maps with minimal artifacts.

Furthermore, our coordinate-based approach allows flexible generation of feature maps at arbitrary resolutions by adjusting the input coordinate resolution—unlike traditional upsampling methods, which are restricted to fixed scaling factors due to their multi-layer structures.
Finally, we note that while attention mechanisms can be computationally expensive, our cross-attention remains relatively efficient because it processes a much smaller set of low-resolution tokens as keys and values, rather than the progressively higher-resolution feature maps as in multi-layer upsampler architectures. In fact, as our experiments (\cref{tab:efficiency}) later show, our upsampler's inference speed is comparable to bilinear upsampling.

\begin{figure*}[t]
     \centering
     \begin{subfigure}[b]{0.19\textwidth}
         \centering
         \includegraphics[width=\textwidth]{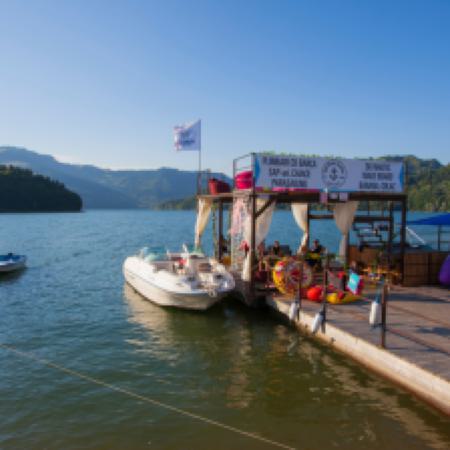}
         \label{fig:original}
     \end{subfigure}
     \hfill
     \begin{subfigure}[b]{0.19\textwidth}
         \centering
         \includegraphics[width=\textwidth]{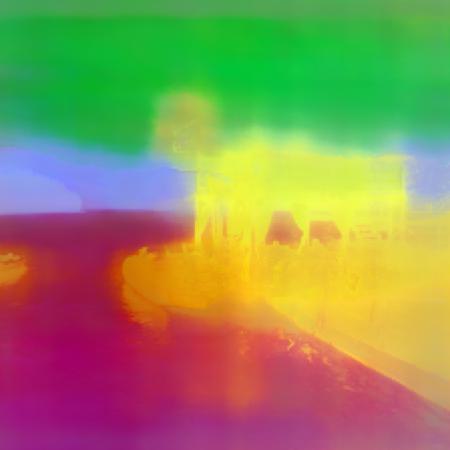}
         \label{fig:pseudo-gt-1}
     \end{subfigure}
     \hfill
     \begin{subfigure}[b]{0.19\textwidth}
         \centering
         \includegraphics[width=\textwidth]{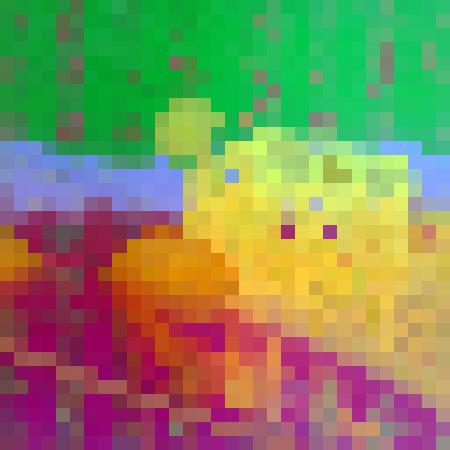}
         \label{fig:three sin x}
     \end{subfigure}
     \hfill
     \begin{subfigure}[b]{0.19\textwidth}
         \centering
         \includegraphics[width=\textwidth]{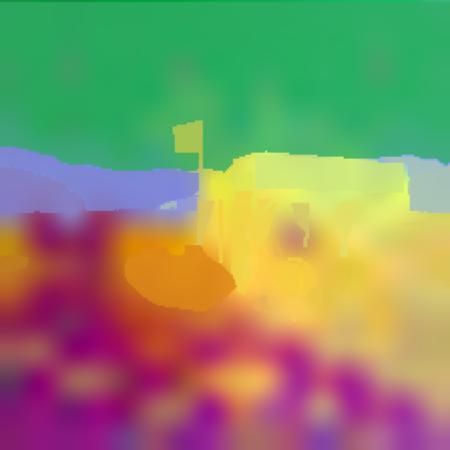}
         \label{fig:five over x}
     \end{subfigure}
     \begin{subfigure}[b]{0.19\textwidth}
         \centering
         \includegraphics[width=\textwidth]{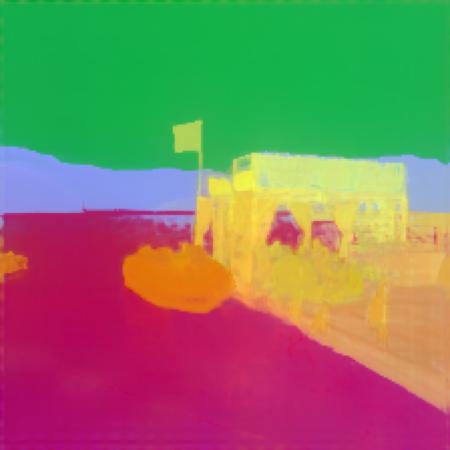}
         \label{fig:five over x}
     \end{subfigure}
     \\
     \vspace{-0.3cm}
     \begin{subfigure}[b]{0.19\textwidth}
         \centering
         \includegraphics[width=\textwidth]{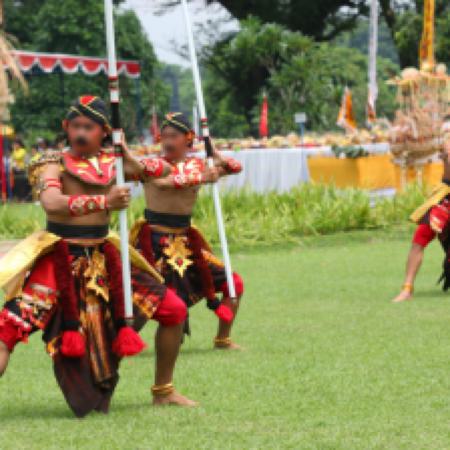}
         \caption{Original image}
         \label{fig:original}
     \end{subfigure}
     \hfill
     \begin{subfigure}[b]{0.19\textwidth}
         \centering
         \includegraphics[width=\textwidth]{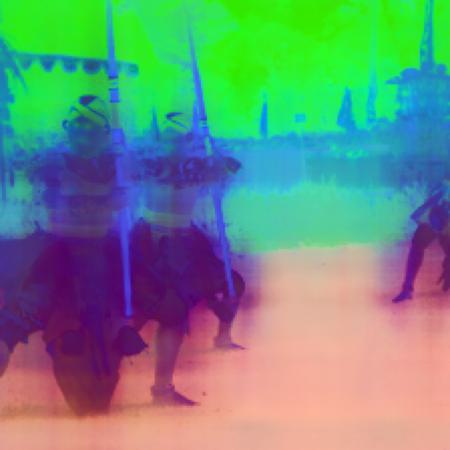}
         \caption{Per-image optimized}
         \label{fig:pseudo-gt-1}
     \end{subfigure}
     \hfill
     \begin{subfigure}[b]{0.19\textwidth}
         \centering
         \includegraphics[width=\textwidth]{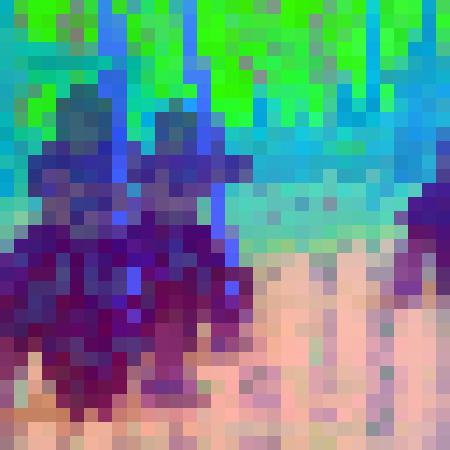}
         \caption{2x features (LiFT)}
         \label{fig:three sin x}
     \end{subfigure}
     \hfill
     \begin{subfigure}[b]{0.19\textwidth}
         \centering
         \includegraphics[width=\textwidth]{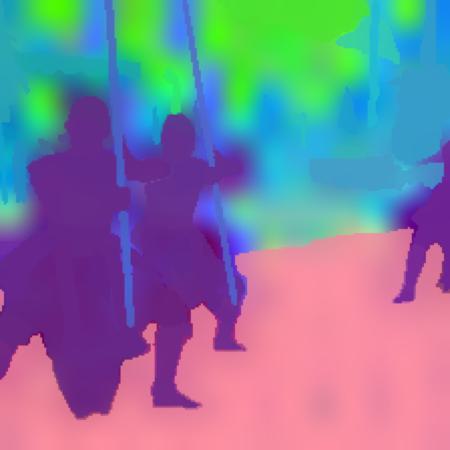}
         \caption{Mask-Bicubic (Stage 1)}
         \label{fig:five over x}
     \end{subfigure}
     \begin{subfigure}[b]{0.19\textwidth}
         \centering
         \includegraphics[width=\textwidth]{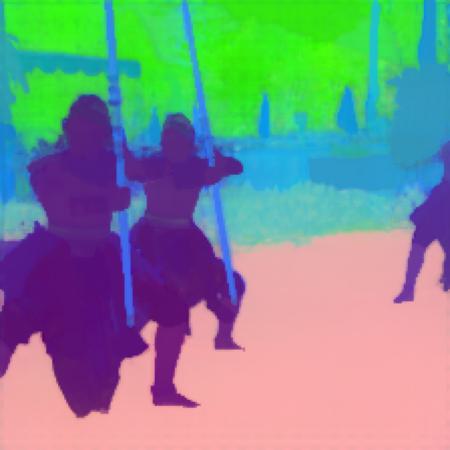}
         \caption{Self-Distilled (Stage 2)}
         \label{fig:five over x}
     \end{subfigure} \\

\caption{\textbf{Visualization of different pseudo-GT.} Both Mask-Bicubic and Self-Distilled are proposed by our work. We set $\alpha=0.8$ (in~\cref{eq:mask-adjustment}) to balance sharp boundaries from masks and fine-grained details from high-res features.}
        \label{fig:pseudo-gt}
\vspace{-0.3cm}

\end{figure*}
\begin{table}[t]
    \centering
    

\resizebox{\linewidth}{!}{%

\begin{tabular}{lccccc}
 \textbf{Metrics} & {\rotatebox{55}{\textbf{Depth map}}}
 & {\rotatebox{55}{\textbf{Per-image implicit}}} 
 & {\rotatebox{55}{\textbf{2x features (LiFT)}}} 
 & {\rotatebox{55}{\textbf{Mask-Bicubic}}} 
 & {\rotatebox{55}{\textbf{Self-Distilled}}} \\
\midrule
\rowcolor[gray]{0.9}
High-Res Loss  &\circlecheck & \circlecheck & \circlecross & \circlecheck & \circlecheck \\
Task-Agnostic  &\circlecross & \circlecheck & \circlecheck & \circlecheck  & \circlecheck\\
\rowcolor[gray]{0.9}
Geometry Fidelity  & \circlecheck &\circlecheck & \circlecross & \circlecheck & \circlecheck\\
Free of Noise/Artifacts & \circlecheck & \circlecross & \circlecross & \circlecross& \circlecheck\\
\rowcolor[gray]{0.9}
Scalability & \circlecheck &\circlecross & \circlecheck & \circlecheck & \circlecheck\\
\midrule
COCO Seg. (IoU) & 56.02 & 59.08 & 52.90 & 59.87 & \textbf{61.11}\\
\rowcolor[gray]{0.9}
Cityscapes Seg. (IoU) & 44.82 &  46.05 & 35.85 & 50.43 & \textbf{53.10}\\ 
\bottomrule
\end{tabular}
}
    \caption{\textbf{Comparison of different pseudo-GT choices.}}
    \label{tab:quant-pseudo-gt}
\end{table}

\section{Training Objective}
\label{sec:method}
While the architecture of the upsampler establishes its capacity, the training objective ultimately sets the performance ceiling. 
A central challenge for training a task-agnostic feature upsampler is the absence of full-resolution groundtruth features. As a result, previous work has either relied on proxy tasks or constructed pseudo-GT at lower resolutions. For instance, FeatUp~\cite{fu2024featup} employs a multi-view reconstruction task: the predicted high-resolution features, $\hat{F}_{\text{HR}}$, are first transformed via an affine mapping $t$, then downsampled, and finally compared against the low-resolution features extracted from images that go through the same transformation, i.e., 
\begin{equation*}
    \mathcal{L}_{\text{FeatUp}} = \mathcal{D}\bigl(f(t(I)), \sigma_{\downarrow}(t(\hat{F}_{\text{HR}})\bigr).
\end{equation*} 
In contrast, LiFT~\cite{suri2024lift} directly uses features from 2$\times$ larger images $I_{2\times}$ as pseudo-GT, constraining the upsampler to predict 2$\times$ upsampled features $\hat{F}_{\text{2$\times$}}$ via
\begin{equation*}
\mathcal{L}_{\text{LiFT}}=\mathcal{D}\bigl(\hat{F}_{\text{2$\times$}}, f(I_{2\times}) \bigr). 
\end{equation*}
However, since both $\mathcal{L}_{\text{FeatUp}}$ and $\mathcal{L}_{\text{LiFT}}$ are at relatively low resolutions (only 1/16 or 1/8 of the target resolution), they provide only weak supervisory signals for capturing fine-grained details inherent in high-resolution outputs, potentially leaving the upsampled features under-constrained.

In this work, we employ a self-distillation strategy to generate high-quality pseudo-GT for supervising full-resolution features, as shown in~\cref{fig:pipeline}. First, we train an upsampler using high-resolution class-agnostic masks, which emphasize sharp boundaries and geometric awareness. In the second stage, we enhance training through self-distillation. Specifically, we first initialize a teacher and a student upsampler both from the trained upsampler in Stage 1. Then, the teacher processes high-resolution image crops, and its outputs serve as supervision for the corresponding student upsampler outputs. 
By distilling more detailed and accurate feature maps from the teacher, Stage 2 further reduces noise and enhances sharpness in the student's outputs. 
Overall, by constructing high-quality pseudo-GT feature maps at full-resolution, our approach enables the upsampler to capture more detailed structures and boundaries, ultimately pushing the limits of feature upsampling.

\subsection{Stage 1: Training with class-agnostic masks}
Low-resolution features and their upsampled features using existing upsamplers are often too blurry and noisy to be used as pseudo-GT.
In contrast, Segment Anything Model (SAM)~\cite{kirillov2023segment} produces full-resolution, class-agnostic masks that capture fine-grained details such as small object parts and boundaries. Previous work has shown that training with these masks leads to strong task generalization~\cite{kirillov2023segment, ravi2024sam2, rasheed2024glamm, ren2024grounded}. 
These qualities make them well suited for refining feature maps to reduce artifacts and noise while promoting smoothness in homogeneous regions.

To leverage these masks, we first upsample a low-resolution feature map to full resolution via bicubic interpolation, yielding $F_{\text{Bicubic}}$. Then, for each mask $m \in M=\{m_1, m_2, ..., m_N\}$, we compute the mean feature,  $\overline{{F_{\text{Bicubic}}[m]}}$, and blend it with the original features, yielding a \textit{mask-refined} feature map at pixels within mask $m$:
\begin{equation}\label{eq:mask-adjustment}
    F_{\text{Mask-Bicubic}}[m] = \alpha * \overline{{F_{\text{Bicubic}}[m]}} + (1-\alpha) * F_\text{{Bicubic}}[m],
\end{equation}
where $\alpha \in [0, 1]$ controls the degree of mask refinement.

We then supervise the upsampler using the loss:
\begin{equation}
    \mathcal{L}_{\text{Mask-Bicubic}} = ||\hat{F}_{\text{HR}}- F_{\text{Mask-Bicubic}}||_2,
\end{equation}
which encourages high-resolution outputs with  homogeneous features within each mask region, thereby leveraging the rich structural information from the class-agnostic masks. 
Note while these masks can also refine features from more complicated upsampling methods such as FeatUp~\cite{fu2024featup} or JBU~\cite{kopf2007joint}, we observe that simple bicubic upsampling already yields strong results, making it the preferred choice for Stage 1 training.

\begin{table}[t]
    \centering
    

\resizebox{\linewidth}{!}{%

\begin{tabular}{lccccc}
 \textbf{Metrics} & {\rotatebox{55}{\textbf{Depth map}}}
 & {\rotatebox{55}{\textbf{Per-image implicit}}} 
 & {\rotatebox{55}{\textbf{2x features (LiFT)}}} 
 & {\rotatebox{55}{\textbf{Mask-Bicubic}}} 
 & {\rotatebox{55}{\textbf{Self-Distilled}}} \\
\midrule
\rowcolor[gray]{0.9}
High-Res Loss  &\circlecheck & \circlecheck & \circlecross & \circlecheck & \circlecheck \\
Task-Agnostic  &\circlecross & \circlecheck & \circlecheck & \circlecheck  & \circlecheck\\
\rowcolor[gray]{0.9}
Geometry Fidelity  & \circlecheck &\circlecheck & \circlecross & \circlecheck & \circlecheck\\
Free of Noise/Artifacts & \circlecheck & \circlecross & \circlecross & \circlecross& \circlecheck\\
\rowcolor[gray]{0.9}
Scalability & \circlecheck &\circlecross & \circlecheck & \circlecheck & \circlecheck\\
\midrule
COCO Seg. (IoU) & 56.02 & 59.08 & 52.90 & 59.87 & \textbf{61.11}\\
\rowcolor[gray]{0.9}
Cityscapes Seg. (IoU) & 44.82 &  46.05 & 35.85 & 50.43 & \textbf{53.10}\\ 
\bottomrule
\end{tabular}
}
    \caption{\textbf{Comparison of different pseudo-GT choices.}}
    \label{tab:quant-pseudo-gt}
\end{table}

\begin{table*}[t]
    \centering
    
     \resizebox{0.75\textwidth}{!}{%
    \begin{tabular}{l c c c c c c}
        \toprule
        & \multicolumn{2}{c}{Semantic Seg.} & \multicolumn{2}{c}{Depth Estimation} & \multicolumn{2}{c}{Normal Estimation} \\
        \cmidrule(lr){2-3}\cmidrule(lr){4-5}\cmidrule(lr){6-7}
       Method  & COCO mIoU $\uparrow$ & CS mIoU $\uparrow$ 
        & RMSE $\downarrow$ & Recall $\uparrow$ 
        & RMSE $\downarrow$ & Recall $\uparrow$  \\
        \midrule
        Low-res             &  51.21 & 36.54&  \underline{0.1071} & \underline{89.08} & 32.29 & 69.56    \\
        Bilinear            &  56.15 & \underline{44.79} &    0.1132 & 87.68 & \underline{32.27} & \underline{70.03} \\
        LiFT                &  53.35    &  35.80    & 0.1078 & 88.71 & 32.31 & 69.78    \\
        FeatUp          &  \underline{56.30} & 44.19  &  0.1092 & 88.57 & 32.25 & 69.83 \\
        \textbf{LoftUp}               &\textbf{61.11} &  \textbf{53.10}   &   \textbf{0.0921} & \textbf{91.35} & \textbf{30.79} & \textbf{72.76}   \\
        \bottomrule
    \end{tabular}
    }
    \caption{\textbf{Comparison of feature upsamplers across tasks of semantics segmentation, depth estimation, and normal estimation.}  
    For each metric, we boldface the best performance and underline the second best.}
    \label{tab:probe}
\end{table*}
\begin{table*}[t]
    \centering
     \resizebox{\textwidth}{!}{%
    \begin{tabular}{l c c c c c c c c c}
        \toprule
        & \multicolumn{3}{c}{Video Obj. Seg.} & \multicolumn{3}{c}{Open-Voc Seg.} & \multicolumn{3}{c}{Interactive Seg.} \\
        \cmidrule(lr){2-4}\cmidrule(lr){5-7}\cmidrule(lr){8-10}
       Method  & J Mean & F Mean & J \& F Mean & COCO~\cite{lin2014microsoft} & CS~\cite{Cordts2016Cityscapes} & ADE~\cite{zhou2017scene} & GrabCut~\cite{rother2004grabcut} & Berkeley~\cite{martin2001berkeley} & DAVIS~\cite{perazzi2016davis}  \\
        \midrule
        Low-res             &  42.05 & 31.27 & 36.66 & 25.70 & 34.48 & 19.50  & 64.90 & 55.77 & 52.82 \\
        Bilinear            &  42.62 & 33.90 & 38.26  & 25.78 & 34.56 & 19.53&65.04 & 55.83& 54.26\\
        LiFT                &  \underline{47.68} & 36.78 & 42.23 & 25.96& \underline{35.39} & 19.50 & 29.66 & 31.99& 41.14 \\
        FeatUp          &  45.70 & \underline{42.90} & \underline{44.30} & \underline{26.61} & 35.01 &  \underline{19.99} &\underline{65.89} & \underline{56.67} & \underline{55.03} \\
        \textbf{LoftUp}   &\textbf{58.72} & \textbf{61.79} & \textbf{60.25}& \textbf{27.82} & \textbf{38.82} & \textbf{21.29} & \textbf{78.49} & \textbf{65.24} & \textbf{67.31}  \\
        \bottomrule
    \end{tabular}
    }
    \caption{\textbf{Comparison of feature upsamplers across tasks of video object segmentation, zero-shot open-vocabulary segmentation, and interactive segmentation.} Following previous works~\cite{lan2024proxyclip, huang2024renovating, liu2023simpleclick, kirillov2023segment, jabri2020space, suri2024lift}, we report J Mean, F Mean, and their average for video object segmentation, mIoU for open-vocabulary segementation, and IoU@1 Click for interactive segmentation. }
    \label{tab:more-applications}
    \vspace{-0.4cm}
\end{table*}

\subsection{Stage 2: Training with self-distillation}
We further enhance our upsampler training using self-distillation. To distill high-quality features, we adopt a teacher-student dual-branch design. Both the teacher and student upsamplers are initialized with the upsampler trained in Stage 1. The teacher is updated via an Exponential Moving Average (EMA) during training, ensuring a stable and progressively improving target.

For the student branch, each image is resized to a fixed resolution $I \in \mathbb{R}^{H\times W}$ (224px or 336px), matching the VFM's input requirements. 
For the teacher branch, the original image is resized to a larger size $I_{\text{HR}} \in \mathbb{R}^{tH\times tW}$ with $t \in [2,4]$.
The models then process on a crop from $I_{\text{HR}}$ that also matches the VFM's resolution, i.e., $\text{crop}(I_{\text{HR}}) \in \mathbb{R}^{H\times W}$.
Note our original images, sourced from the SA1B dataset~\cite{kirillov2023segment}, have a minimum resolution of 1500px on the shortest side and is thus larger than $tH \times tW$. While a higher $t$ yields more fine-grained teacher outputs, it also lowers the supervision resolution on the student ($H/t \times W/t$), reducing supervision effectiveness. In  practice, we find $t\in[2,4]$ strikes a good balance.
The teacher's output on this crop, $f_{\text{teacher}}\bigl(\text{crop}(I_{\text{HR}})\bigr) \in \mathbb{R}^{H\times W}$, is subsequently downsampled via $\sigma_{\downarrow}(\cdot)$ to match the resolution of the corresponding student output, $\text{crop}\bigl(f_{\text{student}}(I)\bigr) \in \mathbb{R}^{H/t \times W/t}$, and used as pseudo GT for the student.
Formally, the self-distillation loss is defined as:
\begin{equation*}
    \mathcal{L}_{\text{Self-Distilled}} = \mathcal{D}\Bigl(\sigma_{\downarrow}\bigl(f_{\text{teacher}}(\operatorname{crop}(I_{\text{HR}})\bigr),  \operatorname{crop}\bigl(f_{\text{student}}(I)\bigr)\Bigr),
\end{equation*}
where $\mathcal{D}$ denotes a discrepancy function between features.
In our experiments, we adopt the affinity matrix loss~\cite{wysoczanska2024clipdinoiser, Puy_2024_threepillars, wang2022noramlizedcut}, which consistently outperforms the standard $L_2$ loss.
This self-distillation strategy leverages the teacher's ability to handle an easier task—generating high-quality features for cropped regions—thus providing a more reliable pseudo-GT to guide the student. 
Furthermore, to capitalize on the sharp boundary priors provided by the class-agnostic masks, we apply the same mask refinement described in \cref{eq:mask-adjustment} to the teacher's features. For simplicity, we denote this refined output as $F_{\text{Self-Distilled}}$. As shown in~\cref{fig:pseudo-gt}, $F_{\text{Self-Distilled}}$ reduces the blurry artifacts in $F_{\text{Mask-Bicubic}}$ and captures the underlying geometry much better.

\boldparagraph{Further Discussion on Pseudo-GT Choices}
Earlier in this section, we identified low-resolution loss as a bottleneck in prior training objectives. Here, we expand on that discussion by addressing additional key factors uncovered during our pseudo-GT exploration. In particular,~\cref{tab:quant-pseudo-gt} compares our approach with three alternative pseudo-GT choices: (1) depth maps,  (2) per-image optimized high-resolution features from FeatUp-Implicit~\cite{fu2024featup}, and (3) features from 2$\times$ larger inputs, as in LiFT~\cite{suri2024lift}. 
First, pseudo-GT should be \textbf{task-agnostic} to ensure generalizability across downstream tasks. For example, although depth maps contain fine-grained geometric details, they lack semantic information, potentially limiting the upsampler's ability to maintain global semantic consistency.
Furthermore, pseudo-GT must \textbf{capture image geometry accurately}, maintain sharp boundaries and clear color transitions, and \textbf{minimize artifacts and noise}. As shown in~\cref{fig:pseudo-gt}, per-image optimized features exhibit halo artifacts around objects and speckle noise in smooth regions, while 2$\times$ features introduces significant mosaic artifacts. $F_{\text{Mask-Bicubic}}$ still suffers from blurriness and blobby artifacts inherited from bicubic upsampling. Among these, only $F_{\text{Self-Distilled}}$ meets these requirements, demonstrating the effectiveness of our self-distillation framework. Finally, \textbf{scalability} is another crucial factor. Per-image optimization is computationally expensive, requiring 1 minute per HR feature map on an A100 GPU and 74MB for storage —equating to 35 GPU days and 3.5TB storage for just 50K images. In contrast, other methods generate pseudo-GT in real time with a single forward pass, taking less than 0.1 seconds per image.
Overall, we find that $F_{\text{Self-Distilled}}$ best satisfies these requirements and delivers the strongest performance.

\begin{figure*}[t]
    \centering
    
    \begin{subfigure}[b]{0.16\textwidth}
         \centering
         \includegraphics[width=\textwidth]{sec/figs/vis_semseg/imgs/000000000785.jpg}
         \label{fig:image}
     \end{subfigure}
     \hfill
     \begin{subfigure}[b]{0.16\textwidth}
         \centering
         \includegraphics[width=\textwidth]{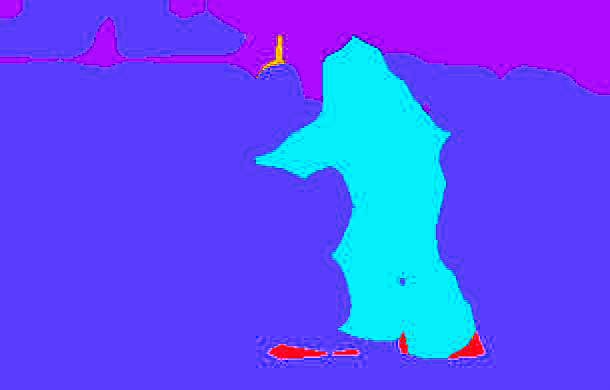}
         \label{fig:bilinear-openvoc}
     \end{subfigure}
     \hfill
     \begin{subfigure}[b]{0.16\textwidth}
         \centering
         \includegraphics[width=\textwidth]{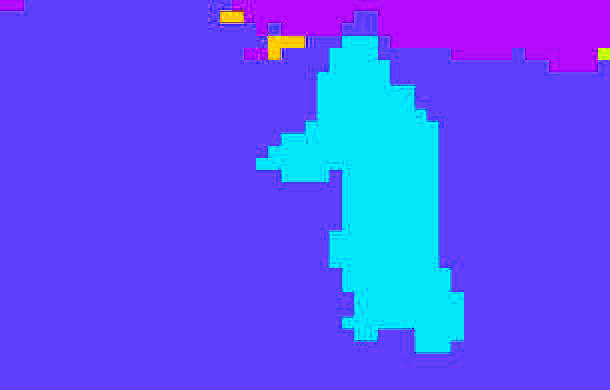}
         \label{fig:five over x}
     \end{subfigure}
     \hfill
     \begin{subfigure}[b]{0.16\textwidth}
         \centering
         \includegraphics[width=\textwidth]{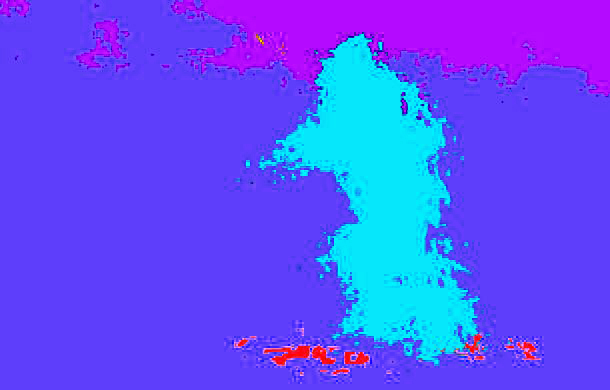}
         \label{fig:three sin x}
     \end{subfigure}
     \hfill
     \begin{subfigure}[b]{0.16\textwidth}
         \centering
         \includegraphics[width=\textwidth]{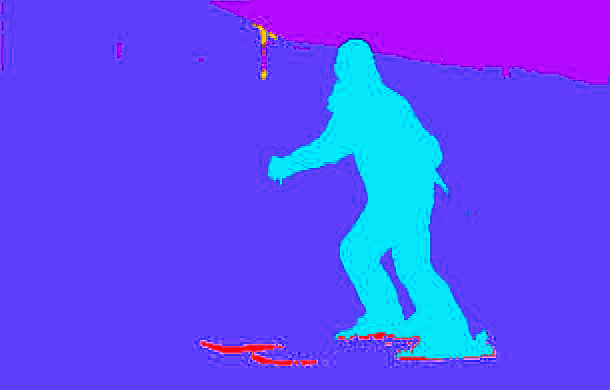}
         \label{fig:five over x}
     \end{subfigure}
     \hfill
     \begin{subfigure}[b]{0.16\textwidth}
         \centering
         \includegraphics[width=\textwidth]{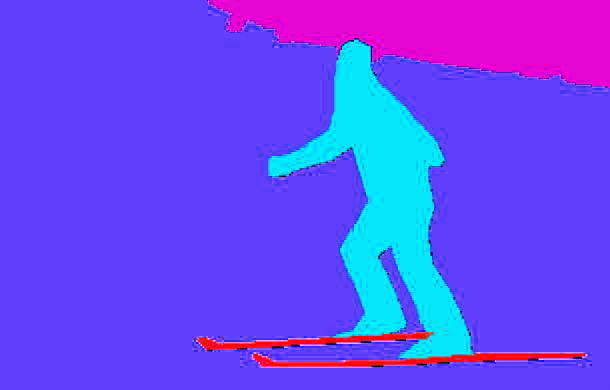}
         \label{fig:five over x}
     \end{subfigure} \\
     \vspace{-0.2cm}
     \begin{subfigure}[b]{0.16\textwidth}
         \centering
         \includegraphics[width=\textwidth]{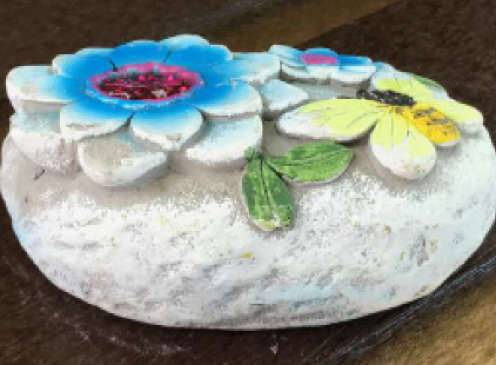}
         \caption{Original image}
         \label{fig:image}
     \end{subfigure}
     \hfill
     \begin{subfigure}[b]{0.16\textwidth}
         \centering
         \includegraphics[width=\textwidth]{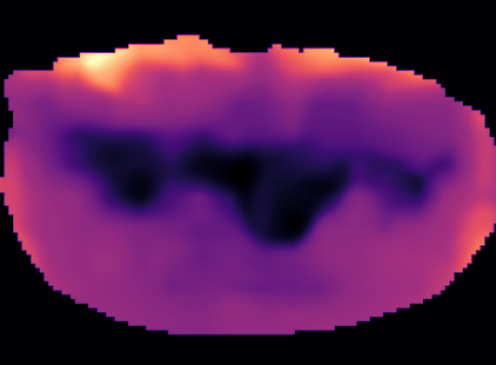}
         \caption{Bilinear}
         \label{fig:bilinear-openvoc}
     \end{subfigure}
     \hfill
     \begin{subfigure}[b]{0.16\textwidth}
         \centering
         \includegraphics[width=\textwidth]{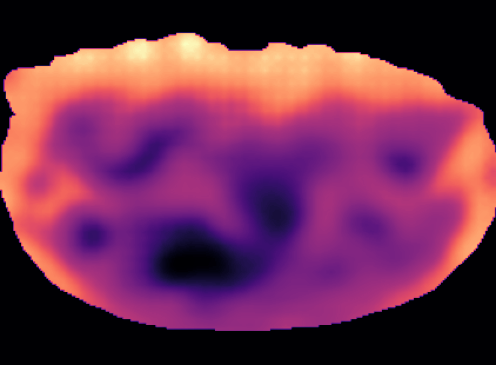}
         \caption{LiFT}
         \label{fig:five over x}
     \end{subfigure}
     \hfill
     \begin{subfigure}[b]{0.16\textwidth}
         \centering
         \includegraphics[width=\textwidth]{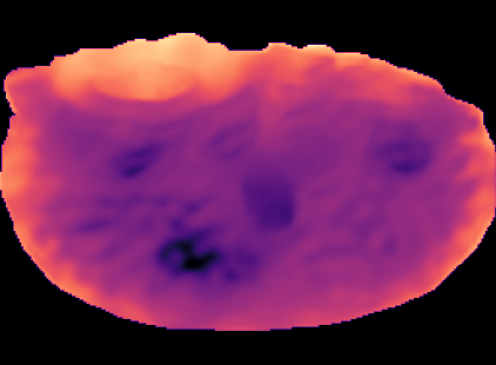}
         \caption{FeatUp}
         \label{fig:three sin x}
     \end{subfigure}
     \hfill
     \begin{subfigure}[b]{0.16\textwidth}
         \centering
         \includegraphics[width=\textwidth]{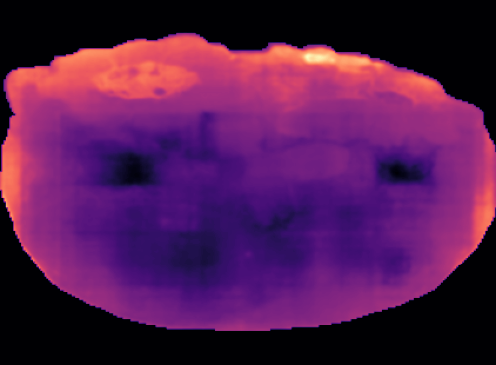}
         \caption{\textbf{LoftUp}}
         \label{fig:five over x}
     \end{subfigure}
     \hfill
     \begin{subfigure}[b]{0.16\textwidth}
         \centering
         \includegraphics[width=\textwidth]{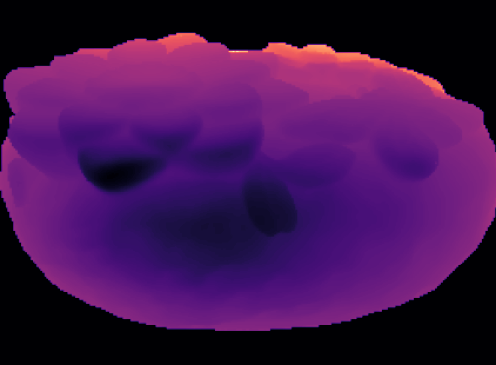}
         \caption{Groundtruth}
         \label{fig:five over x}
     \end{subfigure}
     \vspace{-0.1cm}
    \caption{\textbf{Visualization of predictions examples} on semantic segmentation on COCO-Stuff~\cite{caesar2018coco} and depth estimation on NAVI~\cite{jampani2023navi}. We provide additional visualization in the supplements.}
    \vspace{-0.3cm}
    \label{fig:vis-preds}
\end{figure*}

\section{Experiments}
\label{sec:exp}

In this section, we compare LoftUp with other upsampling approaches on various tasks and conduct in-depth analysis on the strengths of LoftUp.
By default, we use the DINOv2-S/14 model~\cite{oquab2024dinov} as  VFM. 
Experiments with CLIP~\cite{radford2021learning} and RADIO~\cite{am-radio} show that DINOv2 produces the best downstream performance. Detailed results for CLIP and RADIO are provided in the supplements. 
Following previous works~\cite{fu2024featup, suri2024lift}, we resize input images to a resolution of 224 and upsample the VFM features to match the input resolution (14$\times$ for DINOv2-S/14). All upsamplers are trained on a 1M-image subset of the SA1B dataset~\cite{kirillov2023segment} and additional implementation details are available in the supplements.

\subsection{Cross-Task Comparison}
We compare LoftUp with two most recent task-agnostic feature upsamplers: FeatUp (JBU)~\cite{fu2024featup} and LiFT~\cite{suri2024lift}, along with bilinear upsampling as a baseline.
As demonstrated in~\cref{tab:probe} and~\cref{tab:more-applications}, LoftUp consistently outperforms all alternatives on six downstream tasks.

Since we expect high-resolution features to \textit{enhance fine-grained scene understanding}, we first evaluate them on \textbf{semantic segmentation} following~\cite{fu2024featup,hamilton2022unsupervised}. We train a linear projection on upsampled features to predict coarse classes in COCO-Stuff~\cite{caesar2018coco} and Cityscapes~\cite{Cordts2016Cityscapes}. LoftUp achieves 7.3\% and 15.6\% relative improvements over previous best methods. As expected, Cityscapes benefits more from feature upsampling due to its high number of small objects.

To assess the upsamplers' \textit{generalizability to geometry prediction},
we follow~\cite{el2024probing} and evaluate \textbf{depth and normal estimation} using a lightweight DPT decoder head~\cite{Ranftl_2021_dpt} on the NAVI dataset~\cite{jampani2023navi}. Among the compared methods, LoftUp is the \textit{only} upsampling method that significantly outperforms the low-res baseline, achieving 2.5\% and 4.4\% relative improvements in recall.


We further evaluate the \textit{temporal consistency} of upsampled features through \textbf{video object segmentation} on DAVIS2017~\cite{Pont-Tuset_arXiv_2017_davis}, which requires consistent object tracking across video frames. Following prior works~\cite{suri2024lift, jabri2020space}, we use feature affinity maps to track objects across frames and report J Mean (mIoU, reflecting region similarity), F Mean (mean F-score, reflecting contour accuracy) and their average. LoftUp achieves significant performance improvements, with a 39.6\% increase in J Mean and a 97.6\% increase in F Mean over the low-resolution baseline, demonstrating a strong ability to identify object boundaries.

Finally, to assess the \textit{flexibility of usage across modalities} like text and click embeddings, we evaluate on \textbf{zero-shot open-vocabulary segmentation} with ProxyCLIP~\cite{lan2024proxyclip} where DINOv2 features adjust CLIP features spatially for better vision-language alignment, and 
on \textbf{interactive segmentation} using a modified SimpleClick~\cite{liu2023simpleclick} where visual features are incorporated with click embeddings for segmentation. Results show that LoftUp is the \textit{only} method to achieve consistent, significant improvements over the low-res and bilinear baselines, demonstrating its adaptability to multimodal alignment.

\begin{figure}
    \centering
    \vspace{-0.4cm}
    \includegraphics[width=0.8\linewidth]{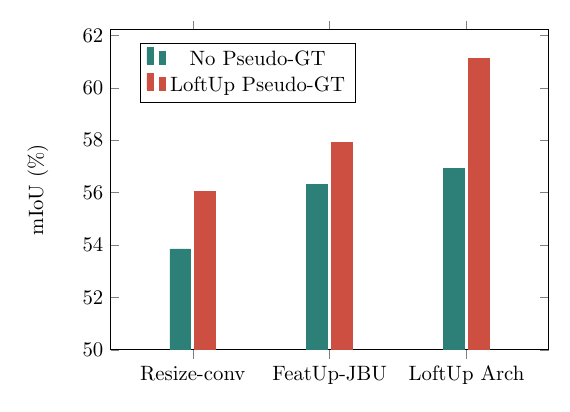}
    \vspace{-0.3cm}
    \caption{\textbf{LoftUp pseudo-GT improves different upsamplers.}}
    \label{fig:pseudo-gt-loss}
\end{figure}

\begin{table}[t]
  \centering
  \resizebox{0.47\textwidth}{!}{%
  \begin{tabular}{l|cccc}
  \toprule
    Architecture &  COCO & Cityscapes & Depth Rec. & Normal Rec.\\
     \midrule
    Low-res & 51.21 & 36.54 & 89.08 & \underline{69.56}\\
 Resize-conv         &   56.05 & \underline{45.94}  & 88.38 & 65.59\\ 
        LIIF & 52.24 & 40.36 & 79.85 & 46.06 \\
        FeatUp-JBU & \underline{57.90} & 44.83 & \underline{89.38} & 68.83 \\ 
    \textbf{LoftUp} & \textbf{61.11} & \textbf{53.10} & \textbf{91.35} & \textbf{72.61}\\
       \bottomrule
  \end{tabular}
  }
  \caption{\textbf{Architecture comparison} on semantic segmentation and depth and normal estimation. All upsamplers are trained using  LoftUp training objective.}
  \label{tab:arch-compare}
  \vspace{-0.5cm}
\end{table}

\begin{figure}[t]
     \centering
     \begin{subfigure}[b]{0.4\linewidth}
         \centering
         \includegraphics[width=\linewidth]{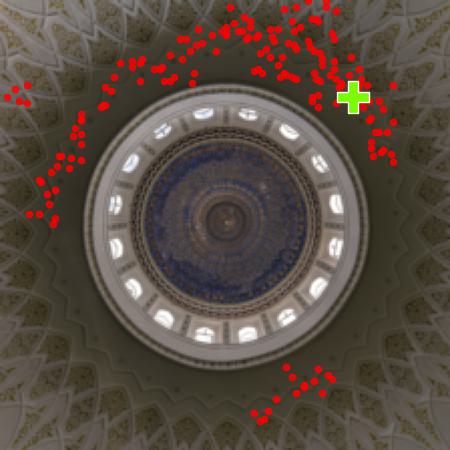}
         \label{fig:original}
     \end{subfigure}
     \begin{subfigure}[b]{0.4\linewidth}
         \centering
         \includegraphics[width=\linewidth]{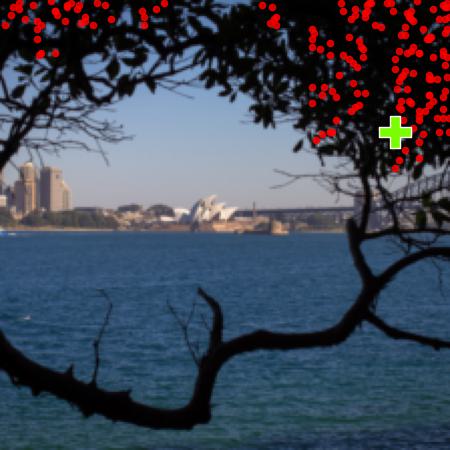}
         \label{fig:original}
     \end{subfigure} \\
     \vspace{-0.3cm}
     \begin{subfigure}[b]{0.4\linewidth}
         \centering
         \includegraphics[width=\linewidth]{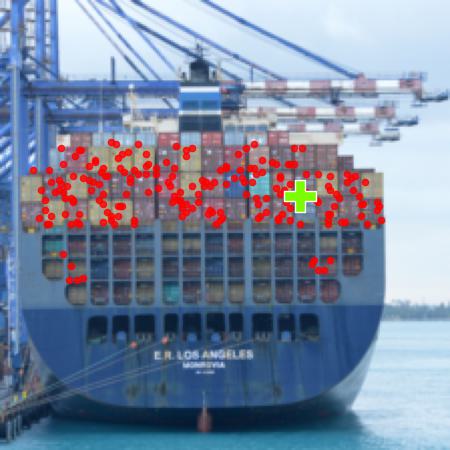}
         \label{fig:original}
     \end{subfigure}
     \begin{subfigure}[b]{0.4\linewidth}
         \centering
         \includegraphics[width=\linewidth]{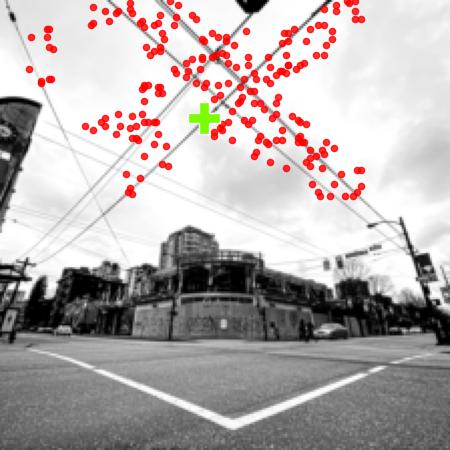}
         \label{fig:original}
     \end{subfigure} \\
\vspace{-0.3cm}

\caption{\textbf{Visualization of attended regions} (shown as \textcolor{red}{dots}) in the low-resolution features corresponding to a high-resolution pixel (marked as a \textcolor{green}{cross}). The density of \textcolor{red}{dots} represents the values in the attention map. LoftUp leverages relevant information from the global feature map to upsample features at each pixel.}
        \label{fig:attn-map}
\vspace{-0.4cm}

\end{figure}
\subsection{More Analysis of LoftUp Design}
\vspace{-0.2cm}
\boldparagraph{Our pseudo-GT provides strong training signal} In~\cref{fig:pseudo-gt} and~\cref{tab:quant-pseudo-gt}, we already discuss and compare different pseudo-GT choices both qualitatively and quantitatively by training LoftUp architecture using different pseudo-GTs. In~\cref{fig:pseudo-gt-loss}, we further show that our pseudo-GT can be used to enhance other upsamplers such as resize-conv and FeatUp (JBU), demonstrating broad applicability of the strong, full-resolution supervision signal of our pseudo-GT alone.

\vspace{0.15cm}
\noindent\textbf{Our architecture significantly outperforms other alternatives} when training under the same LoftUp training objective as in~\cref{sec:method}.
As shown in~\cref{tab:arch-compare}, we compare LoftUp with resize-conv, LIIF, and FeatUp-JBU. We demonstrate that while other methods may bias towards certain tasks, LoftUp achieves the strongest performance across all of the tasks, indicating its stronger capacity to leverage the high-quality pseudo-GT.

\boldparagraph{LoftUp benefits from the long-range, content-aware image-feature interaction thanks to the cross-attention mechanism}
Instead of using a local kernel as in most previous feature upsampling works, our cross-attention mechanism effectively allows global attention and content-aware upsampling. 
In~\cref{fig:attn-map}, we visualize the attended region of a pixel in high-res coordinate (in \textcolor{green}{cross}) in the corresponding low-resolution features (in \textcolor{red}{dots}) where the density of \textcolor{red}{dots} reflects the values of attention maps.
As shown, the attended regions often exhibit semantically similar patterns across the image. Consequently, our upsampler leverages relevant information from the entire image, enhancing global semantic consistency and refining details at object boundaries.

\begin{table}[t]
  \centering
  \resizebox{0.45\textwidth}{!}{%
  \begin{tabular}{l|cc|cc}
  \toprule
    Method &  Input res & Feature res & COCO & Cityscapes\\
     \midrule
    Low-res  & 224 & 16  & 51.21 & 36.54 \\
    2x-large input & 448 & 32 & 55.42& 50.29\\
    4x-large input & 896 & 64 & 56.79 & 58.45\\
     \midrule
     \multirow{8}{*}{LoftUp}  & \multirow{4}{*}{224} & 56 & 59.87 & 48.20 \\
     &  & 112 & 60.75 & 51.04\\
     &  & 224 & \textbf{61.11} & \textbf{53.10}\\
     &  & 448 & 60.61 & 52.29\\
    \cmidrule(lr){2-5}
     &  \multirow{4}{*}{448} & 56 & 60.16 & 51.57\\
     &  & 112 & 61.50 & 56.46 \\
      &  & 224  & 61.78 & 58.55\\
     &  & 448 & \textbf{62.26} & \textbf{61.24}\\
       \bottomrule
  \end{tabular}
  }
  \caption{\textbf{LoftUp maintains strong performance with diverse input sizes and upsampling scales.}}
  \label{tab:any-res}
\end{table}

\boldparagraph{LoftUp supports arbitrary upsampling scales via its coordinate-based design}
As shown in~\cref{tab:any-res}, LoftUp maintains strong performance across varying input and feature resolutions. Here, ``input res'' refers to the image resolution fed into the VFM, while ``feature res'' denotes the final feature map size generated by LoftUp. Notably, LoftUp outperforms the original VFM backbone even when the latter processes images at 2$\times$ resolution—an approach that incurs nearly 4$\times$ the computational and memory cost. This highlights LoftUp's efficiency and adaptability, making it well-suited for diverse scaling needs in downstream tasks.

\boldparagraph{LoftUp also enjoys relatively high efficiency} In~\cref{tab:efficiency}, we compare the parameter size and inference time of each upsampler. Interestingly, while LoftUp has slightly more parameters than most alternatives (still less than 20\% of the VFM), it is the second fastest upsampler after LiFT, which only does 2x upsampling. 
This efficiency arises because LoftUp directly maps coordinates to features, bypassing multiple intermediate upsampling layers used by other methods.
Moreover, compared to the coordinate-based FeatUp-Implicit that requires per-image optimization, LoftUp is both significantly faster and more effective.

\begin{table}[t]
  \resizebox{0.5\textwidth}{!}{%
  \begin{tabular}{l|cc|cc}
  \toprule
    Architecture & COCO & Cityscapes & Params (M) & Infer time (s/img) \\
     \midrule
    Bilinear & 56.15 & 44.79 & 22.1 & 0.0604\\
    \midrule
    Resize-conv & 56.05 & \underline{45.94} & 27.5 (+5.4)& 0.0922 \\
    LIIF & 52.24 & 40.36 & 24.0 (+1.9) & 0.1757\\
    LiFT& 53.35 & 35.80& 23.3 (+1.2) & \textbf{0.0773} \\
    FeatUp-JBU & \underline{57.90} & 44.83 & \textbf{22.3 (+0.2)} & 0.1213\\
    FeatUp-Implicit & 51.12 & 44.81& \underline{22.5 (+0.4)} & 54.302 \\
    LoftUp & \textbf{61.11} & \textbf{53.10} & 26.4 (+4.3) & \underline{0.0893}\\
       \bottomrule
  \end{tabular}
  }
  
  \caption{\textbf{Comparison of efficiency for different upsampler architectures.} Inference time is computed on a single A100 GPU.}
  \vspace{-0.3cm}
  \label{tab:efficiency}
\end{table}

\section{Conclusion}
In this work, we systematically explored upsampler architectures and training objectives for feature upsampling and introduced LoftUp, a coordinate-based cross-attention transformer upsampler trained with our proposed self-distilled high-resolution pseudo-GT.
Our experiments demonstrate that LoftUp generates sharp, fine-grained feature maps, enables global content-aware image-feature interaction, supports arbitrary upsampling scales, and achieves fast inference. Notably, LoftUp outperforms existing methods across six diverse downstream tasks. We hope our work inspires further research in feature upsampling and contributes to the vision community by enhancing VFMs for a wide range of downstream applications.

{
    \small
    \bibliographystyle{ieeenat_fullname}
    \bibliography{main}
}

\clearpage

\maketitlesupplementary
This supplementary material to the main paper \textbf{``LoftUp: earning a Coordinate-Based Feature Upsampler for Vision Foundation Models''} is structured as follows:

\begin{itemize}
    \item In~\cref{appx:application-detail}, we explain more implementation details of LoftUp training and the downstream tasks.
    \item In~\cref{appx:more-exps}, we show more quantitative results of LoftUp with CLIP and RADIO backbones and ablate the architecture and dataset size choices of training LoftUp.
    \item In~\cref{appx:visualization}, we provide more visualization of upsampled features, prediction results on various tasks, pseudo-GT, and cross-attention regions.
\end{itemize}

\appendix
\counterwithin{figure}{section} 
\counterwithin{table}{section}
\section{More Implementation Details}\label{appx:application-detail}

\subsection{Training details of LoftUp}
Our LoftUp upsampler is a 2-block cross-attention transformer that incorporates high-res image inputs with coordinates using an additional convolutional layer and low-res VFM features as keys and values in the cross-attention layers. Each transformer block consists of 1 cross-attention layer and 1 feedforward layer as in ViT~\cite{dosovitskiy2020vit}.
To train LoftUp, we use a batch size of 8 and AdamW~\cite{loshchilov2017decoupled} optimizer with a learning rate of 1e-3  in Stage 1 and 1e-4 in Stage 2 for more stable improvement during self-distillation. 
In Stage 2, we take 2 random crops per image to construct $\text{crop}\bigl({I_{\text{HR}}}\bigr)$, and update the teacher upsampler's weights every 10 steps using the EMA of the student upsampler with a decay factor of 0.99.
In both stages, we use $\alpha=0.8$ for mask refinement when constructing pseudo-GT to balance sharp boundaries from masks and the fine-grained details from high-resolution features within each mask region. 
For all upsamplers, including our compared ones, we train for 1 epoch on a 1M-image subset of SA1B dataset~\cite{kirillov2023segment}.

\subsection{Task setups}

\boldparagraph{Semantic segmentation} 
Following~\cite{fu2024featup,hamilton2022unsupervised}, we perform semantic segmentation on coarse classes in COCO-Stuff~\cite{caesar2018coco} (27 classes) and Cityscapes~\cite{Cordts2016Cityscapes} (19 classes) and report mean Intersection-over-Union (mIoU) for each dataset. We train a linear decoder layer on upsampled features with a batch size of 8 and AdamW optimizer~\cite{loshchilov2017decoupled} with a learning rate of 1e-4 for 10 epochs.

\boldparagraph{Depth and normal estimation}
Following~\cite{el2024probing}, we evaluate depth and normal estimation using NAVI dataset~\cite{jampani2023navi} and train a DPT decoder head with 7 convolutional layers on top of the VFM features. We use a batch size of 8  and AdamW optimizer~\cite{loshchilov2017decoupled} with a learning rate of 5e-4 for 10 epochs.
Following prior works~\cite{eigen2014depth, el2024probing, fouhey2015single}, we report the root-mean-squared prediction error (RMSE) for both tasks and recall at $\delta_3$ for scale-invariant depth estimation and at 30$^{\circ}$ for normal estimation. Here $\delta_3$ is computed as the number of pixels whose ratio of depth prediction to groundtruth is less than $1.25^3$:
\begin{equation*}
\delta_3(d^{pr}, d^{gt}) = \frac{1}{N} \sum_{j \in N} \max \left(\frac{d_j^{pr}}{d_j^{gt}}, \frac{d_j^{gt}}{d_j^{pr}}\right) < 1.25^3,
\end{equation*}
where $d^{pr}$ is predicted depth and $d^{gt}$ is groundtruth depth.

\boldparagraph{Video object segmentation}
This task involves propagating an object segmentation mask across video frames, given the ground truth mask for the first frame. Following prior evaluation protocols~\cite{suri2024lift, jabri2020space}, we compute dense feature affinity maps between frames to track objects. Performance is assessed using three metrics: J Mean, F Mean, and J \& F Mean. Specifically, J Mean denotes the average Intersection-over-Union (IoU) between predicted segmentations and groundtruth masks, while F Mean represents the average F-score, measuring contour accuracy via precision and recall against groundtruth boundaries.We evaluate our method on the DAVIS validation set~\cite{Pont-Tuset_arXiv_2017_davis}, a popular benchmark for video object segmentation. The dataset comprises 30 videos of varying lengths, each containing between 1 and 4 objects.

\boldparagraph{Zero-shot Open-Vocabulary Segmentation} We incorporate upsampled VFM features into ProxyCLIP~\cite{lan2024proxyclip}, a state-of-the-art method for zero-shot open-vocabulary segmentation (OVSeg), and evaluate on three popular OVSeg benchmarks: COCO~\cite{lin2014microsoft}, Cityscapes~\cite{Cordts2016Cityscapes}, and ADE20K~\cite{zhou2017scene}. ProxyCLIP enhances CLIP features by leveraging spatial feature correspondence from VFMs as proxy attention, effectively inheriting the strong local consistency of VFMs while retaining CLIP's remarkable zero-shot transferability. Due to the high computational cost of proxy attention, we perform upsampling to 8$\times$ for all upsampling methods. We use CLIP ViT-B/16~\cite{radford2021learning} as the CLIP backbone, DINOv2-S/14~\cite{oquab2024dinov} as the proxy VFM, and set the input resolution to 336px, matching the resolution of the CLIP backbone.

\boldparagraph{Interactive Segmentation} We adapt the SimpleClick~\cite{liu2023simpleclick} architecture to evaluate upsampled features. Specifically, we use a frozen VFM backbone and train a single-layer click encoder that directly adds to the image patch embedding, along with a three-layer convolutional decoder head on top of the upsampled features for interactive segmentation.
For training, we follow prior works~\cite{liu2023simpleclick, sofiiuk2022reviving} and use the SBD dataset~\cite{hariharan2011semantic} to train for 20 epochs with the normalized focal loss~\cite{sofiiuk2019adaptis, sofiiuk2022reviving}. We employ the Adam optimizer~\cite{kingma2014adam} with a learning rate of 5e-5  and a batch size of 8.
For evaluation, following common practice~\cite{liu2023simpleclick, sofiiuk2022reviving, kirillov2023segment}, we sample the first click point as the farthest point from the object boundary, and report the mean IoU of the predicted segmentation masks with the groundtruth, denoted as IoU@1 Click. We report results on three popular interactive segmentation benchmarks: GrabCut~\cite{rother2004grabcut}, Berkeley~\cite{martin2001berkeley}, and DAVIS~\cite{perazzi2016davis}.

\section{More Quantitative Results}\label{appx:more-exps}

\begin{table}[th]
    \centering
    \resizebox{\linewidth}{!}{%
    \begin{tabular}{cc|cc}
    \toprule
       Resolution & Upsampler & COCO & Cityscapes\\
       \midrule
     \multirow{4}{*}{224}   & NA  &  40.30& 30.79 \\
      & Bilinear    & 47.12  & 39.84 \\
       & FeatUp  & 52.08 & 33.50 \\
       & LoftUp & \textbf{52.58} & \textbf{44.66} \\
         \midrule
  \multirow{4}{*}{448} & NA &  42.14 &37.36 \\
  & Bilinear  & 48.32& 45.58\\
  & FeatUp  &  52.55 & 40.00\\
   & LoftUp & \textbf{53.87}& \textbf{50.14} \\
 \bottomrule
    \end{tabular}
    }
    \caption{\textbf{Comparison of feature upsampers when VFM is CLIP-B/16~\cite{radford2021learning}.}}
    \label{tab:clip}
\end{table}

\begin{table}[th]
    \centering
    \resizebox{\linewidth}{!}{%
    \begin{tabular}{cc|cc}
    \toprule
       Resolution & Upsampler & COCO & Cityscapes\\
       \midrule
  \multirow{4}{*}{224}      & NA  &  51.00 & 34.42\\
      & Bilinear    &  56.77 & 43.09\\
       & FeatUp  & 56.59  & 42.23\\
       & LoftUp & \textbf{58.36} & \textbf{46.98} \\
         \midrule
 \multirow{4}{*}{448}  &NA & 58.94 & 49.30\\
  &Bilinear & 62.29 & 57.42 \\
  & FeatUp &  62.18&  56.58\\
  & LoftUp & \textbf{63.55} & \textbf{60.83}\\ 
 \bottomrule
    \end{tabular}
    }
    \caption{\textbf{Comparison of feature upsampers when VFM is RADIOv2.5-B~\cite{am-radio}.}}
    \label{tab:radio}
\end{table}

\boldparagraph{Upsampling CLIP and RADIO} In~\cref{tab:clip} and~\cref{tab:radio}, we compare LoftUp with FeatUp and a bilinear upsampling baseline using CLIP~\cite{radford2021learning} and RADIO~\cite{am-radio} as VFM backbones. The upsamplers are trained following the same procedure as on the DINOv2 backbone and evaluated at resolutions 224 and 448. As with DINOv2, LoftUp consistently outperforms all baselines when using CLIP and RADIO as VFM backbone, demonstrating the general applicability of our approach across different VFMs.

\boldparagraph{Ablation on the architecture and training data size} In~\cref{tab:arch-ablation}, we conduct an ablation study on both the architecture components of LoftUp and the training data size. For convenience, the upsamplers are trained using only the Stage 1 training objective.
Specifically, in experiment (a), we demonstrate that employing a sinusoidal positional encoding for the low-resolution features—combined with a 3x3 convolutional layer to process the high-resolution coordinates and image inputs—yields improved performance. This result is in line with prior work showing that sinusoidal positional encodings excel in coordinate-based methods~\cite{yu2020pixelnerf, mildenhall2021nerf, Ye_2023_featurenerf} and that stronger image processing layers help better integrate high-resolution information.
In experiment (b), we observe that two blocks of the cross-attention transformer are sufficient for optimal feature upsampling, with performance saturating at greater depths. Finally, in experiment (c), we find that training with a larger dataset improves performance, although the benefits begin to diminish as the dataset size increases. Consequently, we select a 1M-subset of the SA1B dataset~\cite{kirillov2023segment} to achieve the best balance among data diversity, model performance, and training time.

\begin{table}[t]
  \centering
  \resizebox{\linewidth}{!}{%
  \begin{tabular}{ccccc|cc}
  \toprule
   & Feat PE & Image conv &  \# blocks  & \# Train data  & COCO & Cityscapes\\
     \midrule
\multirow{5}{*}{(a)}& no & 1x1 & 2  & 50k & 56.46  & 43.13 \\
             & learnable & 1x1 & 2 & 50k & 57.89 & 46.06 \\
           & learnable & 3x3& 2     & 50k & 58.40 & 47.35 \\
              & Sine & 1x1& 2 & 50k & 58.10 & 47.24 \\
              & Sine & 3x3 & 2& 50k & \textbf{58.65} & \textbf{48.63} \\
  \midrule
\multirow{3}{*}{(b)}   & Sine & 3x3 &  1& 50k & 57.94 & 48.13 \\
 & Sine & 3x3 & 2& 50k & \textbf{58.65} & \textbf{48.63} \\
 & Sine & 3x3 &  3& 50k & 58.35 & 48.32\\
 \midrule
\multirow{3}{*}{(c)} & Sine & 3x3&  2 & 50k & 58.65 & 48.63 \\
 & Sine & 3x3 &  2& 200k & 59.40 & 49.84\\ 
 & Sine & 3x3 & 2  & 1M & \textbf{59.87} & \textbf{50.43}\\  


       \bottomrule
  \end{tabular}
  }
  \vspace{0.1cm}
  \caption{\textbf{Ablation of architecture and training data size choices.} Upsamplers are trained using only Stage 1 loss for convenience. }
  \label{tab:arch-ablation}
\end{table}

\section{More Visualization}\label{appx:visualization}

\begin{figure*}[t]
    \centering
    \begin{subfigure}[b]{0.12\textwidth}
         \centering
         \includegraphics[width=\textwidth]{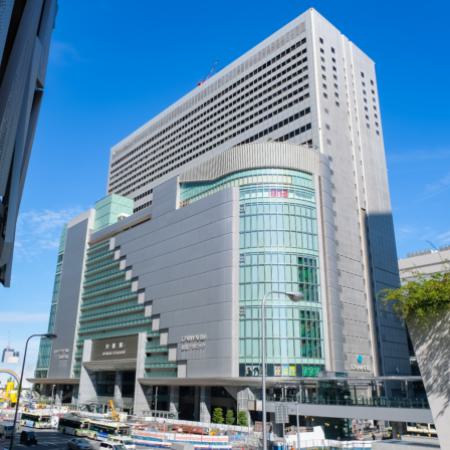}
         \label{fig:image}
     \end{subfigure}
     \hfill
     \begin{subfigure}[b]{0.12\textwidth}
         \centering
         \includegraphics[width=\textwidth]{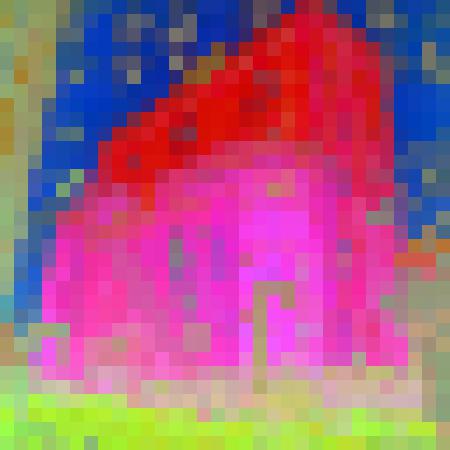}
         \label{fig:image}
     \end{subfigure}
     \hfill
     \begin{subfigure}[b]{0.12\textwidth}
         \centering
         \includegraphics[width=\textwidth]{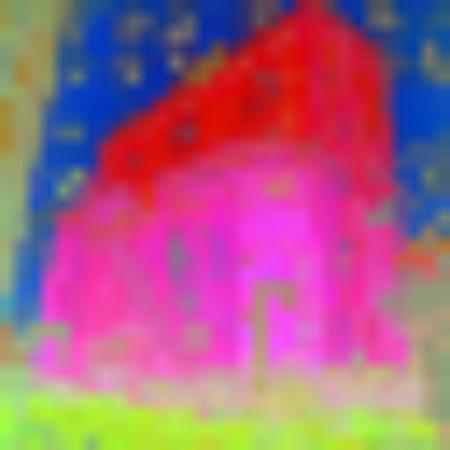}
         \label{fig:bilinear-openvoc}
     \end{subfigure}
     \hfill
     \begin{subfigure}[b]{0.12\textwidth}
         \centering
         \includegraphics[width=\textwidth]{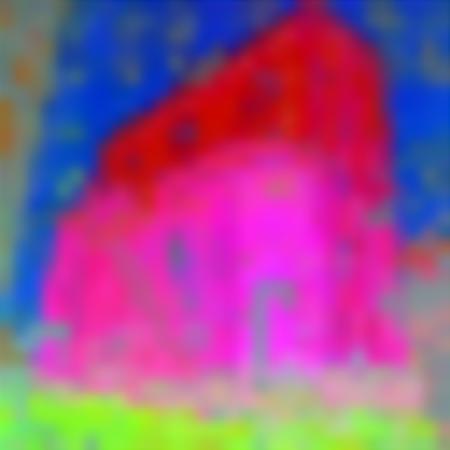}
         \label{fig:three sin x}
     \end{subfigure}
     \hfill
     \begin{subfigure}[b]{0.12\textwidth}
         \centering
         \includegraphics[width=\textwidth]{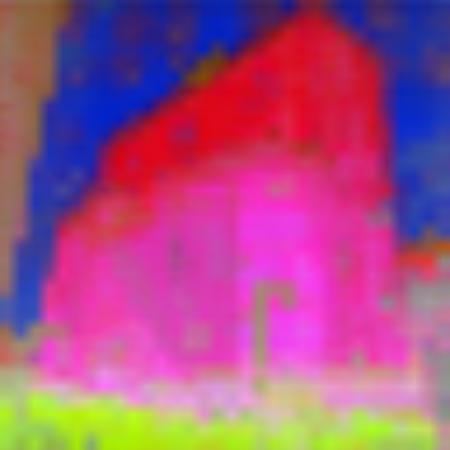}
         \label{fig:five over x}
     \end{subfigure}
     \hfill
     \begin{subfigure}[b]{0.12\textwidth}
         \centering
         \includegraphics[width=\textwidth]{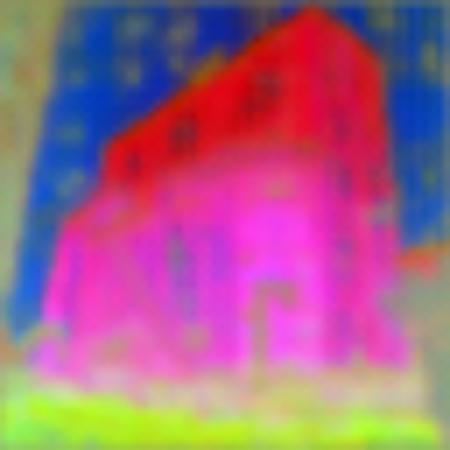}
         \label{fig:five over x}
     \end{subfigure}
     \hfill
     \begin{subfigure}[b]{0.12\textwidth}
         \centering
         \includegraphics[width=\textwidth]{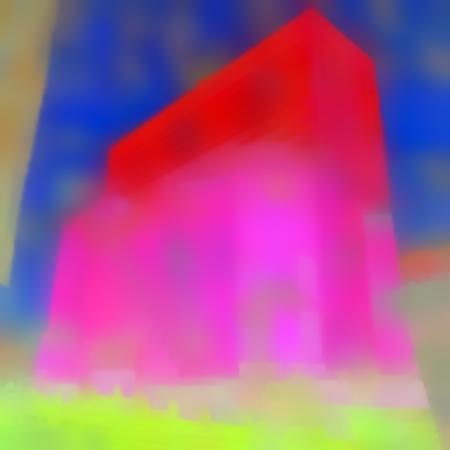}
         \label{fig:five over x}
     \end{subfigure}
     \hfill
     \begin{subfigure}[b]{0.12\textwidth}
         \centering
         \includegraphics[width=\textwidth]{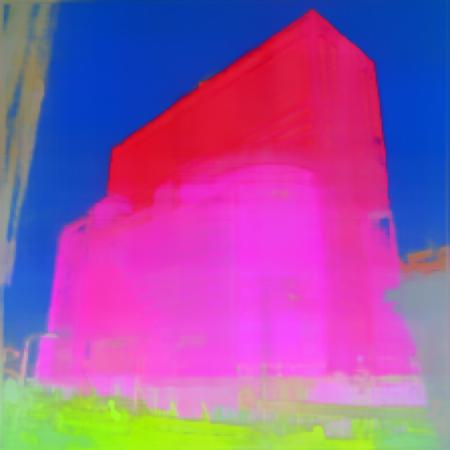}
         \label{fig:five over x}
     \end{subfigure} \\
     \vspace{-0.2cm}
      \begin{subfigure}[b]{0.12\textwidth}
         \centering
         \includegraphics[width=\textwidth]{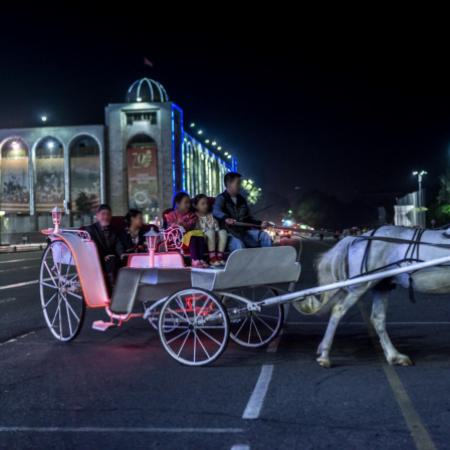}
         \label{fig:image}
     \end{subfigure}
     \hfill
     \begin{subfigure}[b]{0.12\textwidth}
         \centering
         \includegraphics[width=\textwidth]{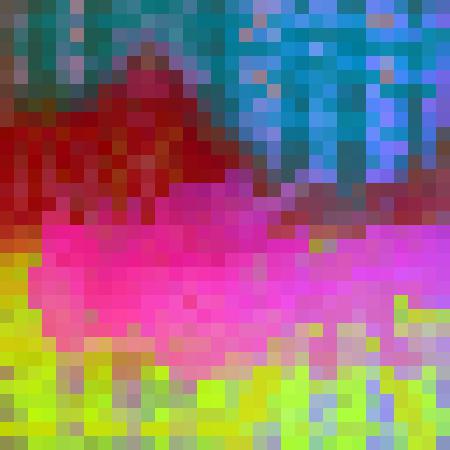}
         \label{fig:image}
     \end{subfigure}
     \hfill
     \begin{subfigure}[b]{0.12\textwidth}
         \centering
         \includegraphics[width=\textwidth]{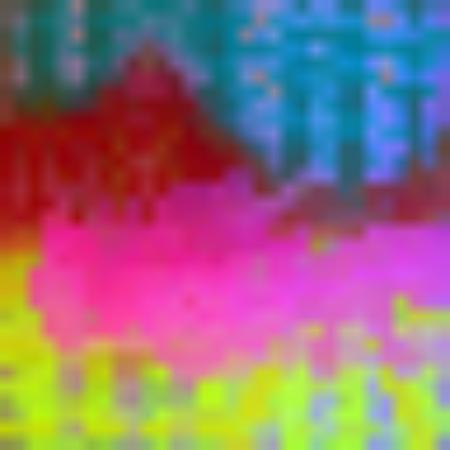}
         \label{fig:bilinear-openvoc}
     \end{subfigure}
     \hfill
     \begin{subfigure}[b]{0.12\textwidth}
         \centering
         \includegraphics[width=\textwidth]{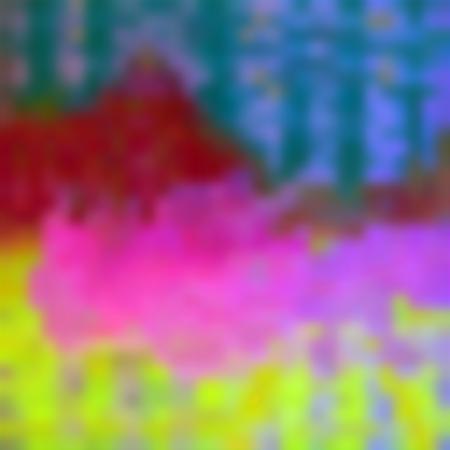}
         \label{fig:three sin x}
     \end{subfigure}
     \hfill
     \begin{subfigure}[b]{0.12\textwidth}
         \centering
         \includegraphics[width=\textwidth]{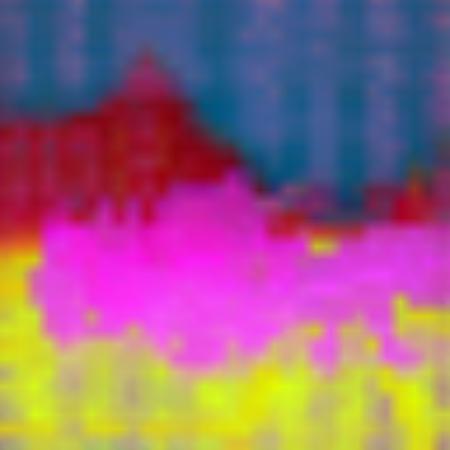}
         \label{fig:five over x}
     \end{subfigure}
     \hfill
     \begin{subfigure}[b]{0.12\textwidth}
         \centering
         \includegraphics[width=\textwidth]{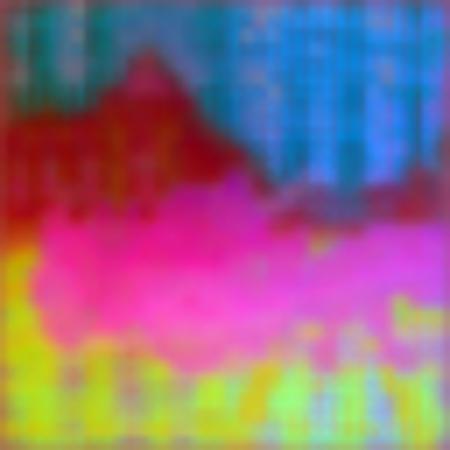}
         \label{fig:five over x}
     \end{subfigure}
     \hfill
     \begin{subfigure}[b]{0.12\textwidth}
         \centering
         \includegraphics[width=\textwidth]{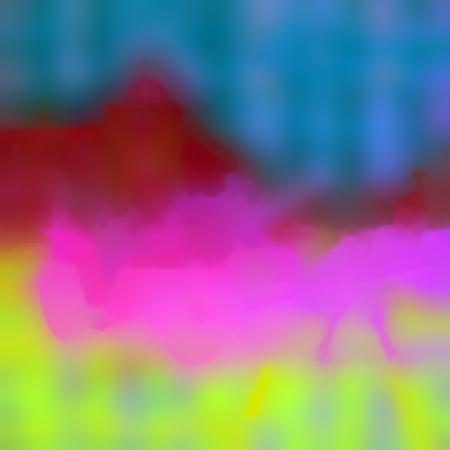}
         \label{fig:five over x}
     \end{subfigure}
     \hfill
     \begin{subfigure}[b]{0.12\textwidth}
         \centering
         \includegraphics[width=\textwidth]{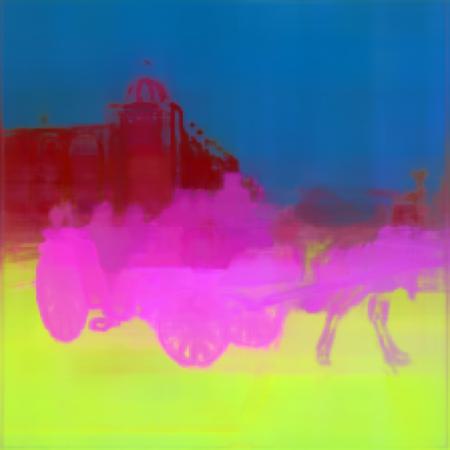}
         \label{fig:five over x}
     \end{subfigure} \\
     \vspace{-0.2cm}
      \begin{subfigure}[b]{0.12\textwidth}
         \centering
         \includegraphics[width=\textwidth]{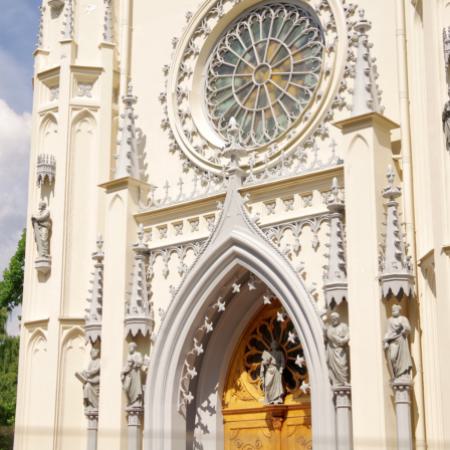}
         \label{fig:image}
     \end{subfigure}
     \hfill
     \begin{subfigure}[b]{0.12\textwidth}
         \centering
         \includegraphics[width=\textwidth]{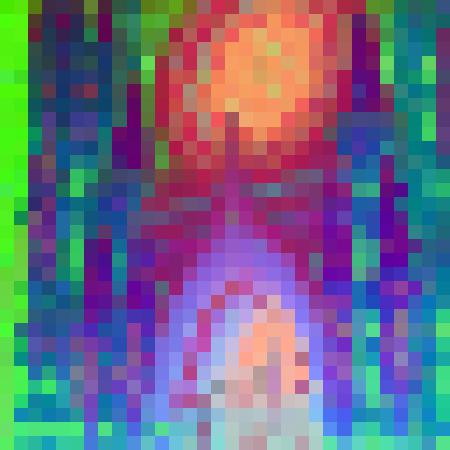}
         \label{fig:image}
     \end{subfigure}
     \hfill
     \begin{subfigure}[b]{0.12\textwidth}
         \centering
         \includegraphics[width=\textwidth]{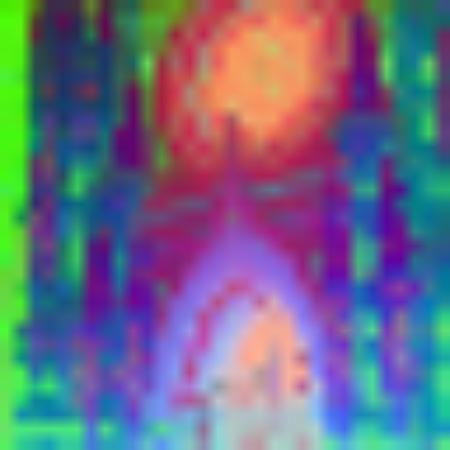}
         \label{fig:bilinear-openvoc}
     \end{subfigure}
     \hfill
     \begin{subfigure}[b]{0.12\textwidth}
         \centering
         \includegraphics[width=\textwidth]{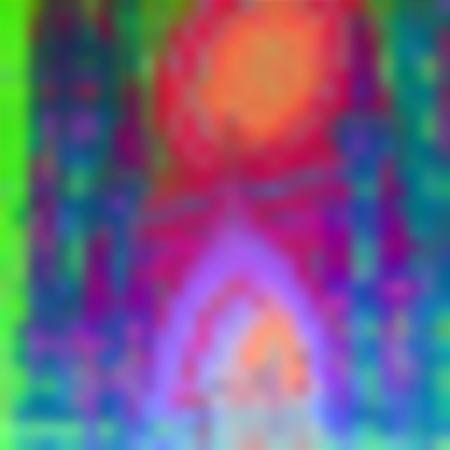}
         \label{fig:three sin x}
     \end{subfigure}
     \hfill
     \begin{subfigure}[b]{0.12\textwidth}
         \centering
         \includegraphics[width=\textwidth]{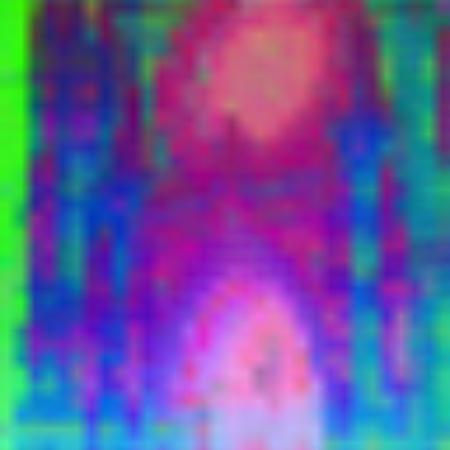}
         \label{fig:five over x}
     \end{subfigure}
     \hfill
     \begin{subfigure}[b]{0.12\textwidth}
         \centering
         \includegraphics[width=\textwidth]{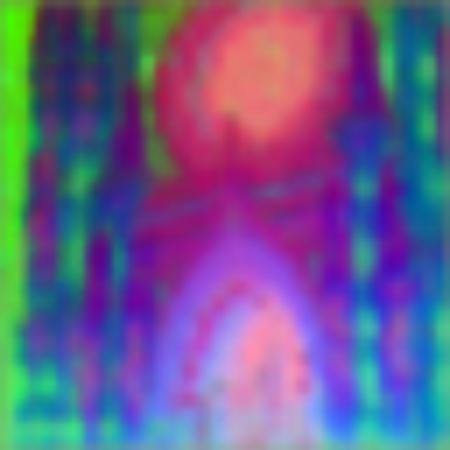}
         \label{fig:five over x}
     \end{subfigure}
     \hfill
     \begin{subfigure}[b]{0.12\textwidth}
         \centering
         \includegraphics[width=\textwidth]{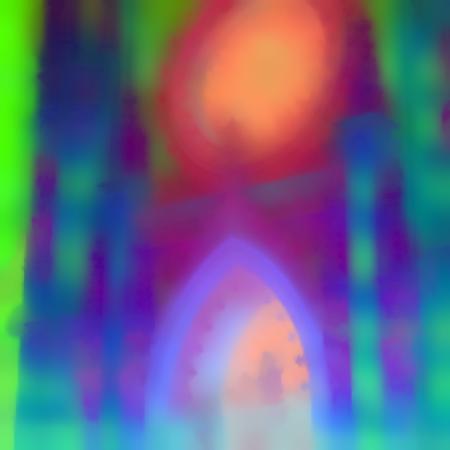}
         \label{fig:five over x}
     \end{subfigure}
     \hfill
     \begin{subfigure}[b]{0.12\textwidth}
         \centering
         \includegraphics[width=\textwidth]{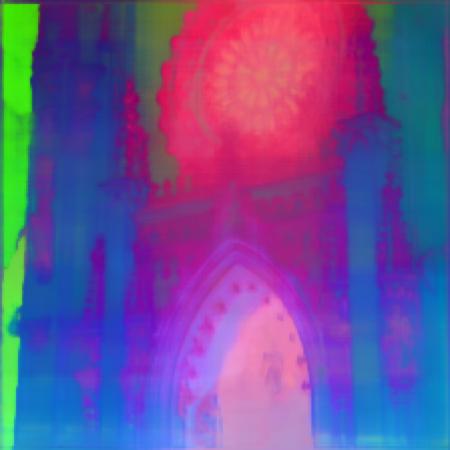}
         \label{fig:five over x}
     \end{subfigure} \\
     \vspace{-0.2cm}
      \begin{subfigure}[b]{0.12\textwidth}
         \centering
         \includegraphics[width=\textwidth]{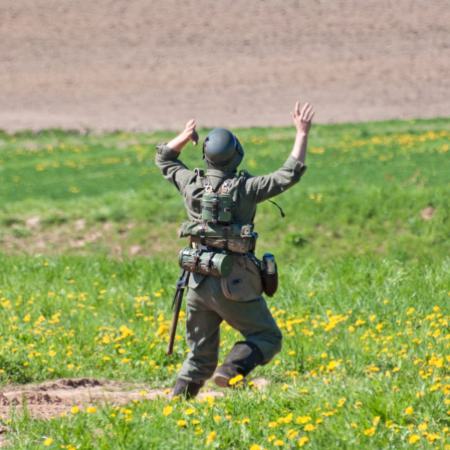}
         \label{fig:image}
     \end{subfigure}
     \hfill
     \begin{subfigure}[b]{0.12\textwidth}
         \centering
         \includegraphics[width=\textwidth]{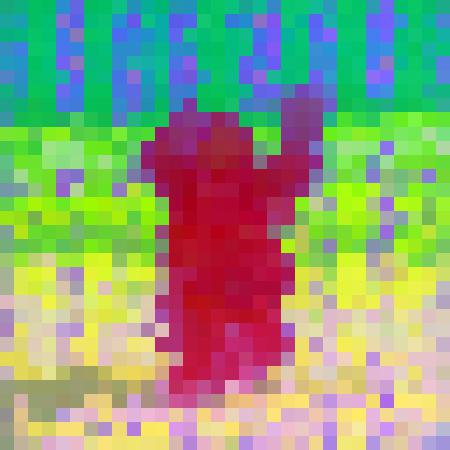}
         \label{fig:image}
     \end{subfigure}
     \hfill
     \begin{subfigure}[b]{0.12\textwidth}
         \centering
         \includegraphics[width=\textwidth]{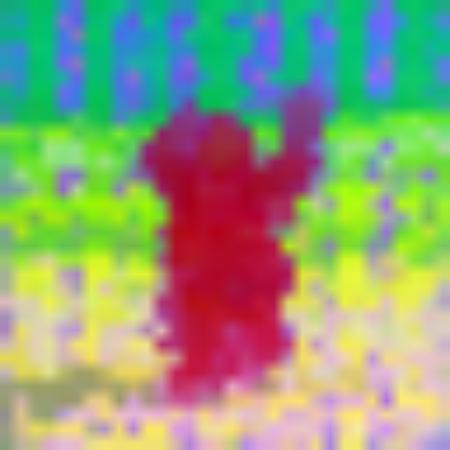}
         \label{fig:bilinear-openvoc}
     \end{subfigure}
     \hfill
     \begin{subfigure}[b]{0.12\textwidth}
         \centering
         \includegraphics[width=\textwidth]{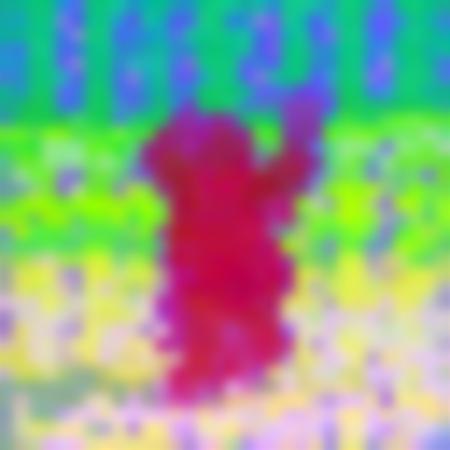}
         \label{fig:three sin x}
     \end{subfigure}
     \hfill
     \begin{subfigure}[b]{0.12\textwidth}
         \centering
         \includegraphics[width=\textwidth]{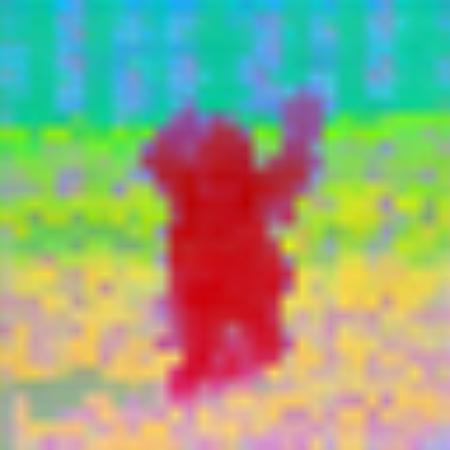}
         \label{fig:five over x}
     \end{subfigure}
     \hfill
     \begin{subfigure}[b]{0.12\textwidth}
         \centering
         \includegraphics[width=\textwidth]{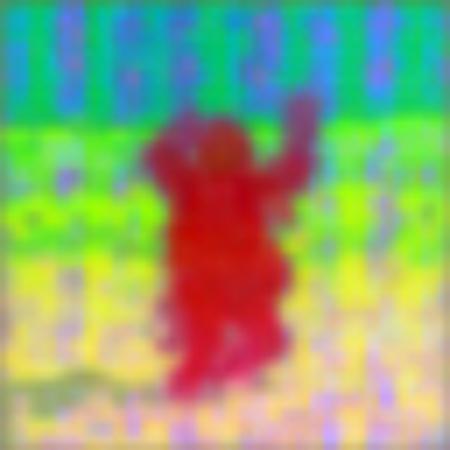}
         \label{fig:five over x}
     \end{subfigure}
     \hfill
     \begin{subfigure}[b]{0.12\textwidth}
         \centering
         \includegraphics[width=\textwidth]{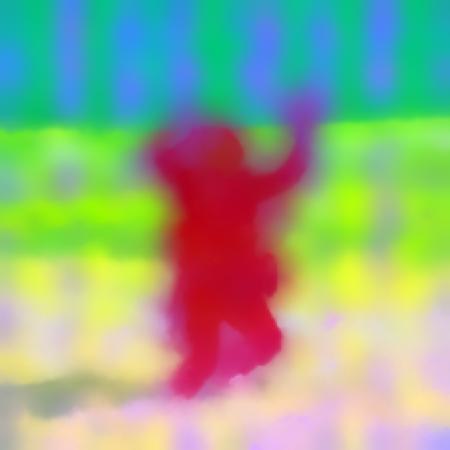}
         \label{fig:five over x}
     \end{subfigure}
     \hfill
     \begin{subfigure}[b]{0.12\textwidth}
         \centering
         \includegraphics[width=\textwidth]{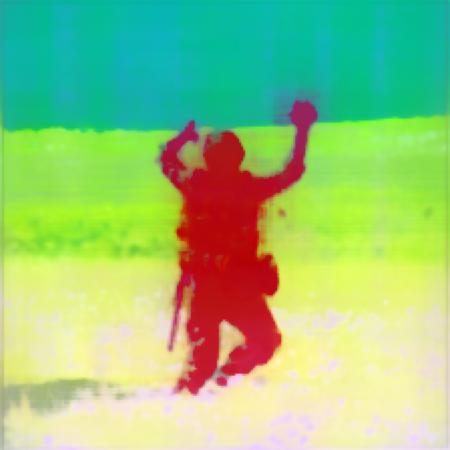}
         \label{fig:five over x}
     \end{subfigure} \\
     \vspace{-0.2cm}
      \begin{subfigure}[b]{0.12\textwidth}
         \centering
         \includegraphics[width=\textwidth]{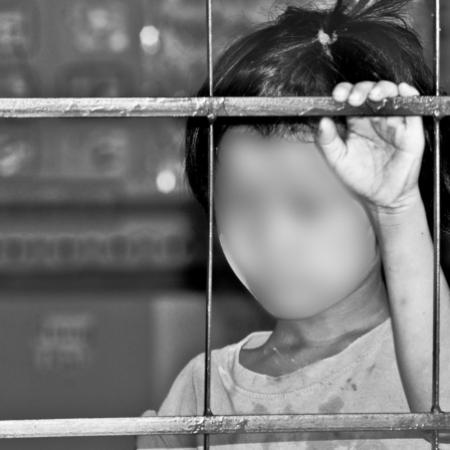}
         \label{fig:image}
     \end{subfigure}
     \hfill
     \begin{subfigure}[b]{0.12\textwidth}
         \centering
         \includegraphics[width=\textwidth]{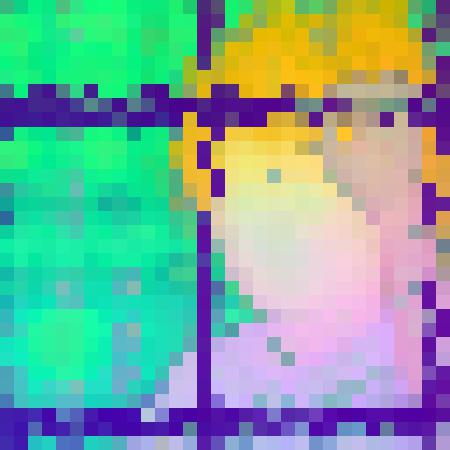}
         \label{fig:image}
     \end{subfigure}
     \hfill
     \begin{subfigure}[b]{0.12\textwidth}
         \centering
         \includegraphics[width=\textwidth]{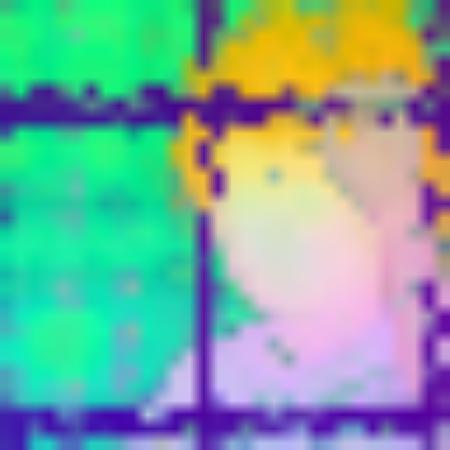}
         \label{fig:bilinear-openvoc}
     \end{subfigure}
     \hfill
     \begin{subfigure}[b]{0.12\textwidth}
         \centering
         \includegraphics[width=\textwidth]{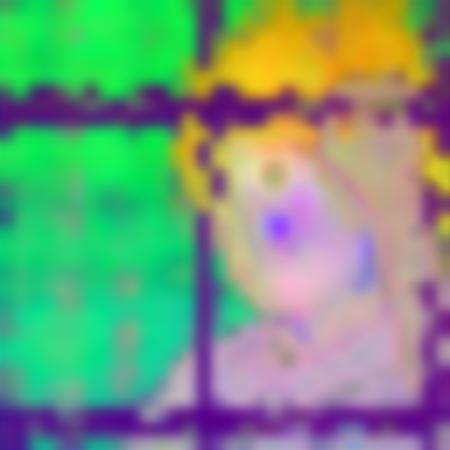}
         \label{fig:three sin x}
     \end{subfigure}
     \hfill
     \begin{subfigure}[b]{0.12\textwidth}
         \centering
         \includegraphics[width=\textwidth]{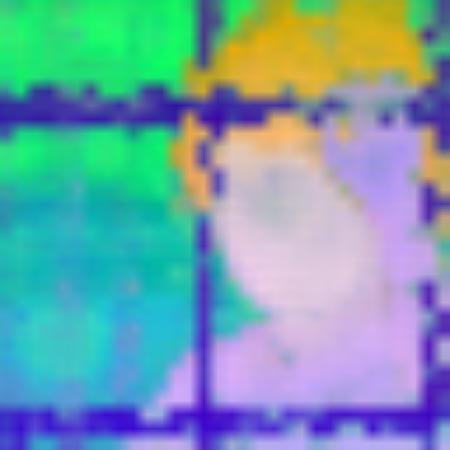}
         \label{fig:five over x}
     \end{subfigure}
     \hfill
     \begin{subfigure}[b]{0.12\textwidth}
         \centering
         \includegraphics[width=\textwidth]{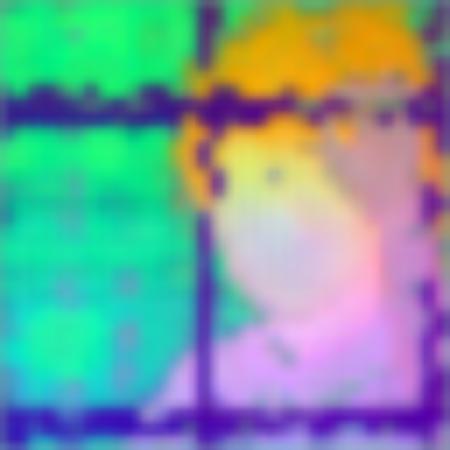}
         \label{fig:five over x}
     \end{subfigure}
     \hfill
     \begin{subfigure}[b]{0.12\textwidth}
         \centering
         \includegraphics[width=\textwidth]{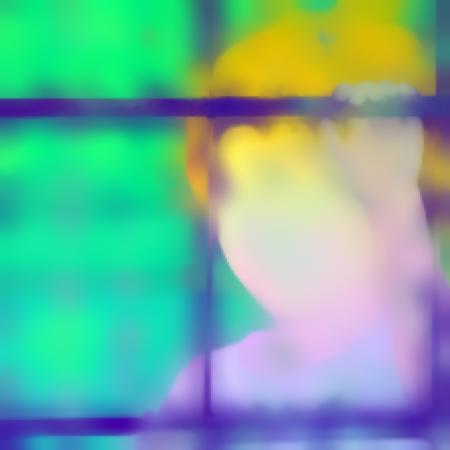}
         \label{fig:five over x}
     \end{subfigure}
     \hfill
     \begin{subfigure}[b]{0.12\textwidth}
         \centering
         \includegraphics[width=\textwidth]{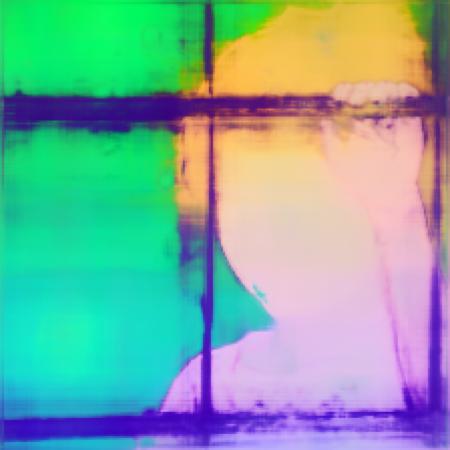}
         \label{fig:five over x}
     \end{subfigure} \\
       \vspace{-0.2cm}
      \begin{subfigure}[b]{0.12\textwidth}
         \centering
         \includegraphics[width=\textwidth]{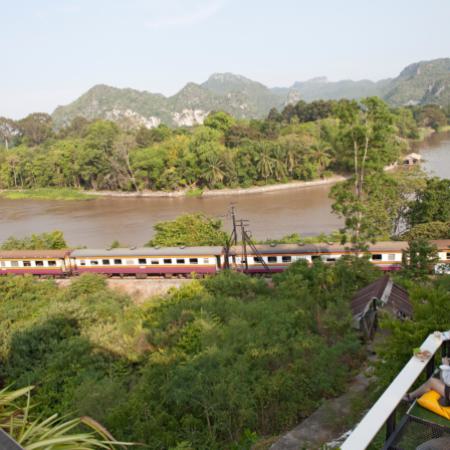}
         \label{fig:image}
     \end{subfigure}
     \hfill
     \begin{subfigure}[b]{0.12\textwidth}
         \centering
         \includegraphics[width=\textwidth]{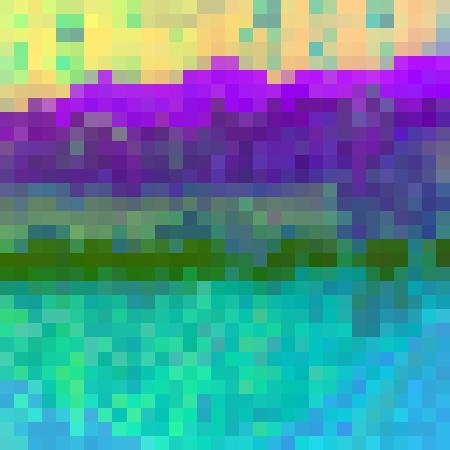}
         \label{fig:image}
     \end{subfigure}
     \hfill
     \begin{subfigure}[b]{0.12\textwidth}
         \centering
         \includegraphics[width=\textwidth]{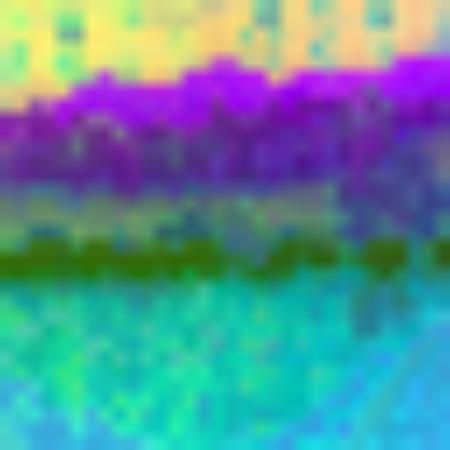}
         \label{fig:bilinear-openvoc}
     \end{subfigure}
     \hfill
     \begin{subfigure}[b]{0.12\textwidth}
         \centering
         \includegraphics[width=\textwidth]{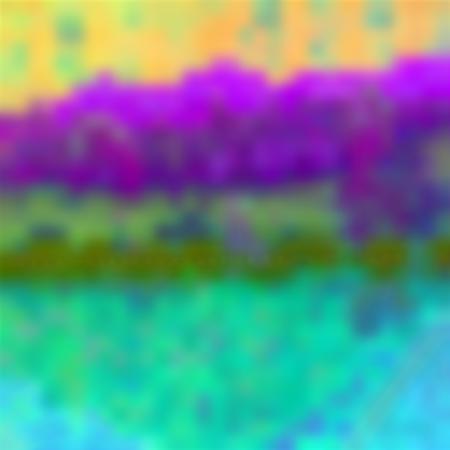}
         \label{fig:three sin x}
     \end{subfigure}
     \hfill
     \begin{subfigure}[b]{0.12\textwidth}
         \centering
         \includegraphics[width=\textwidth]{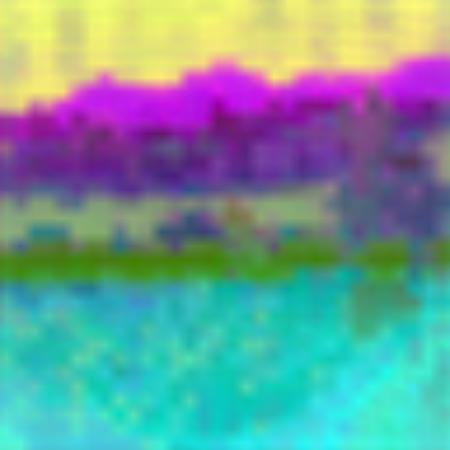}
         \label{fig:five over x}
     \end{subfigure}
     \hfill
     \begin{subfigure}[b]{0.12\textwidth}
         \centering
         \includegraphics[width=\textwidth]{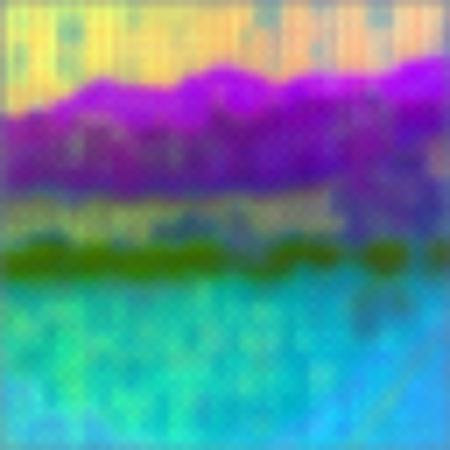}
         \label{fig:five over x}
     \end{subfigure}
     \hfill
     \begin{subfigure}[b]{0.12\textwidth}
         \centering
         \includegraphics[width=\textwidth]{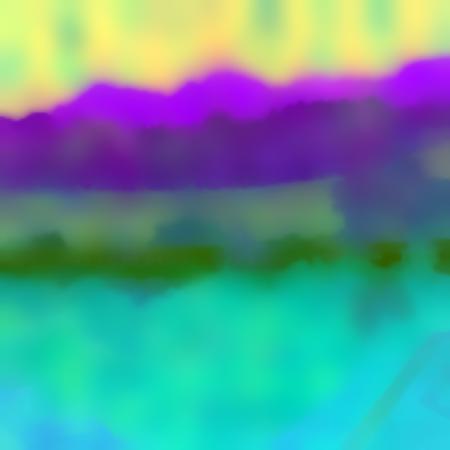}
         \label{fig:five over x}
     \end{subfigure}
     \hfill
     \begin{subfigure}[b]{0.12\textwidth}
         \centering
         \includegraphics[width=\textwidth]{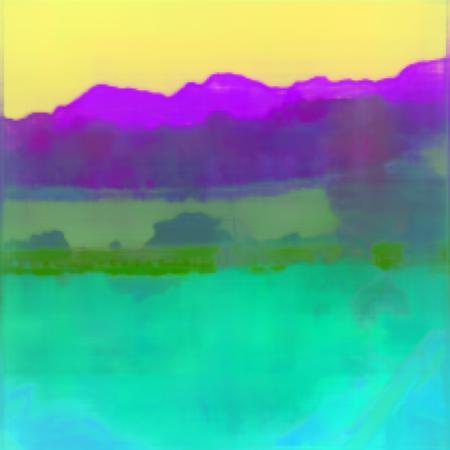}
         \label{fig:five over x}
     \end{subfigure} \\
      \vspace{-0.2cm}
      \begin{subfigure}[b]{0.12\textwidth}
         \centering
         \includegraphics[width=\textwidth]{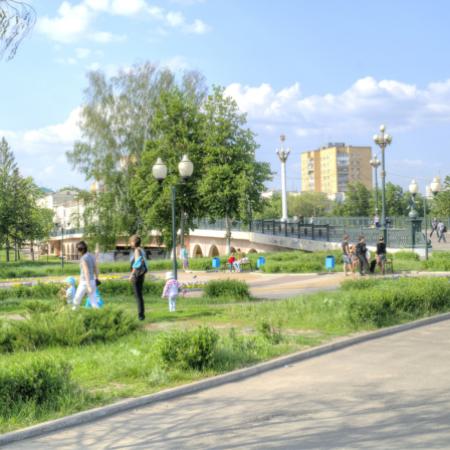}
         \label{fig:image}
     \end{subfigure}
     \hfill
     \begin{subfigure}[b]{0.12\textwidth}
         \centering
         \includegraphics[width=\textwidth]{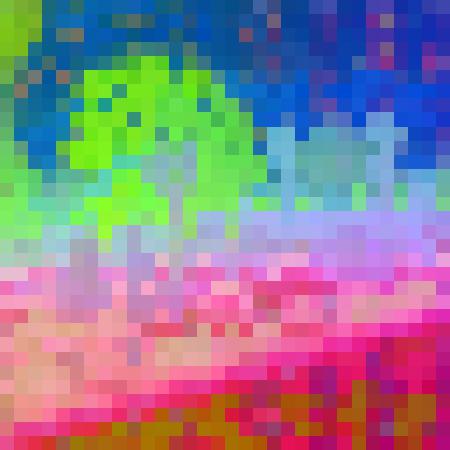}
         \label{fig:image}
     \end{subfigure}
     \hfill
     \begin{subfigure}[b]{0.12\textwidth}
         \centering
         \includegraphics[width=\textwidth]{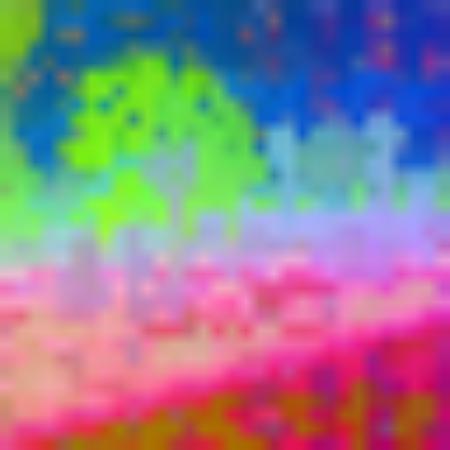}
         \label{fig:bilinear-openvoc}
     \end{subfigure}
     \hfill
     \begin{subfigure}[b]{0.12\textwidth}
         \centering
         \includegraphics[width=\textwidth]{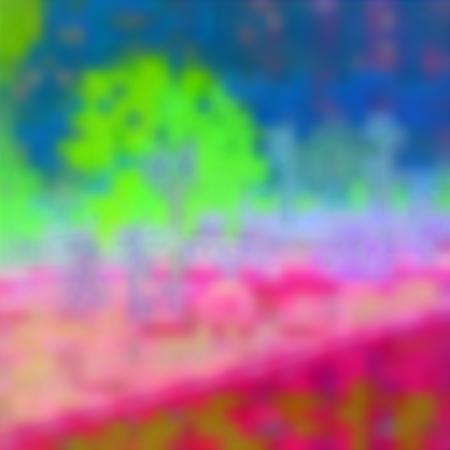}
         \label{fig:three sin x}
     \end{subfigure}
     \hfill
     \begin{subfigure}[b]{0.12\textwidth}
         \centering
         \includegraphics[width=\textwidth]{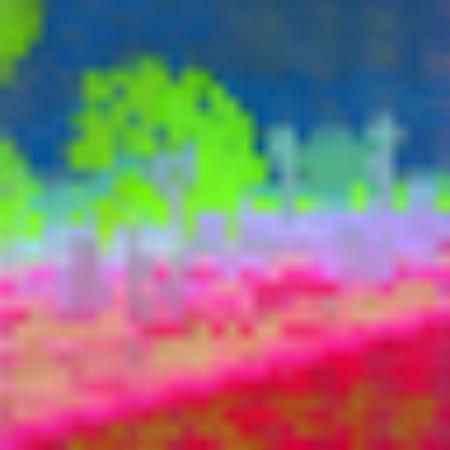}
         \label{fig:five over x}
     \end{subfigure}
     \hfill
     \begin{subfigure}[b]{0.12\textwidth}
         \centering
         \includegraphics[width=\textwidth]{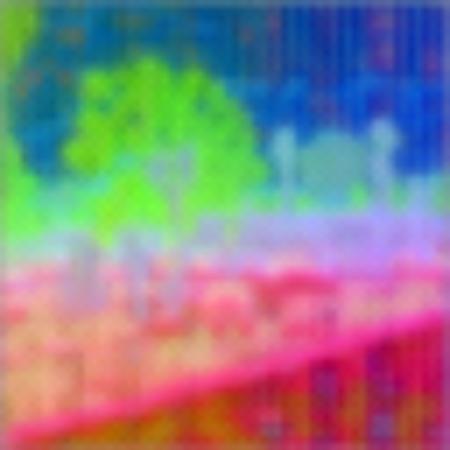}
         \label{fig:five over x}
     \end{subfigure}
     \hfill
     \begin{subfigure}[b]{0.12\textwidth}
         \centering
         \includegraphics[width=\textwidth]{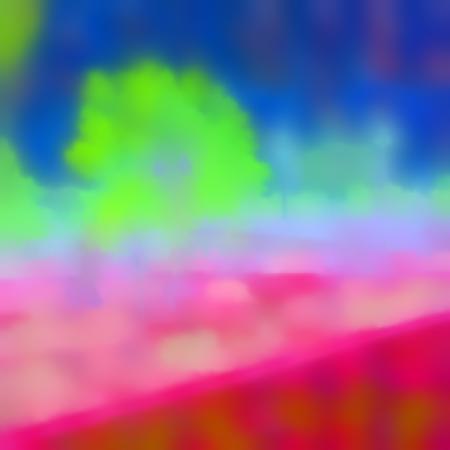}
         \label{fig:five over x}
     \end{subfigure}
     \hfill
     \begin{subfigure}[b]{0.12\textwidth}
         \centering
         \includegraphics[width=\textwidth]{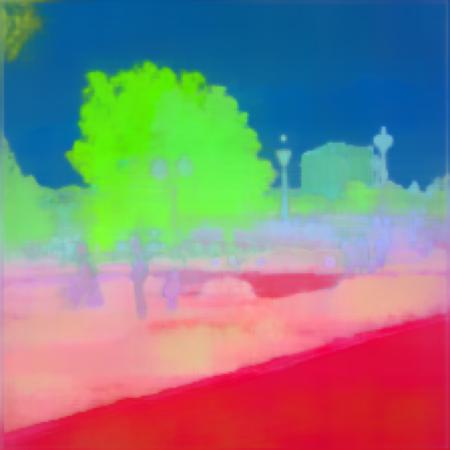}
         \label{fig:five over x}
     \end{subfigure} \\
          \vspace{-0.2cm}
      \begin{subfigure}[b]{0.12\textwidth}
         \centering
         \includegraphics[width=\textwidth]{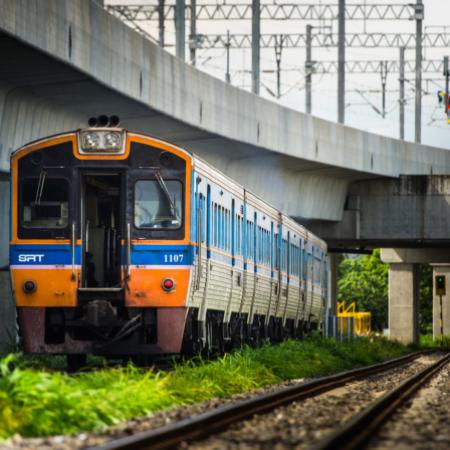}
         \label{fig:image}
     \end{subfigure}
     \hfill
     \begin{subfigure}[b]{0.12\textwidth}
         \centering
         \includegraphics[width=\textwidth]{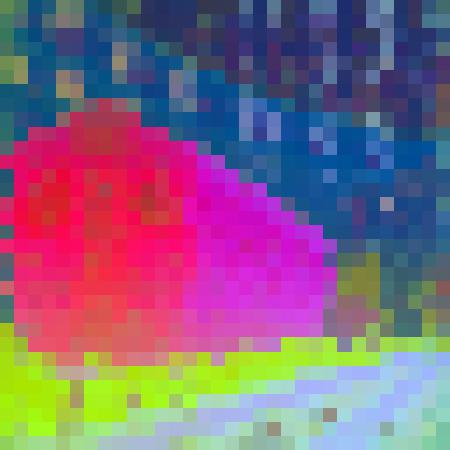}
         \label{fig:image}
     \end{subfigure}
     \hfill
     \begin{subfigure}[b]{0.12\textwidth}
         \centering
         \includegraphics[width=\textwidth]{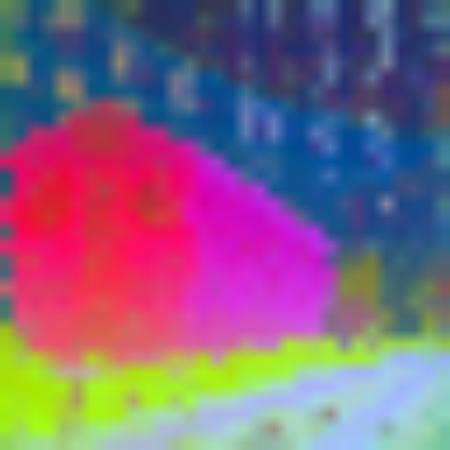}
         \label{fig:bilinear-openvoc}
     \end{subfigure}
     \hfill
     \begin{subfigure}[b]{0.12\textwidth}
         \centering
         \includegraphics[width=\textwidth]{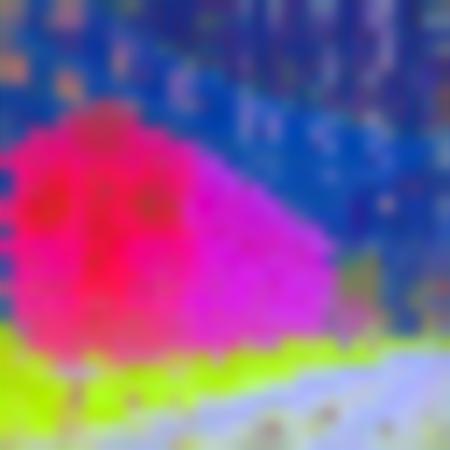}
         \label{fig:three sin x}
     \end{subfigure}
     \hfill
     \begin{subfigure}[b]{0.12\textwidth}
         \centering
         \includegraphics[width=\textwidth]{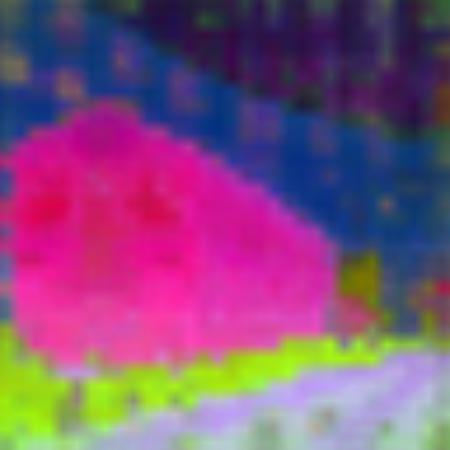}
         \label{fig:five over x}
     \end{subfigure}
     \hfill
     \begin{subfigure}[b]{0.12\textwidth}
         \centering
         \includegraphics[width=\textwidth]{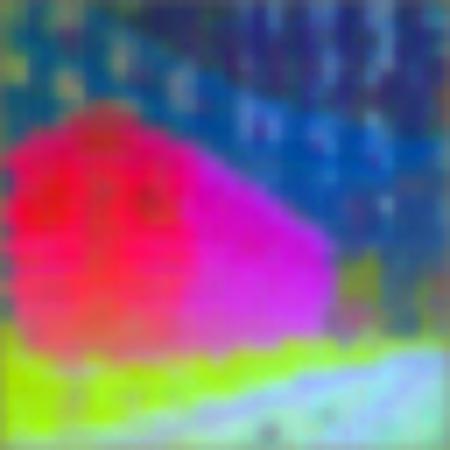}
         \label{fig:five over x}
     \end{subfigure}
     \hfill
     \begin{subfigure}[b]{0.12\textwidth}
         \centering
         \includegraphics[width=\textwidth]{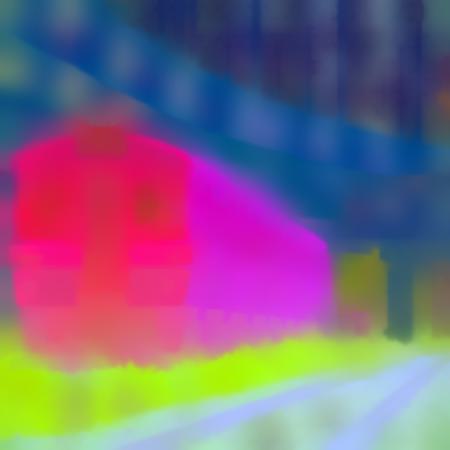}
         \label{fig:five over x}
     \end{subfigure}
     \hfill
     \begin{subfigure}[b]{0.12\textwidth}
         \centering
         \includegraphics[width=\textwidth]{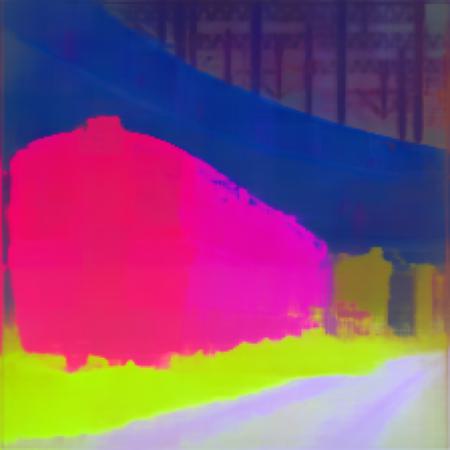}
         \label{fig:five over x}
     \end{subfigure} \\
     \vspace{-0.2cm}
         \begin{subfigure}[b]{0.12\textwidth}
         \centering
         \includegraphics[width=\textwidth]{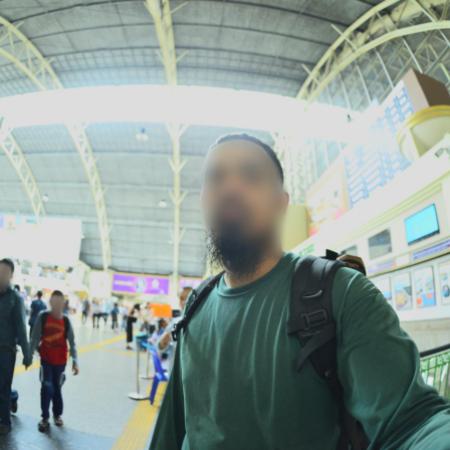}
         \caption{Original image}
         \label{fig:image}
     \end{subfigure}
     \hfill
     \begin{subfigure}[b]{0.12\textwidth}
         \centering
         \includegraphics[width=\textwidth]{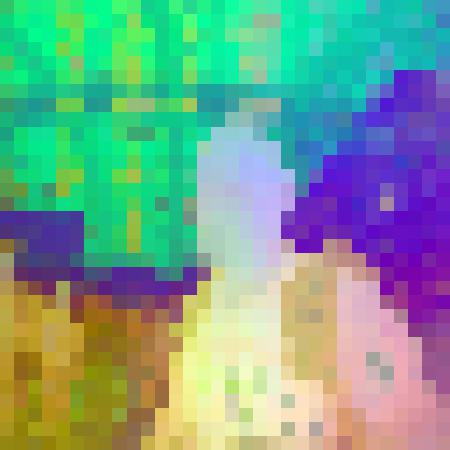}
         \caption{Low res}
         \label{fig:image}
     \end{subfigure}
     \hfill
     \begin{subfigure}[b]{0.12\textwidth}
         \centering
         \includegraphics[width=\textwidth]{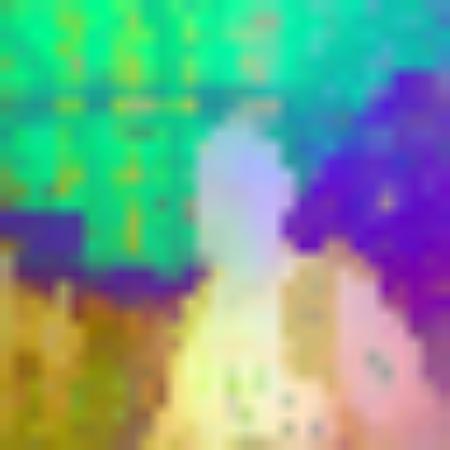}
         \caption{Bilinear}
         \label{fig:bilinear-openvoc}
     \end{subfigure}
     \hfill
     \begin{subfigure}[b]{0.12\textwidth}
         \centering
         \includegraphics[width=\textwidth]{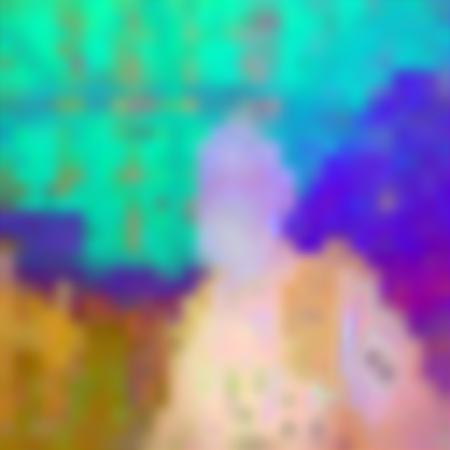}
         \caption{Resize-conv}
         \label{fig:three sin x}
     \end{subfigure}
     \hfill
     \begin{subfigure}[b]{0.12\textwidth}
         \centering
         \includegraphics[width=\textwidth]{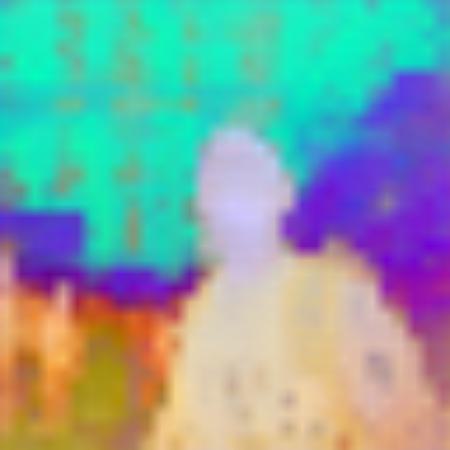}
         \caption{LIIF}
         \label{fig:five over x}
     \end{subfigure}
     \hfill
     \begin{subfigure}[b]{0.12\textwidth}
         \centering
         \includegraphics[width=\textwidth]{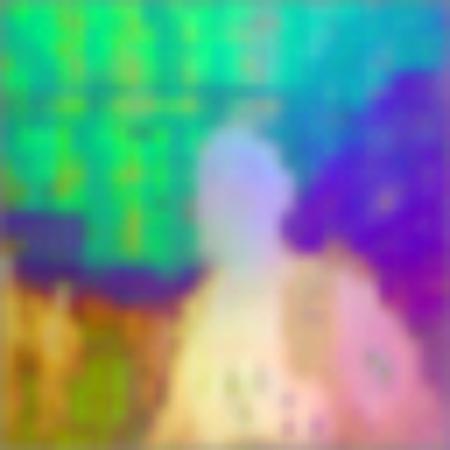}
         \caption{LiFT}
         \label{fig:five over x}
     \end{subfigure}
     \hfill
     \begin{subfigure}[b]{0.12\textwidth}
         \centering
         \includegraphics[width=\textwidth]{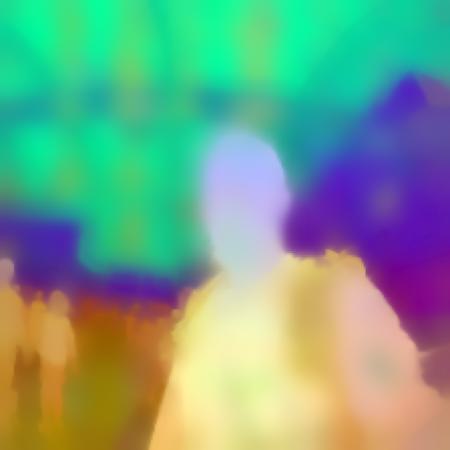}
         \caption{FeatUp}
         \label{fig:five over x}
     \end{subfigure}
     \hfill
     \begin{subfigure}[b]{0.12\textwidth}
         \centering
         \includegraphics[width=\textwidth]{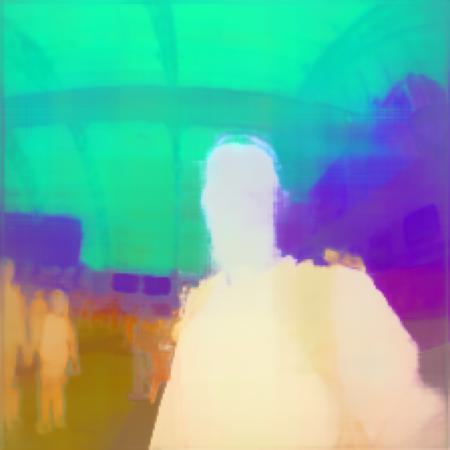}
         \caption{\textbf{LoftUp(Ours)}}
         \label{fig:five over x}
     \end{subfigure} \\
    \caption{\textbf{More visualization of features from various upsamplers.} Backbone is DINOv2-S/14~\cite{oquab2024dinov}.}
    \label{fig:more-vis-feats}
\end{figure*}

\begin{figure*}[t]
    \centering

    \begin{subfigure}[b]{0.16\textwidth}
         \centering
         \includegraphics[width=\textwidth]{sec/figs/vis_semseg_new/imgs/000000007795.jpg}
         \label{fig:image}
     \end{subfigure}
     \hfill
     \begin{subfigure}[b]{0.16\textwidth}
         \centering
         \includegraphics[width=\textwidth]{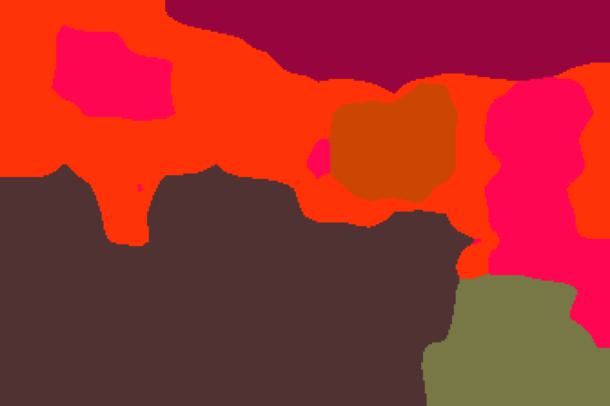}
         \label{fig:bilinear-openvoc}
     \end{subfigure}
     \hfill
     \begin{subfigure}[b]{0.16\textwidth}
         \centering
         \includegraphics[width=\textwidth]{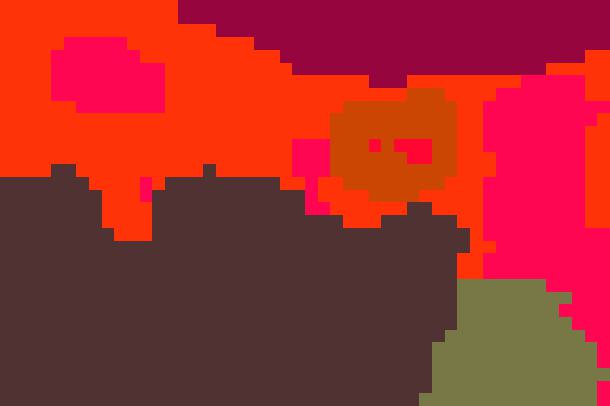}
         \label{fig:five over x}
     \end{subfigure}
     \hfill
     \begin{subfigure}[b]{0.16\textwidth}
         \centering
         \includegraphics[width=\textwidth]{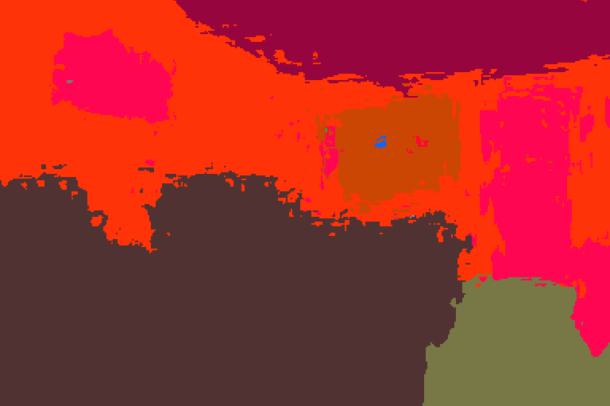}
         \label{fig:three sin x}
     \end{subfigure}
     \hfill
     \begin{subfigure}[b]{0.16\textwidth}
         \centering
         \includegraphics[width=\textwidth]{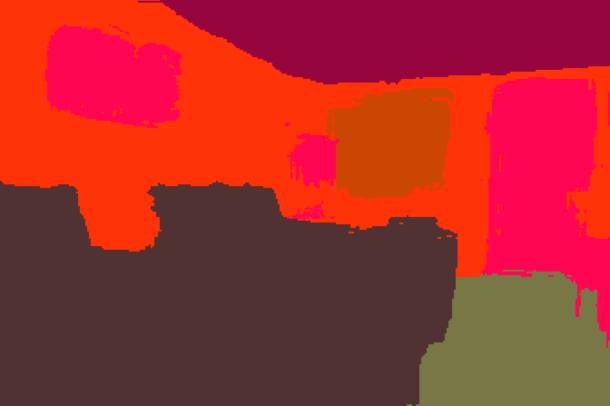}
         \label{fig:five over x}
     \end{subfigure}
     \hfill
     \begin{subfigure}[b]{0.16\textwidth}
         \centering
         \includegraphics[width=\textwidth]{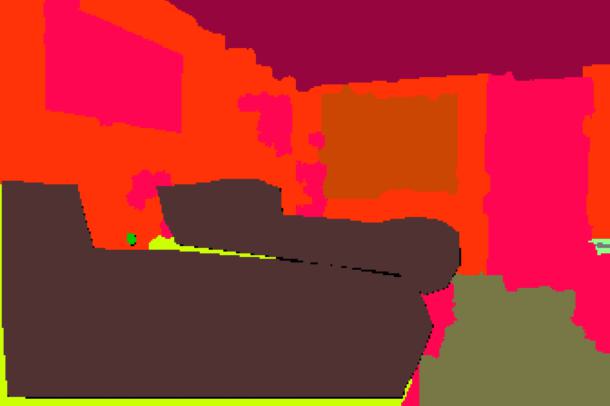}
         \label{fig:five over x}
     \end{subfigure} \\
      \vspace{-0.2cm}
    \begin{subfigure}[b]{0.16\textwidth}
         \centering
         \includegraphics[width=\textwidth]{sec/figs/vis_semseg_new/imgs/000000007511.jpg}
         \label{fig:image}
     \end{subfigure}
     \hfill
     \begin{subfigure}[b]{0.16\textwidth}
         \centering
         \includegraphics[width=\textwidth]{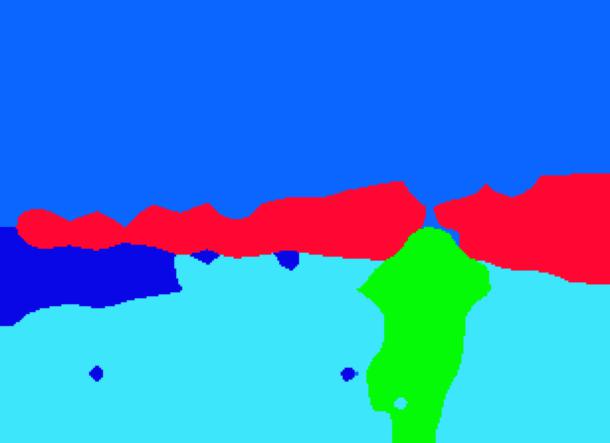}
         \label{fig:bilinear-openvoc}
     \end{subfigure}
     \hfill
     \begin{subfigure}[b]{0.16\textwidth}
         \centering
         \includegraphics[width=\textwidth]{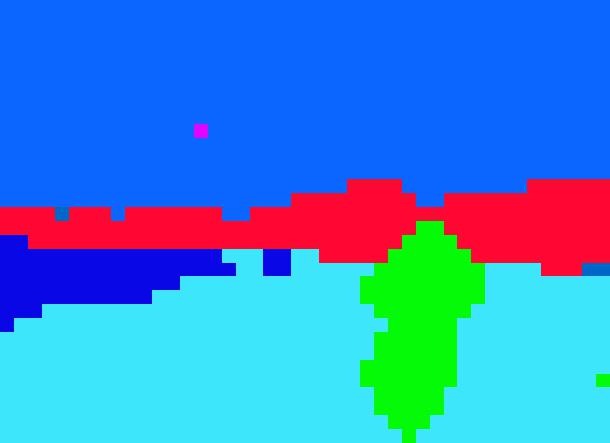}
         \label{fig:five over x}
     \end{subfigure}
     \hfill
     \begin{subfigure}[b]{0.16\textwidth}
         \centering
         \includegraphics[width=\textwidth]{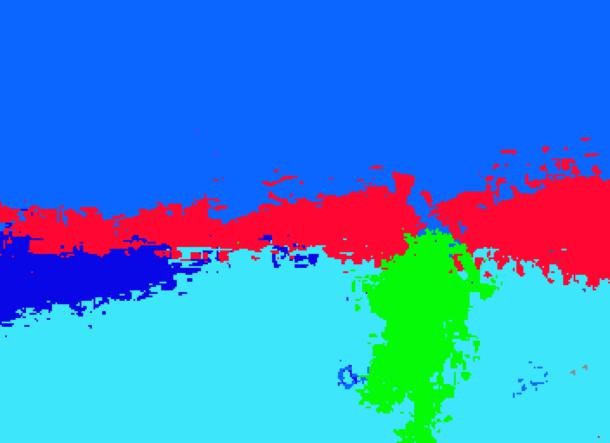}
         \label{fig:three sin x}
     \end{subfigure}
     \hfill
     \begin{subfigure}[b]{0.16\textwidth}
         \centering
         \includegraphics[width=\textwidth]{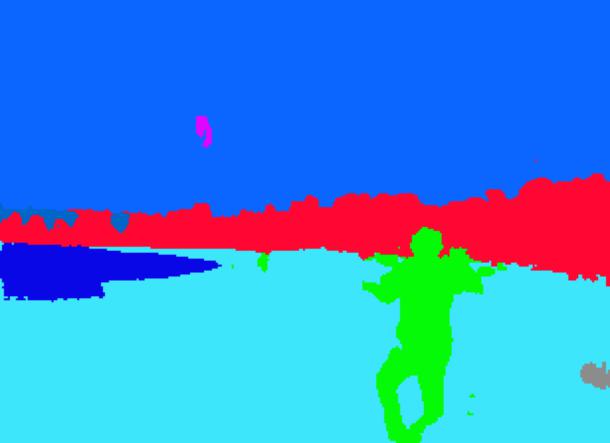}
         \label{fig:five over x}
     \end{subfigure}
     \hfill
     \begin{subfigure}[b]{0.16\textwidth}
         \centering
         \includegraphics[width=\textwidth]{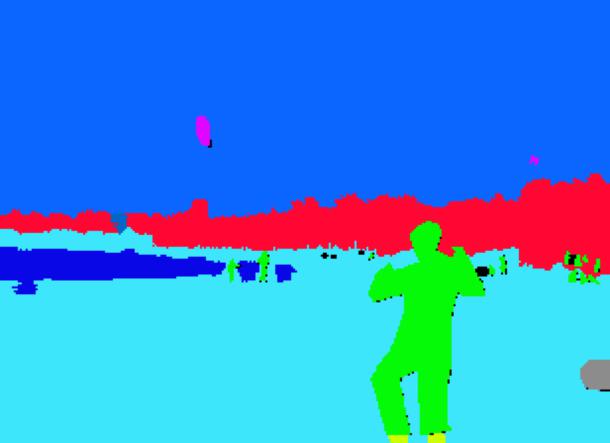}
         \label{fig:five over x}
     \end{subfigure} \\ 
      \vspace{-0.2cm}
    \begin{subfigure}[b]{0.16\textwidth}
         \centering
         \includegraphics[width=\textwidth]{sec/figs/vis_semseg_new/imgs/000000005992.jpg}
         \label{fig:image}
     \end{subfigure}
     \hfill
     \begin{subfigure}[b]{0.16\textwidth}
         \centering
         \includegraphics[width=\textwidth]{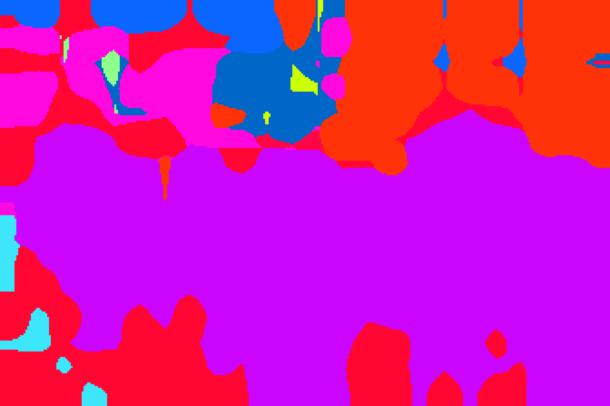}
         \label{fig:bilinear-openvoc}
     \end{subfigure}
     \hfill
     \begin{subfigure}[b]{0.16\textwidth}
         \centering
         \includegraphics[width=\textwidth]{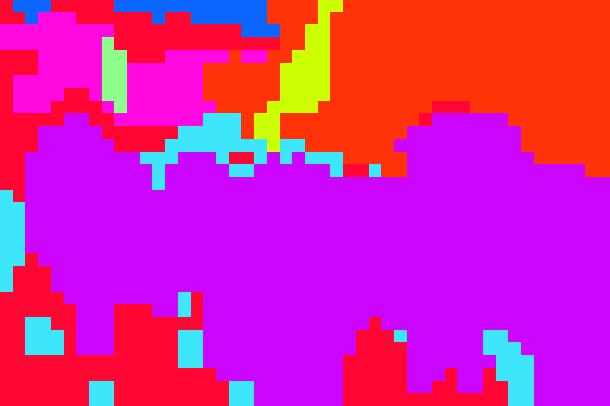}
         \label{fig:five over x}
     \end{subfigure}
     \hfill
     \begin{subfigure}[b]{0.16\textwidth}
         \centering
         \includegraphics[width=\textwidth]{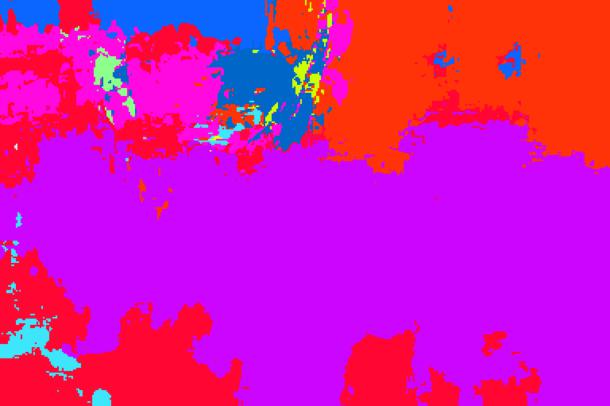}
         \label{fig:three sin x}
     \end{subfigure}
     \hfill
     \begin{subfigure}[b]{0.16\textwidth}
         \centering
         \includegraphics[width=\textwidth]{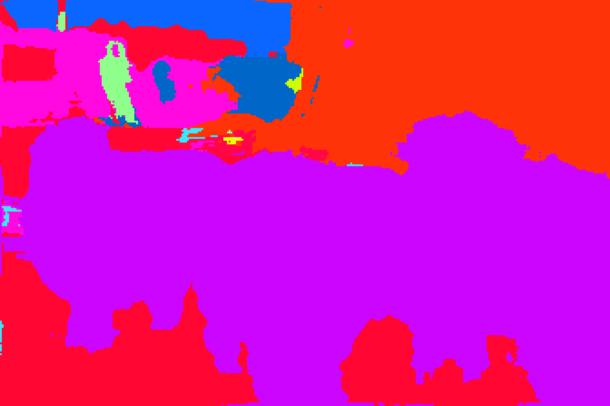}
         \label{fig:five over x}
     \end{subfigure}
     \hfill
     \begin{subfigure}[b]{0.16\textwidth}
         \centering
         \includegraphics[width=\textwidth]{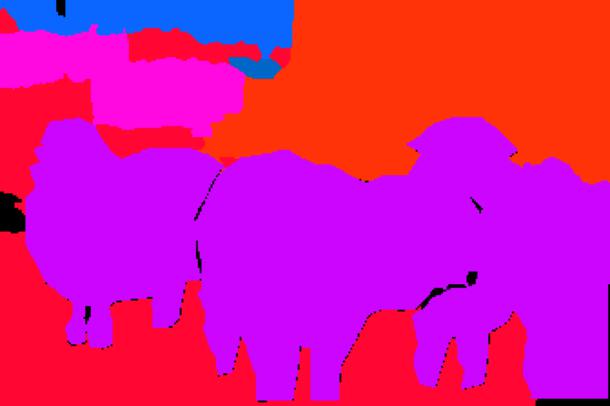}
         \label{fig:five over x}
     \end{subfigure}  \\
     \vspace{-0.2cm}
     \begin{subfigure}[b]{0.16\textwidth}
         \centering
         \includegraphics[width=\textwidth]{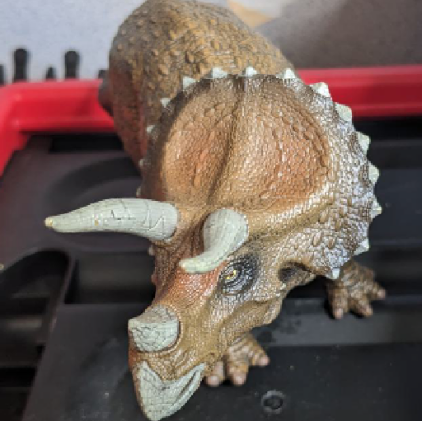}
         \label{fig:image}
     \end{subfigure}
     \hfill
     \begin{subfigure}[b]{0.16\textwidth}
         \centering
         \includegraphics[width=\textwidth]{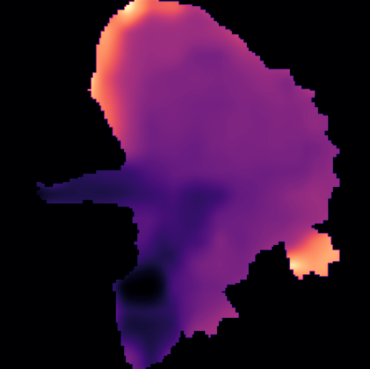}
         \label{fig:bilinear-openvoc}
     \end{subfigure}
     \hfill
     \begin{subfigure}[b]{0.16\textwidth}
         \centering
         \includegraphics[width=\textwidth]{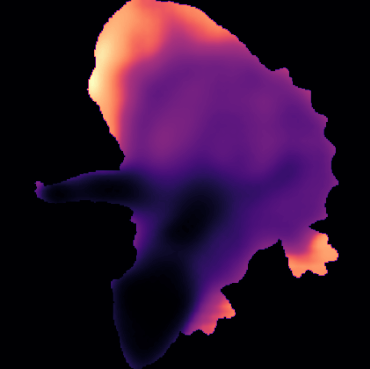}
         \label{fig:five over x}
     \end{subfigure}
     \hfill
     \begin{subfigure}[b]{0.16\textwidth}
         \centering
         \includegraphics[width=\textwidth]{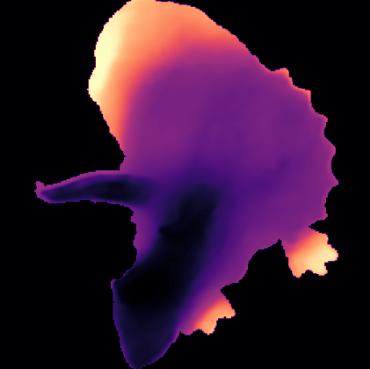}
         \label{fig:three sin x}
     \end{subfigure}
     \hfill
     \begin{subfigure}[b]{0.16\textwidth}
         \centering
         \includegraphics[width=\textwidth]{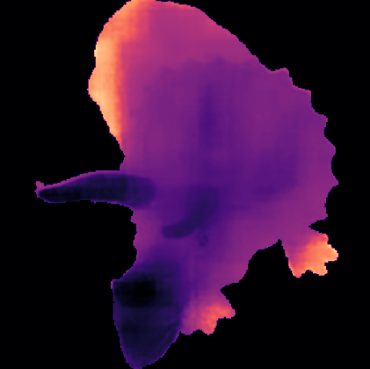}
         \label{fig:five over x}
     \end{subfigure}
     \hfill
     \begin{subfigure}[b]{0.16\textwidth}
         \centering
         \includegraphics[width=\textwidth]{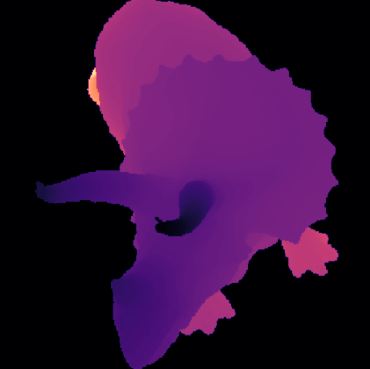}
         \label{fig:five over x}
     \end{subfigure} \\
     \vspace{-0.2cm}
     \begin{subfigure}[b]{0.16\textwidth}
         \centering
         \includegraphics[width=\textwidth]{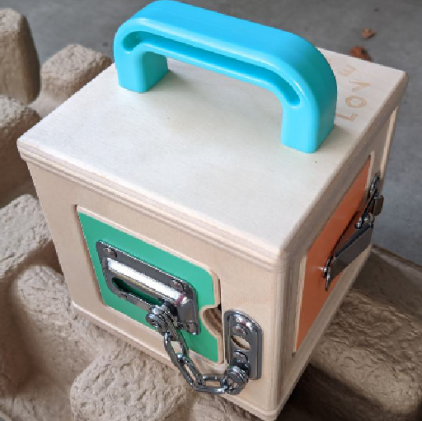}
         \label{fig:image}
     \end{subfigure}
     \hfill
     \begin{subfigure}[b]{0.16\textwidth}
         \centering
         \includegraphics[width=\textwidth]{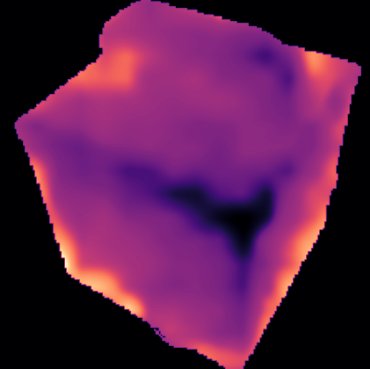}
         \label{fig:bilinear-openvoc}
     \end{subfigure}
     \hfill
     \begin{subfigure}[b]{0.16\textwidth}
         \centering
         \includegraphics[width=\textwidth]{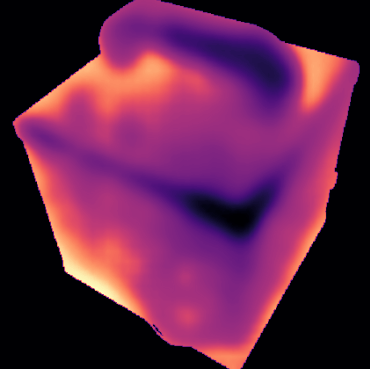}
         \label{fig:five over x}
     \end{subfigure}
     \hfill
     \begin{subfigure}[b]{0.16\textwidth}
         \centering
         \includegraphics[width=\textwidth]{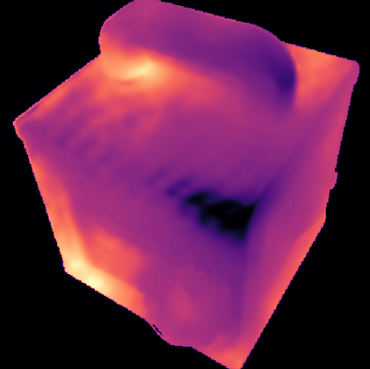}
         \label{fig:three sin x}
     \end{subfigure}
     \hfill
     \begin{subfigure}[b]{0.16\textwidth}
         \centering
         \includegraphics[width=\textwidth]{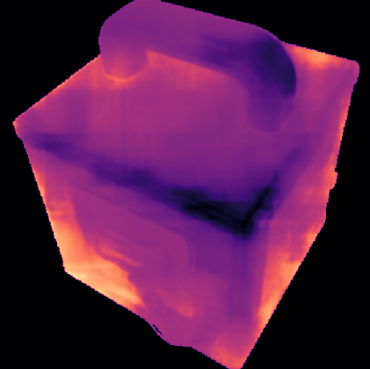}
         \label{fig:five over x}
     \end{subfigure}
     \hfill
     \begin{subfigure}[b]{0.16\textwidth}
         \centering
         \includegraphics[width=\textwidth]{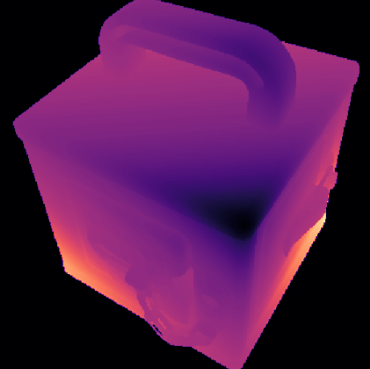}
         \label{fig:five over x}
     \end{subfigure} \\
     \vspace{-0.2cm}
     \begin{subfigure}[b]{0.16\textwidth}
         \centering
         \includegraphics[width=\textwidth]{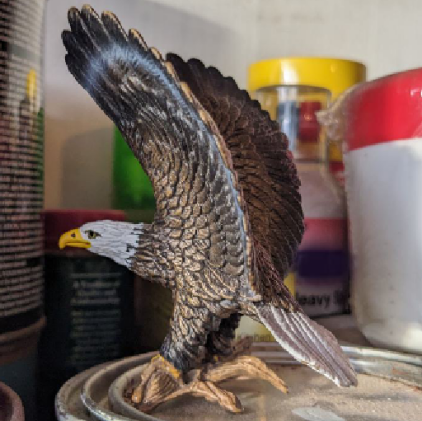}
         \caption{Original image}
         \label{fig:image}
     \end{subfigure}
     \hfill
     \begin{subfigure}[b]{0.16\textwidth}
         \centering
         \includegraphics[width=\textwidth]{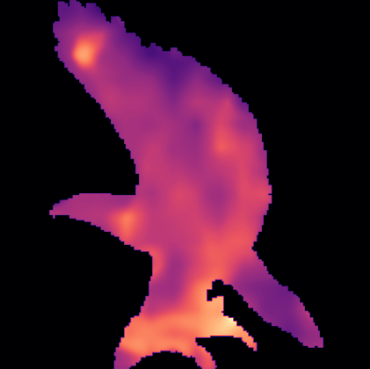}
         \caption{Bilinear}
         \label{fig:bilinear-openvoc}
     \end{subfigure}
     \hfill
     \begin{subfigure}[b]{0.16\textwidth}
         \centering
         \includegraphics[width=\textwidth]{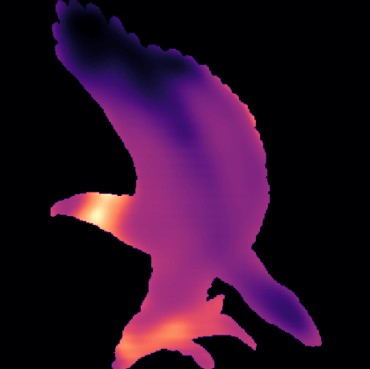}
         \caption{LiFT}
         \label{fig:five over x}
     \end{subfigure}
     \hfill
     \begin{subfigure}[b]{0.16\textwidth}
         \centering
         \includegraphics[width=\textwidth]{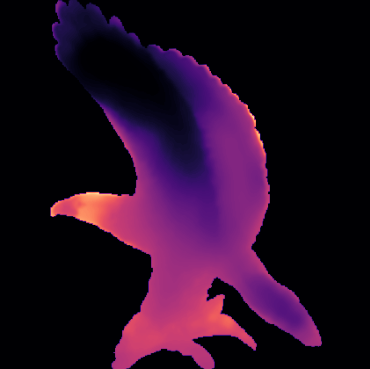}
         \caption{FeatUp}
         \label{fig:three sin x}
     \end{subfigure}
     \hfill
     \begin{subfigure}[b]{0.16\textwidth}
         \centering
         \includegraphics[width=\textwidth]{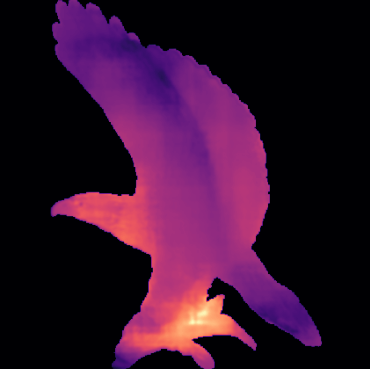}
         \caption{\textbf{LoftUp}}
         \label{fig:five over x}
     \end{subfigure}
     \hfill
     \begin{subfigure}[b]{0.16\textwidth}
         \centering
         \includegraphics[width=\textwidth]{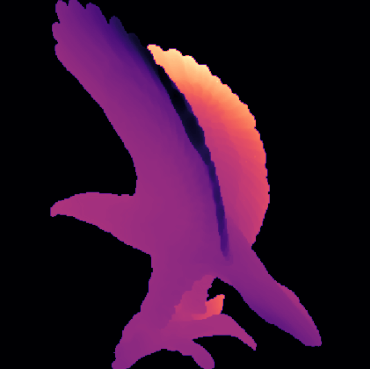}
         \caption{Groundtruth}
         \label{fig:five over x}
     \end{subfigure}
    \caption{\textbf{More visualization of predictions examples} on semantic segmentation on COCO-Stuff~\cite{caesar2018coco} and depth estimation on NAVI~\cite{jampani2023navi}.}
    \label{fig:more-vis-preds}
\end{figure*}

\begin{figure*}[t]
    \centering

\begin{subfigure}[b]{0.16\textwidth}
         \centering
         \includegraphics[width=\textwidth]{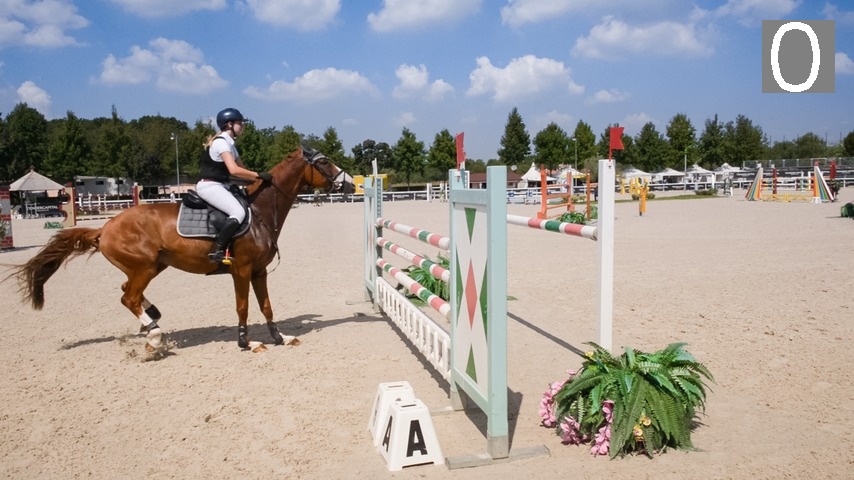}
         \label{fig:image}
     \end{subfigure}
     \hfill
     \begin{subfigure}[b]{0.16\textwidth}
         \centering
         \includegraphics[width=\textwidth]{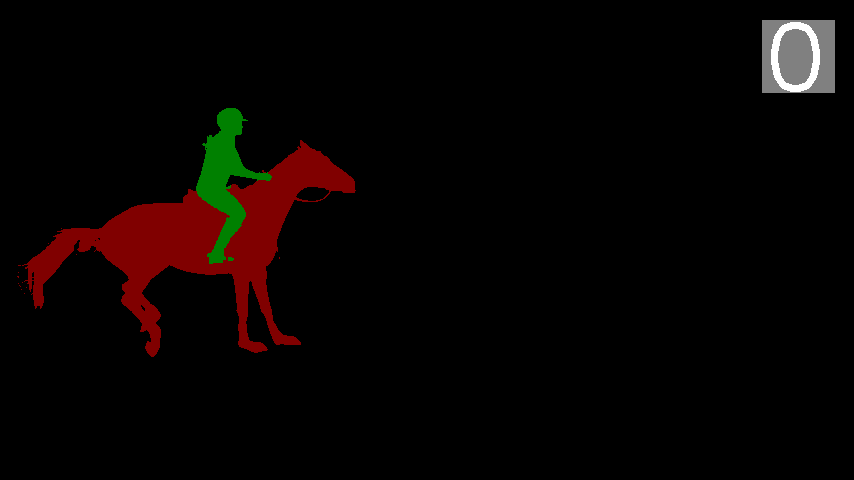}
         \label{fig:bilinear-openvoc}
     \end{subfigure}
     \hfill
     \begin{subfigure}[b]{0.16\textwidth}
         \centering
         \includegraphics[width=\textwidth]{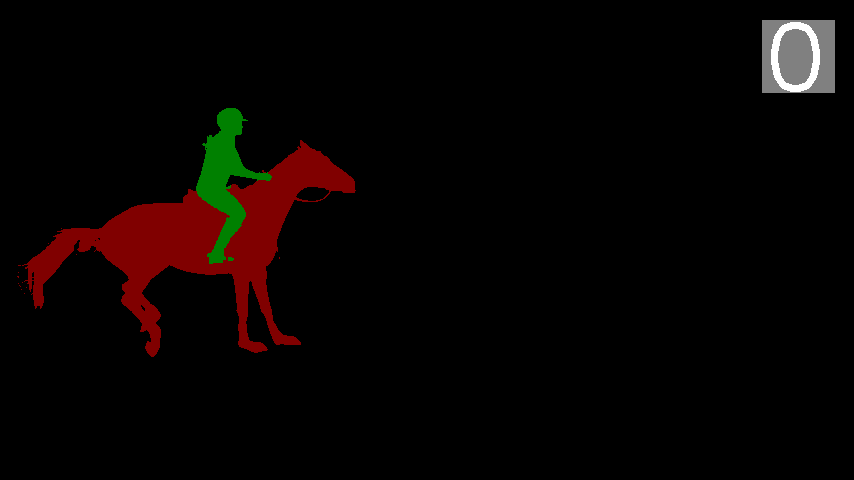}
         \label{fig:five over x}
     \end{subfigure}
     \hfill
     \begin{subfigure}[b]{0.16\textwidth}
         \centering
         \includegraphics[width=\textwidth]{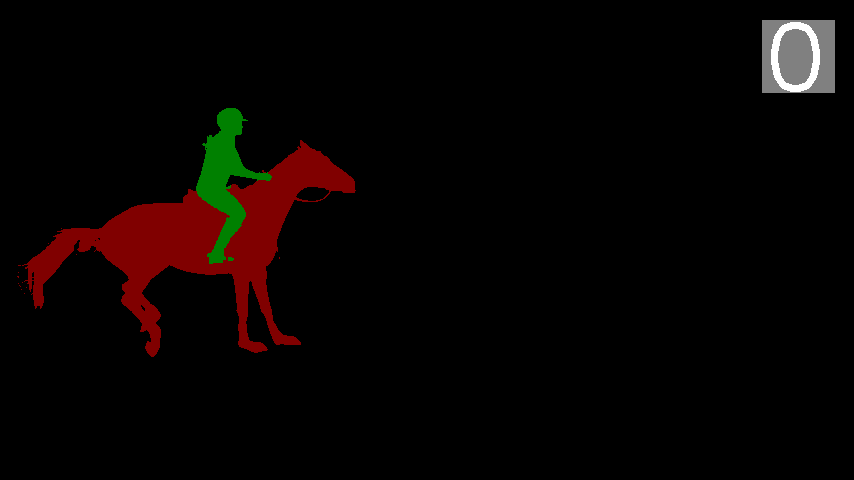}
         \label{fig:three sin x}
     \end{subfigure}
     \hfill
     \begin{subfigure}[b]{0.16\textwidth}
         \centering
         \includegraphics[width=\textwidth]{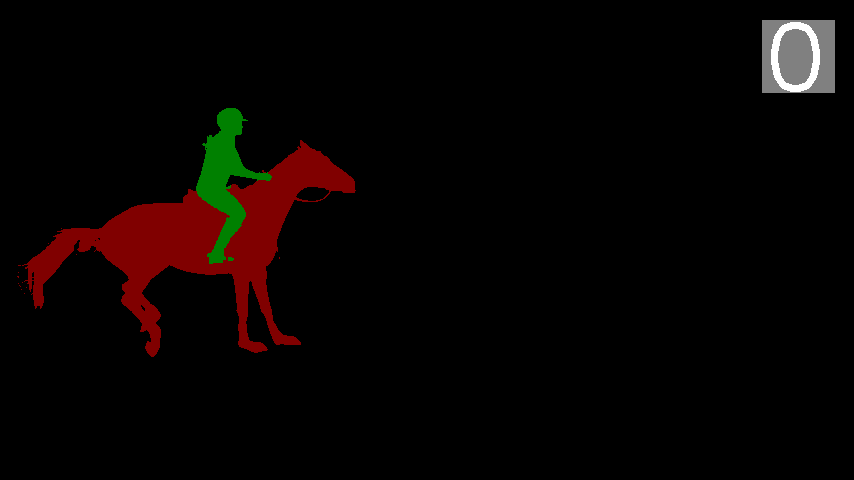}
         \label{fig:five over x}
     \end{subfigure}
     \hfill
     \begin{subfigure}[b]{0.16\textwidth}
         \centering
         \includegraphics[width=\textwidth]{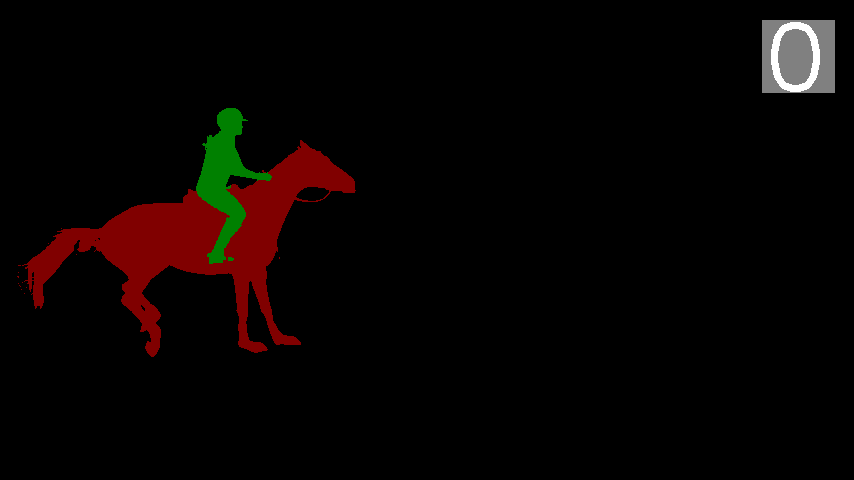}
         \label{fig:five over x}
     \end{subfigure}  \\
     \vspace{-0.2cm}
     \begin{subfigure}[b]{0.16\textwidth}
         \centering
         \includegraphics[width=\textwidth]{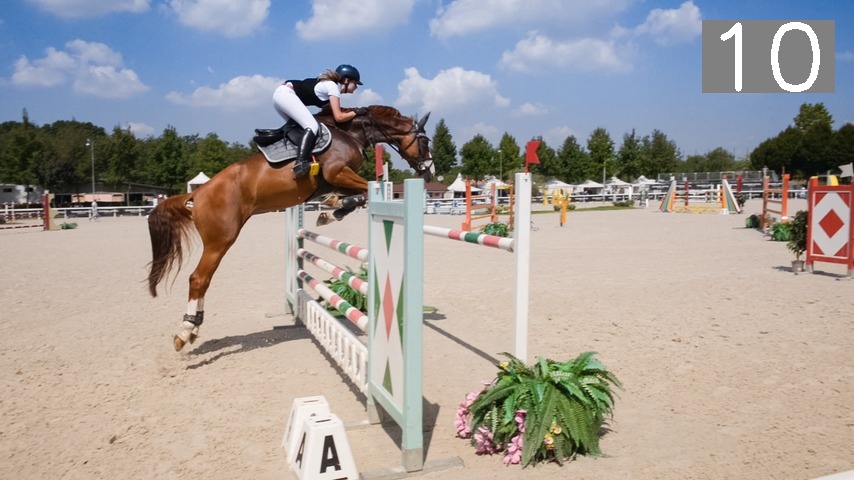}
         \label{fig:image}
     \end{subfigure}
     \hfill
     \begin{subfigure}[b]{0.16\textwidth}
         \centering
         \includegraphics[width=\textwidth]{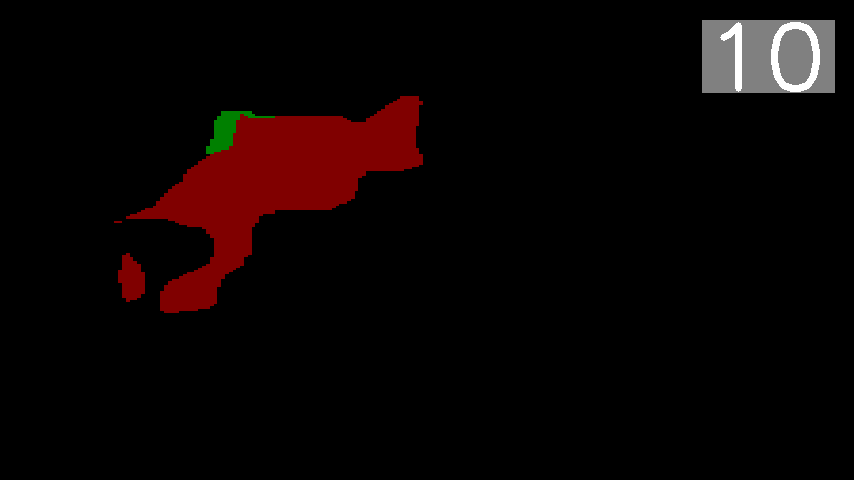}
         \label{fig:bilinear-openvoc}
     \end{subfigure}
     \hfill
     \begin{subfigure}[b]{0.16\textwidth}
         \centering
         \includegraphics[width=\textwidth]{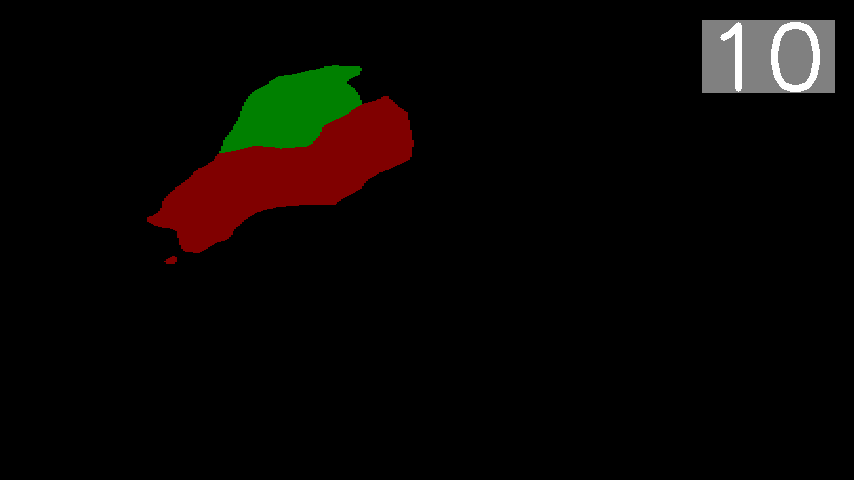}
         \label{fig:five over x}
     \end{subfigure}
     \hfill
     \begin{subfigure}[b]{0.16\textwidth}
         \centering
         \includegraphics[width=\textwidth]{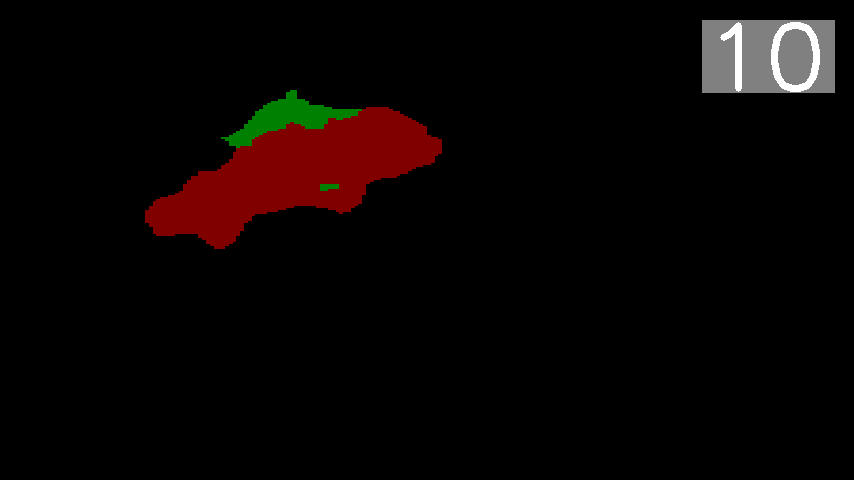}
         \label{fig:three sin x}
     \end{subfigure}
     \hfill
     \begin{subfigure}[b]{0.16\textwidth}
         \centering
         \includegraphics[width=\textwidth]{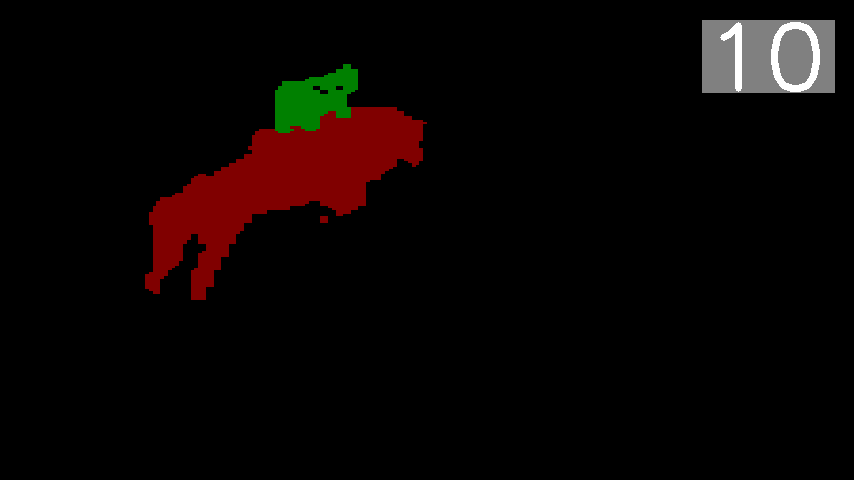}
         \label{fig:five over x}
     \end{subfigure}
     \hfill
     \begin{subfigure}[b]{0.16\textwidth}
         \centering
         \includegraphics[width=\textwidth]{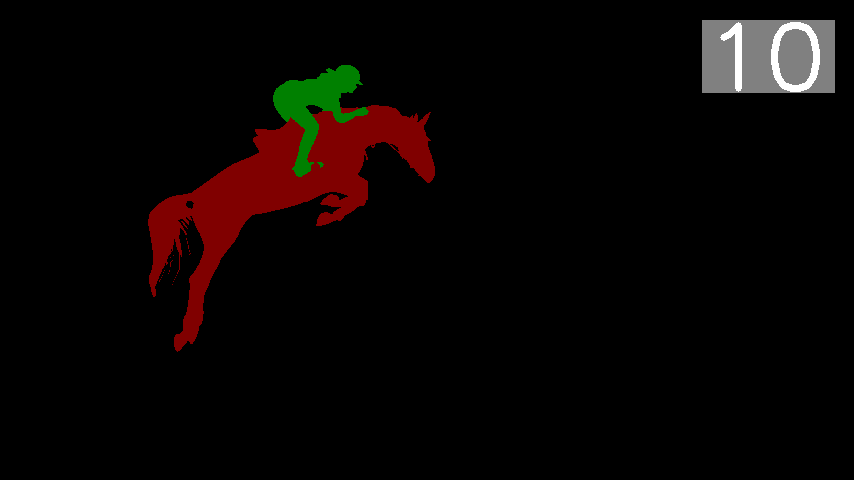}
         \label{fig:five over x}
     \end{subfigure}  \\
          \vspace{-0.2cm}
     \begin{subfigure}[b]{0.16\textwidth}
         \centering
         \includegraphics[width=\textwidth]{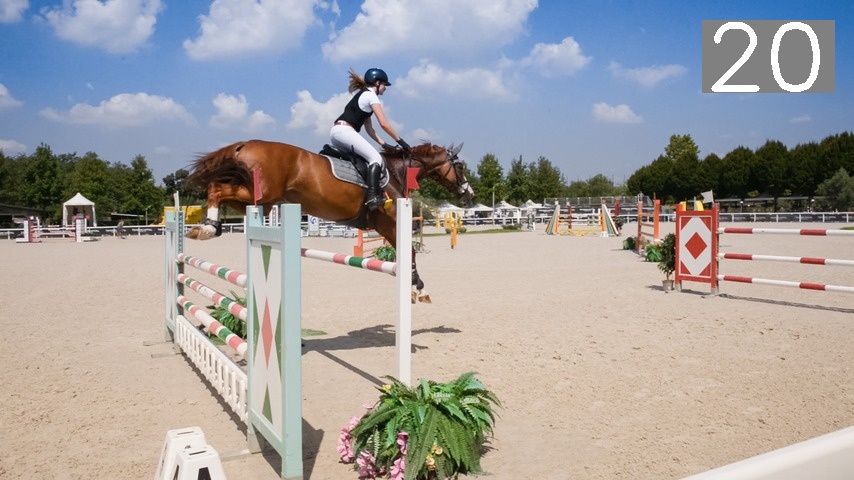}
         \label{fig:image}
     \end{subfigure}
     \hfill
     \begin{subfigure}[b]{0.16\textwidth}
         \centering
         \includegraphics[width=\textwidth]{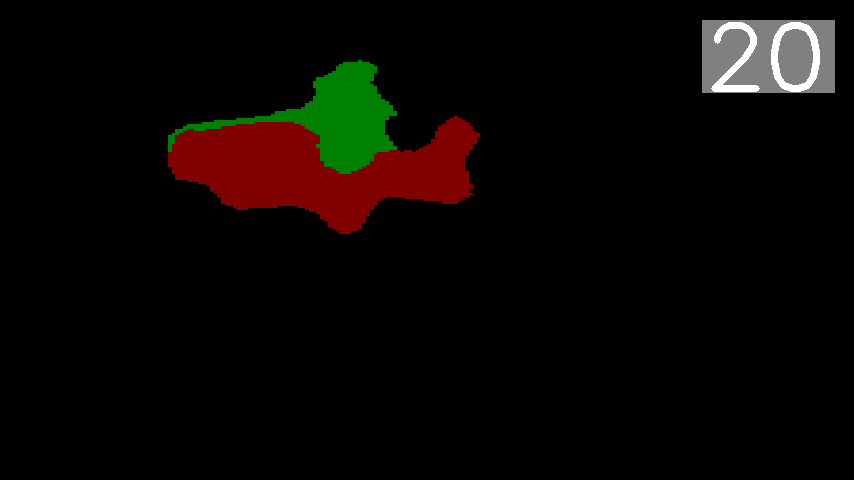}
         \label{fig:bilinear-openvoc}
     \end{subfigure}
     \hfill
     \begin{subfigure}[b]{0.16\textwidth}
         \centering
         \includegraphics[width=\textwidth]{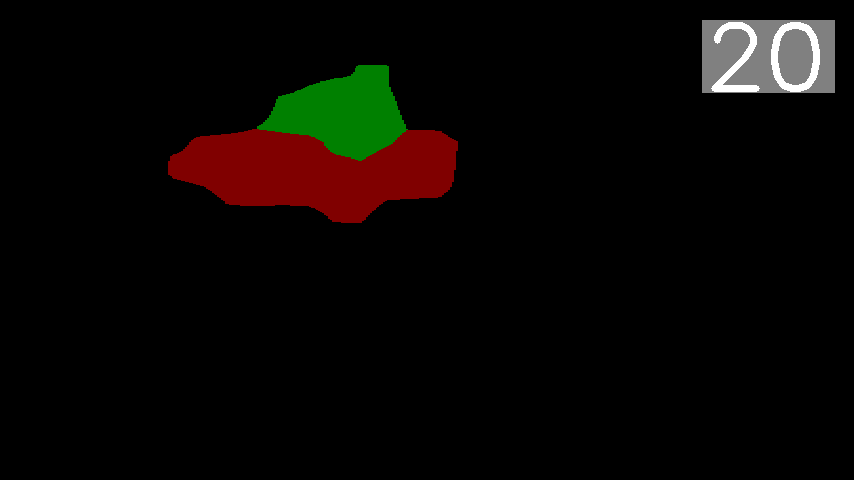}
         \label{fig:five over x}
     \end{subfigure}
     \hfill
     \begin{subfigure}[b]{0.16\textwidth}
         \centering
         \includegraphics[width=\textwidth]{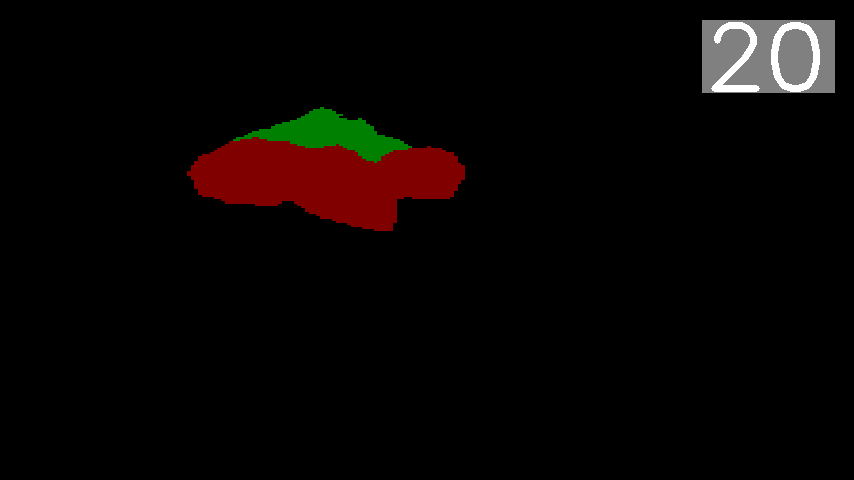}
         \label{fig:three sin x}
     \end{subfigure}
     \hfill
     \begin{subfigure}[b]{0.16\textwidth}
         \centering
         \includegraphics[width=\textwidth]{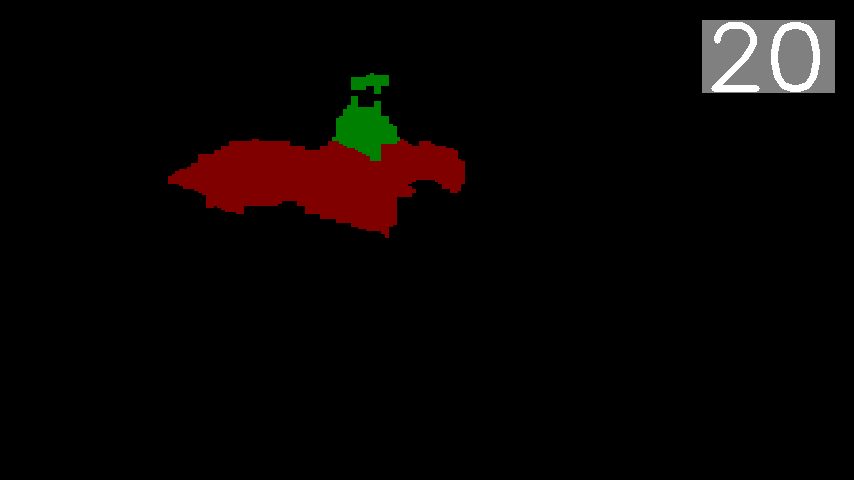}
         \label{fig:five over x}
     \end{subfigure}
     \hfill
     \begin{subfigure}[b]{0.16\textwidth}
         \centering
         \includegraphics[width=\textwidth]{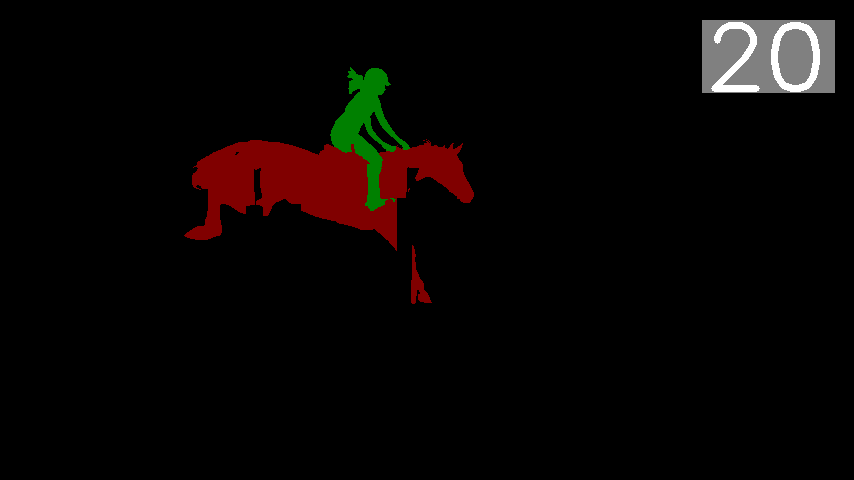}
         \label{fig:five over x}
     \end{subfigure}  \\
      \vspace{-0.2cm}
     \begin{subfigure}[b]{0.16\textwidth}
         \centering
         \includegraphics[width=\textwidth]{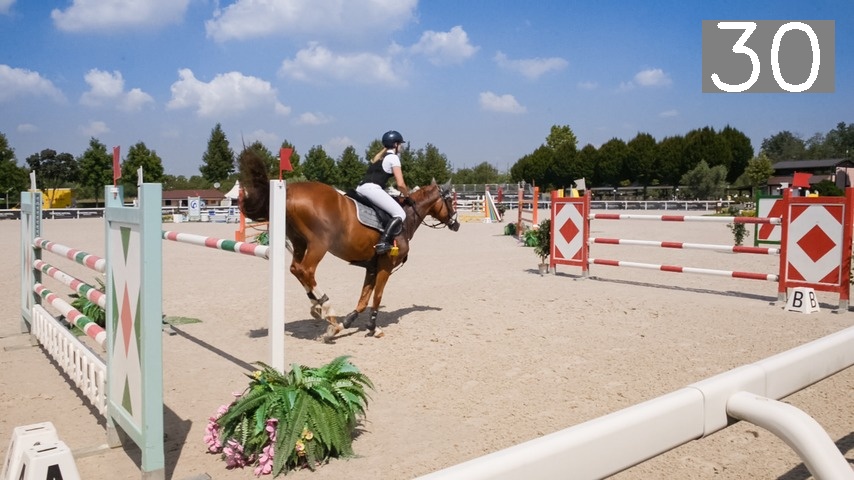}
         \label{fig:image}
     \end{subfigure}
     \hfill
     \begin{subfigure}[b]{0.16\textwidth}
         \centering
         \includegraphics[width=\textwidth]{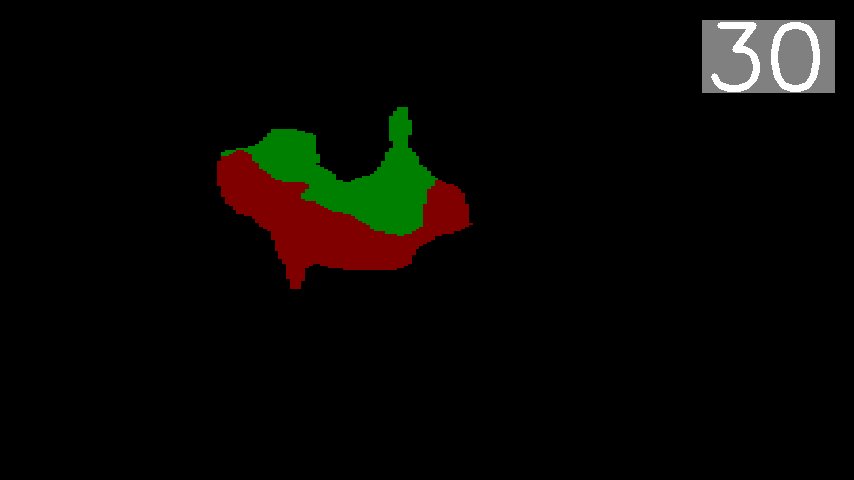}
         \label{fig:bilinear-openvoc}
     \end{subfigure}
     \hfill
     \begin{subfigure}[b]{0.16\textwidth}
         \centering
         \includegraphics[width=\textwidth]{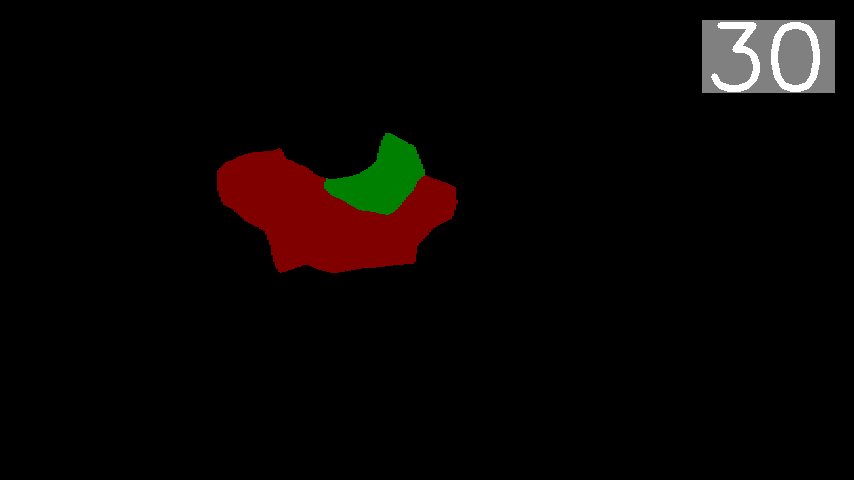}
         \label{fig:five over x}
     \end{subfigure}
     \hfill
     \begin{subfigure}[b]{0.16\textwidth}
         \centering
         \includegraphics[width=\textwidth]{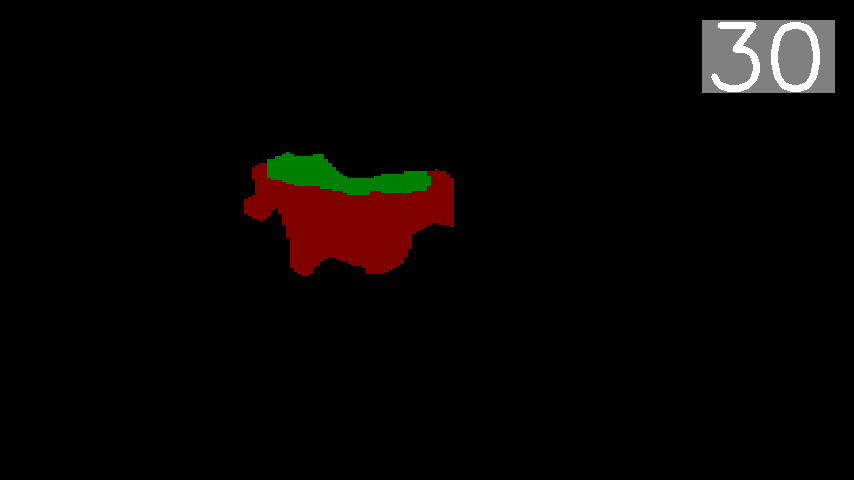}
         \label{fig:three sin x}
     \end{subfigure}
     \hfill
     \begin{subfigure}[b]{0.16\textwidth}
         \centering
         \includegraphics[width=\textwidth]{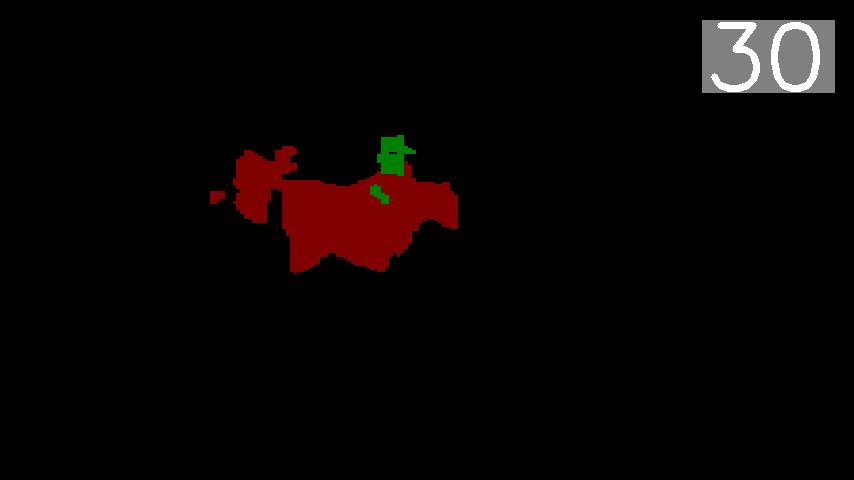}
         \label{fig:five over x}
     \end{subfigure}
     \hfill
     \begin{subfigure}[b]{0.16\textwidth}
         \centering
         \includegraphics[width=\textwidth]{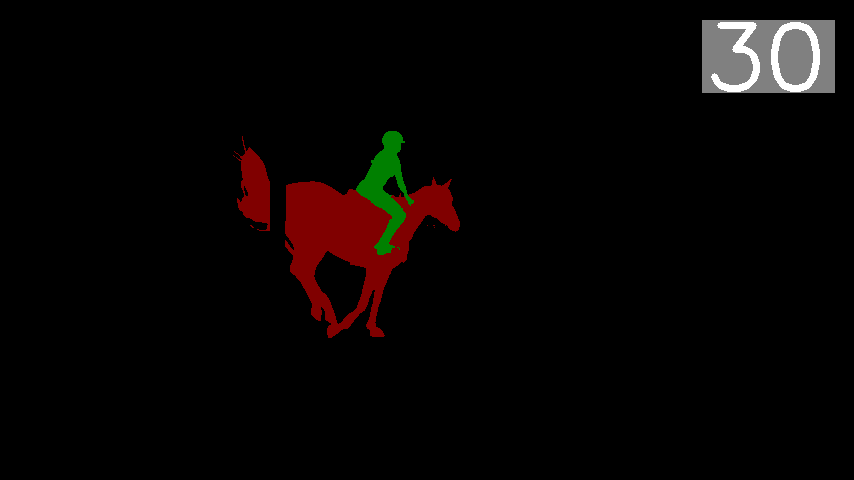}
         \label{fig:five over x}
     \end{subfigure}  \\
   
    \begin{subfigure}[b]{0.16\textwidth}
         \centering
         \includegraphics[width=\textwidth]{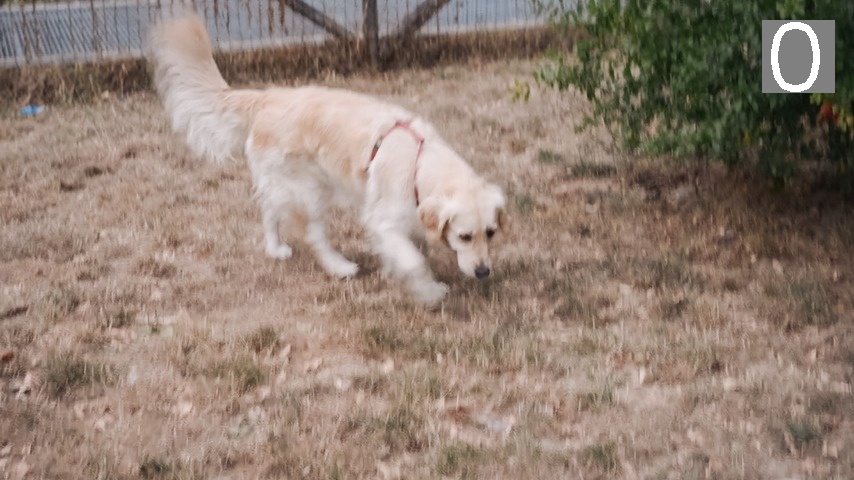}
         \label{fig:image}
     \end{subfigure}
     \hfill
     \begin{subfigure}[b]{0.16\textwidth}
         \centering
         \includegraphics[width=\textwidth]{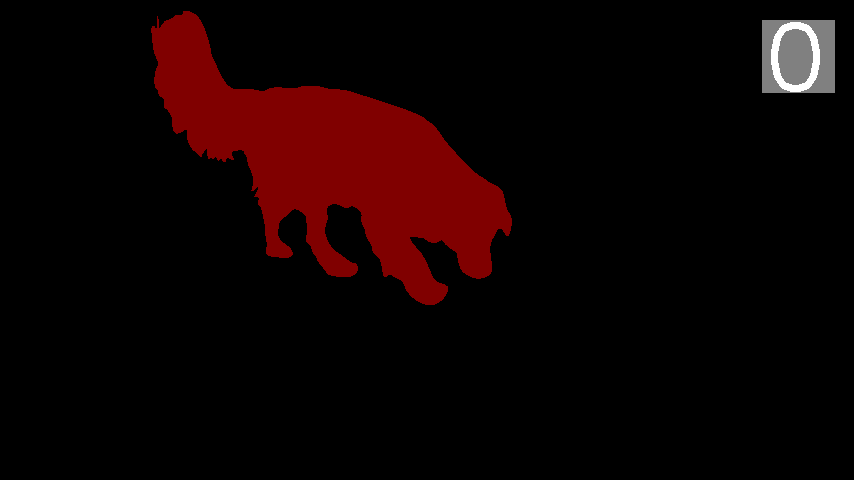}
         \label{fig:bilinear-openvoc}
     \end{subfigure}
     \hfill
     \begin{subfigure}[b]{0.16\textwidth}
         \centering
         \includegraphics[width=\textwidth]{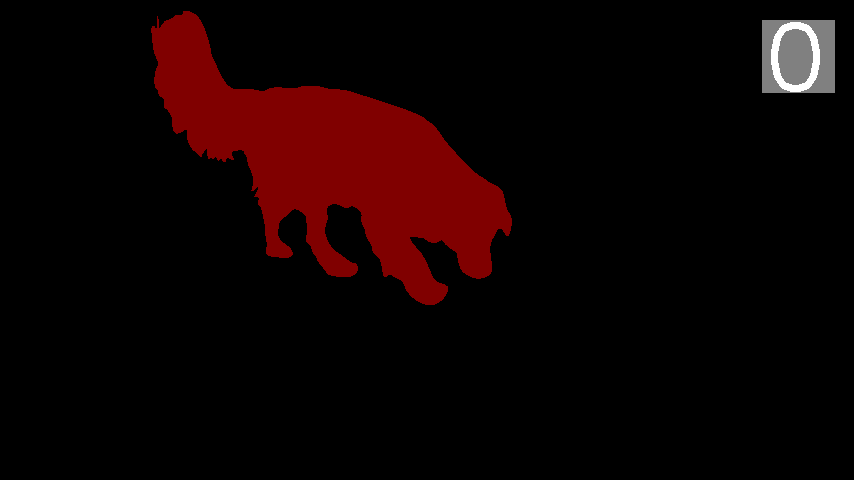}
         \label{fig:five over x}
     \end{subfigure}
     \hfill
     \begin{subfigure}[b]{0.16\textwidth}
         \centering
         \includegraphics[width=\textwidth]{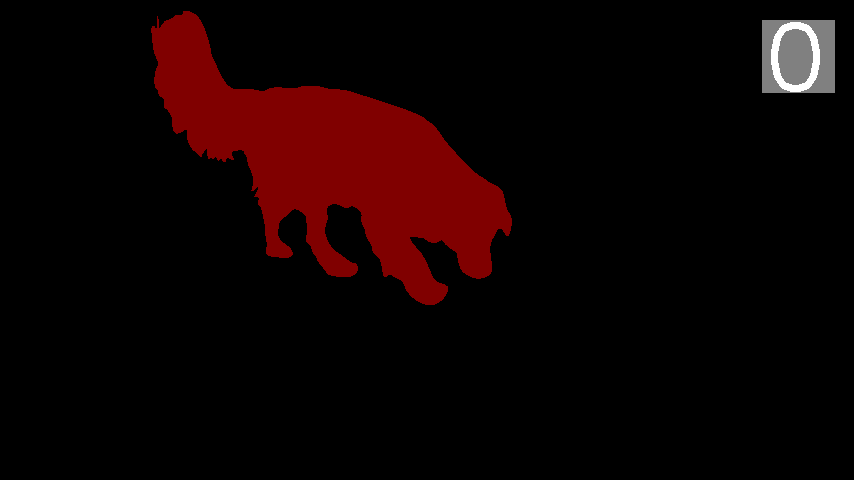}
         \label{fig:three sin x}
     \end{subfigure}
     \hfill
     \begin{subfigure}[b]{0.16\textwidth}
         \centering
         \includegraphics[width=\textwidth]{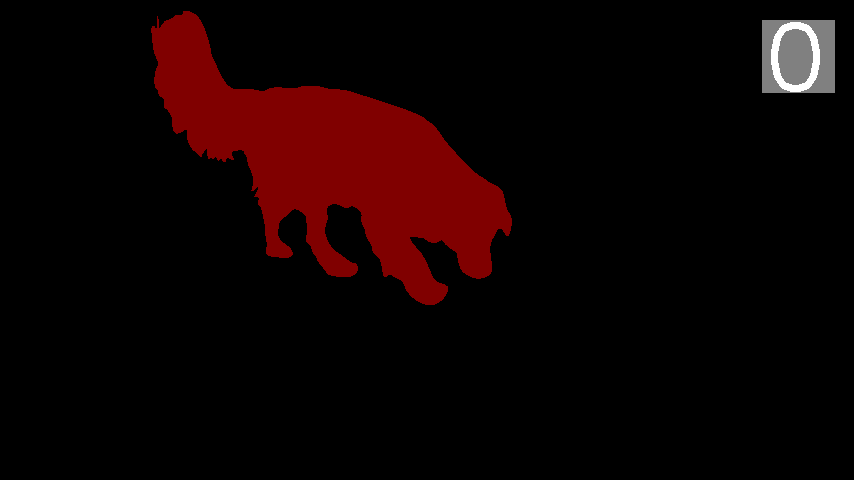}
         \label{fig:five over x}
     \end{subfigure}
     \hfill
     \begin{subfigure}[b]{0.16\textwidth}
         \centering
         \includegraphics[width=\textwidth]{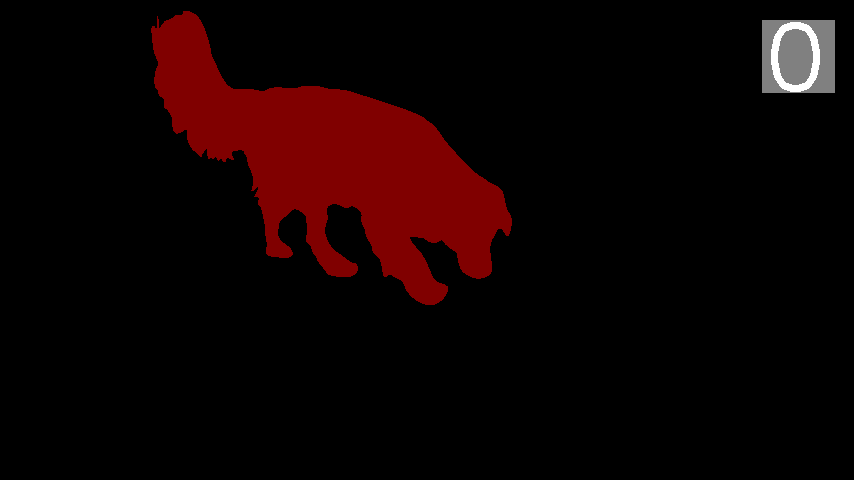}
         \label{fig:five over x}
     \end{subfigure}  \\
     \vspace{-0.2cm}
      \begin{subfigure}[b]{0.16\textwidth}
         \centering
         \includegraphics[width=\textwidth]{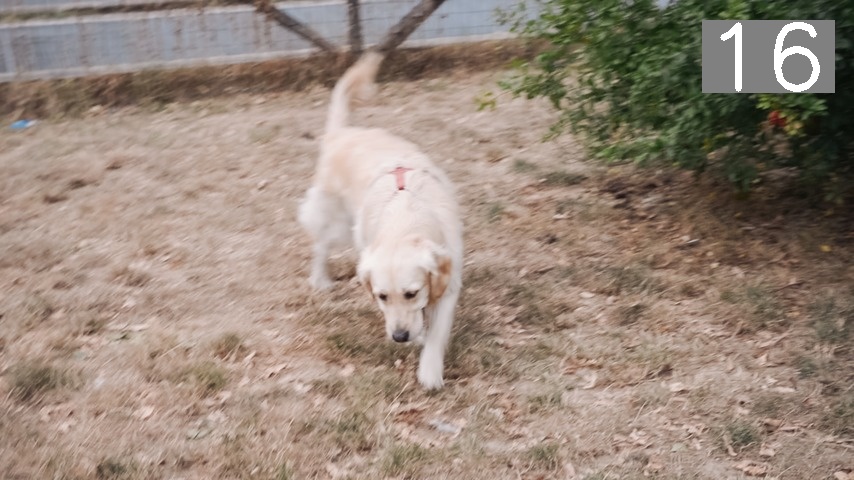}
         \label{fig:image}
     \end{subfigure}
     \hfill
     \begin{subfigure}[b]{0.16\textwidth}
         \centering
         \includegraphics[width=\textwidth]{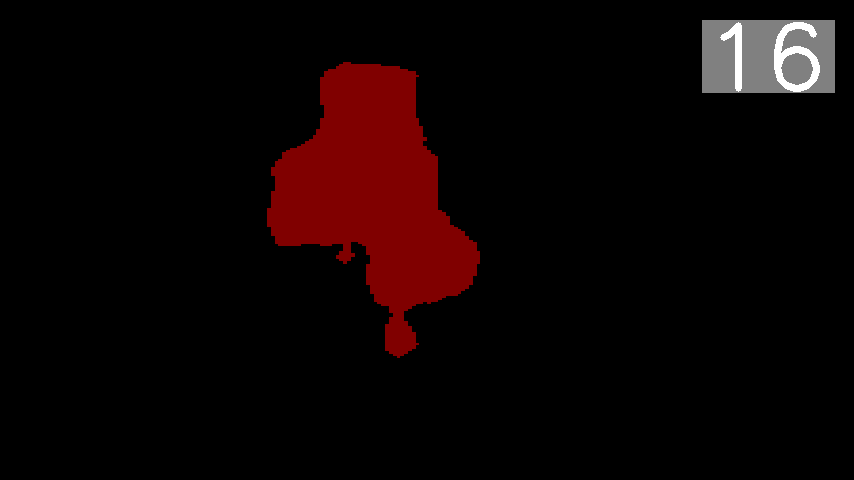}
         \label{fig:bilinear-openvoc}
     \end{subfigure}
     \hfill
     \begin{subfigure}[b]{0.16\textwidth}
         \centering
         \includegraphics[width=\textwidth]{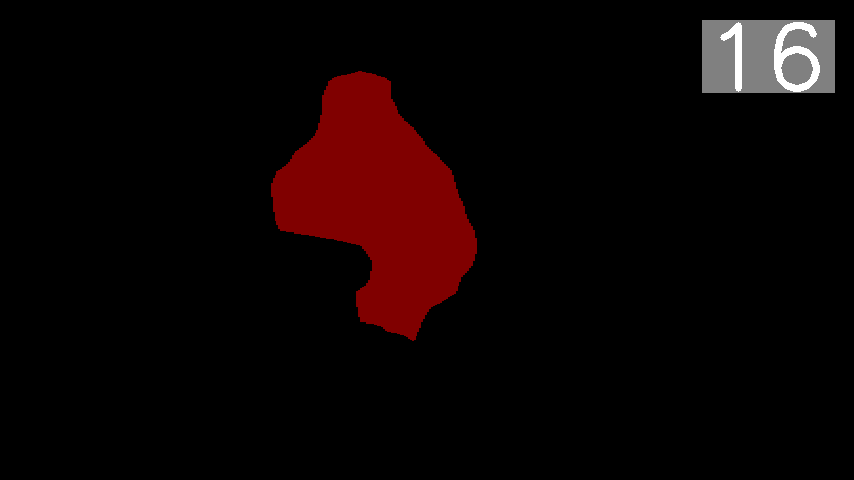}
         \label{fig:five over x}
     \end{subfigure}
     \hfill
     \begin{subfigure}[b]{0.16\textwidth}
         \centering
         \includegraphics[width=\textwidth]{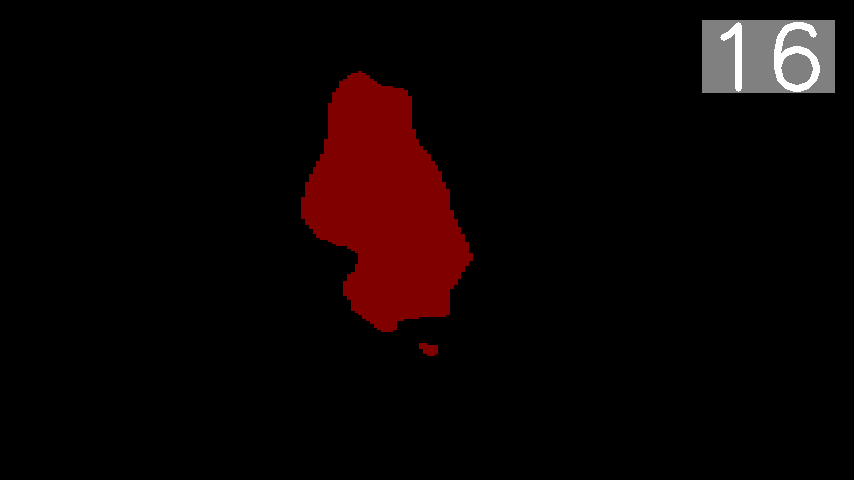}
         \label{fig:three sin x}
     \end{subfigure}
     \hfill
     \begin{subfigure}[b]{0.16\textwidth}
         \centering
         \includegraphics[width=\textwidth]{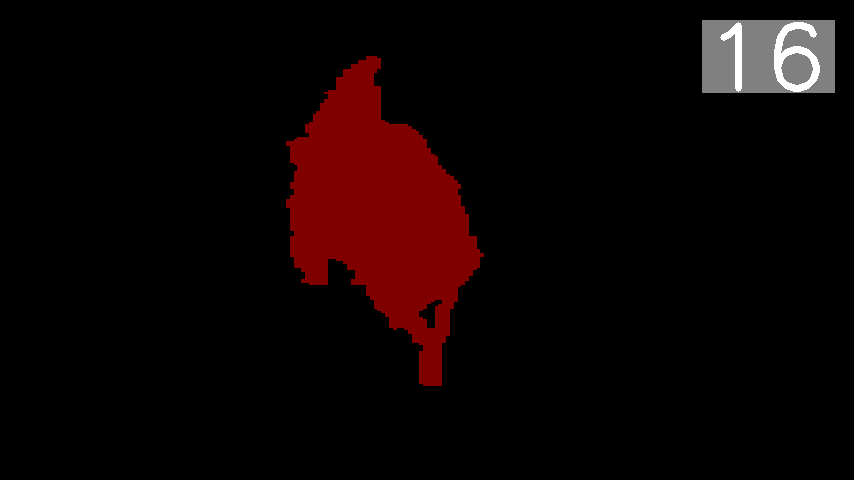}
         \label{fig:five over x}
     \end{subfigure}
     \hfill
     \begin{subfigure}[b]{0.16\textwidth}
         \centering
         \includegraphics[width=\textwidth]{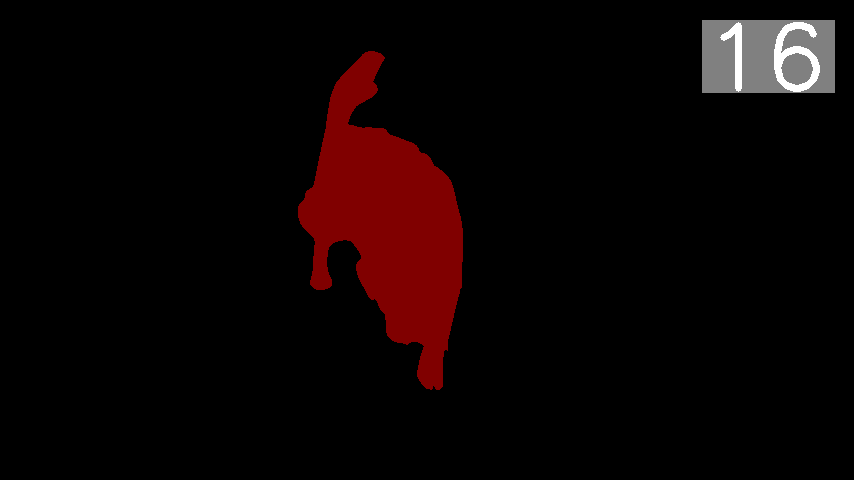}
         \label{fig:five over x}
     \end{subfigure}  \\
      \vspace{-0.2cm}
      \begin{subfigure}[b]{0.16\textwidth}
         \centering
         \includegraphics[width=\textwidth]{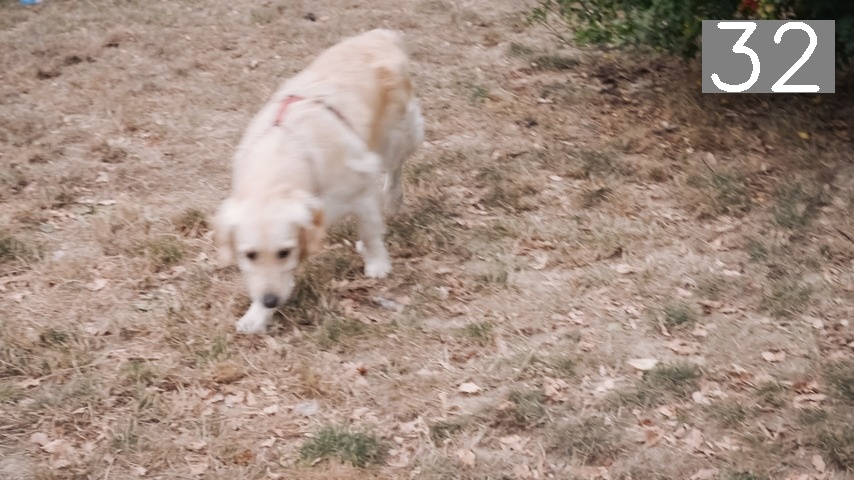}
         \label{fig:image}
     \end{subfigure}
     \hfill
     \begin{subfigure}[b]{0.16\textwidth}
         \centering
         \includegraphics[width=\textwidth]{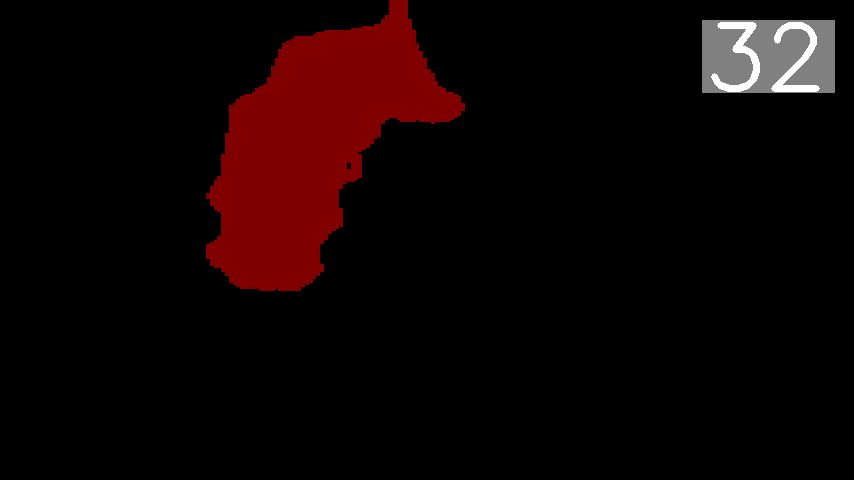}
         \label{fig:bilinear-openvoc}
     \end{subfigure}
     \hfill
     \begin{subfigure}[b]{0.16\textwidth}
         \centering
         \includegraphics[width=\textwidth]{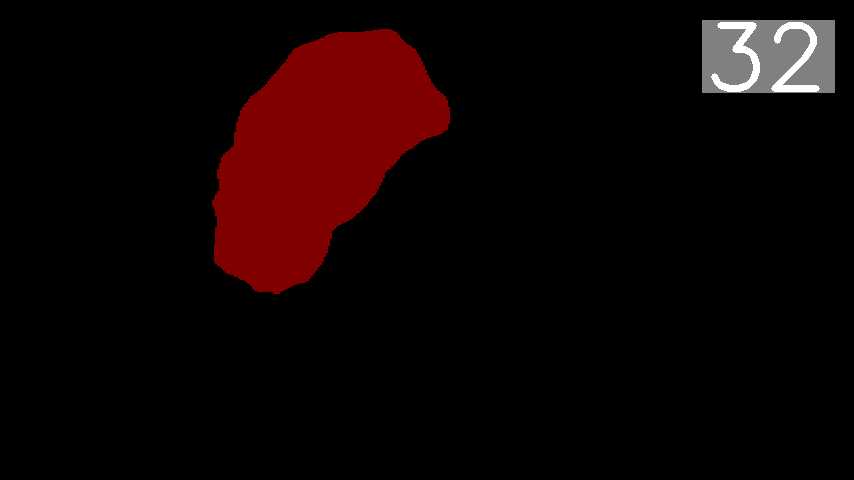}
         \label{fig:five over x}
     \end{subfigure}
     \hfill
     \begin{subfigure}[b]{0.16\textwidth}
         \centering
         \includegraphics[width=\textwidth]{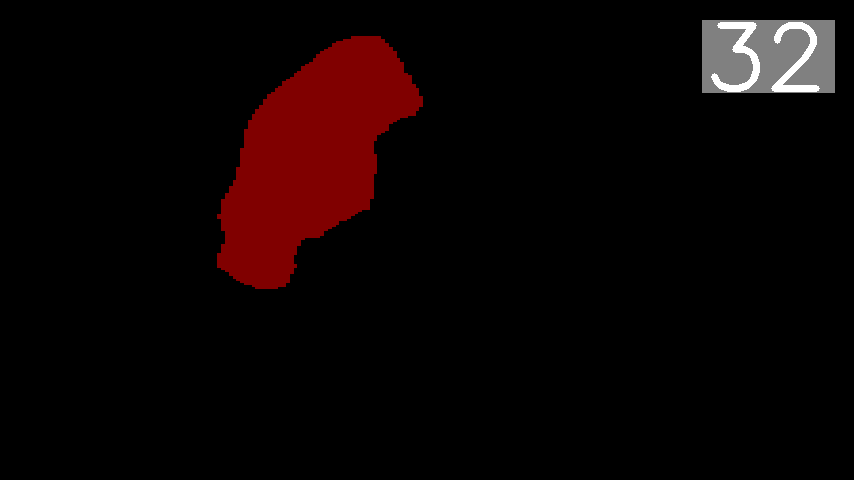}
         \label{fig:three sin x}
     \end{subfigure}
     \hfill
     \begin{subfigure}[b]{0.16\textwidth}
         \centering
         \includegraphics[width=\textwidth]{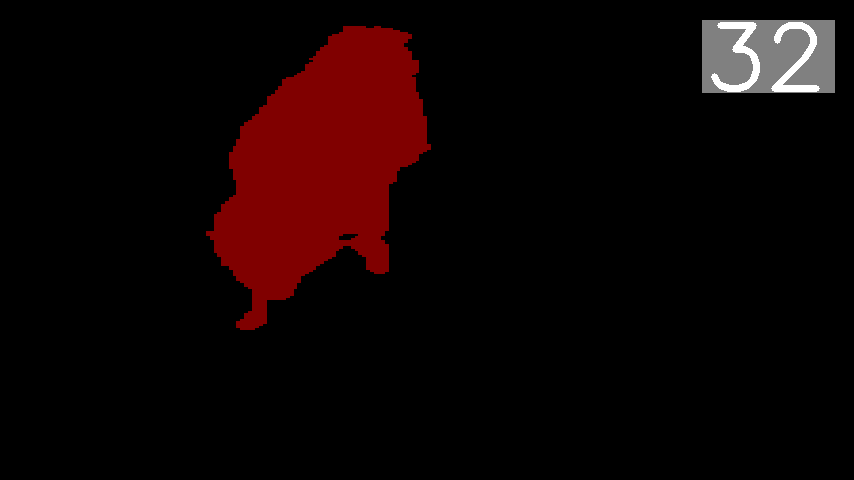}
         \label{fig:five over x}
     \end{subfigure}
     \hfill
     \begin{subfigure}[b]{0.16\textwidth}
         \centering
         \includegraphics[width=\textwidth]{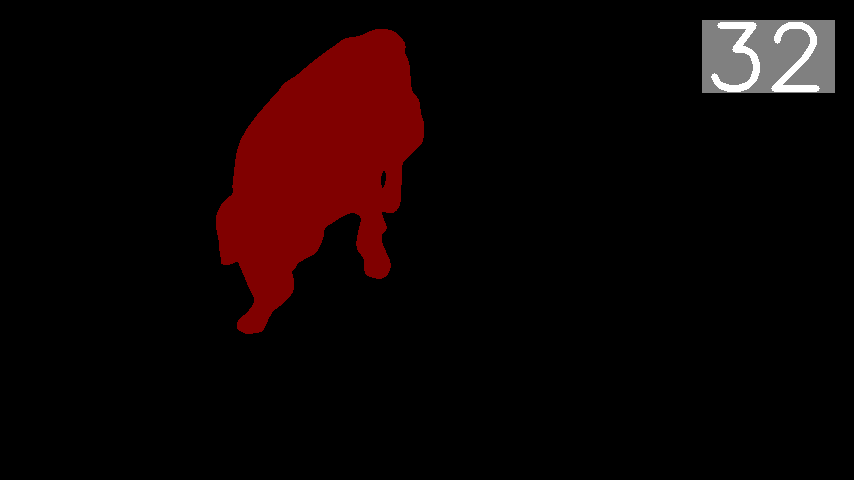}
         \label{fig:five over x}
     \end{subfigure}  \\
     \vspace{-0.2cm}
     \begin{subfigure}[b]{0.16\textwidth}
         \centering
         \includegraphics[width=\textwidth]{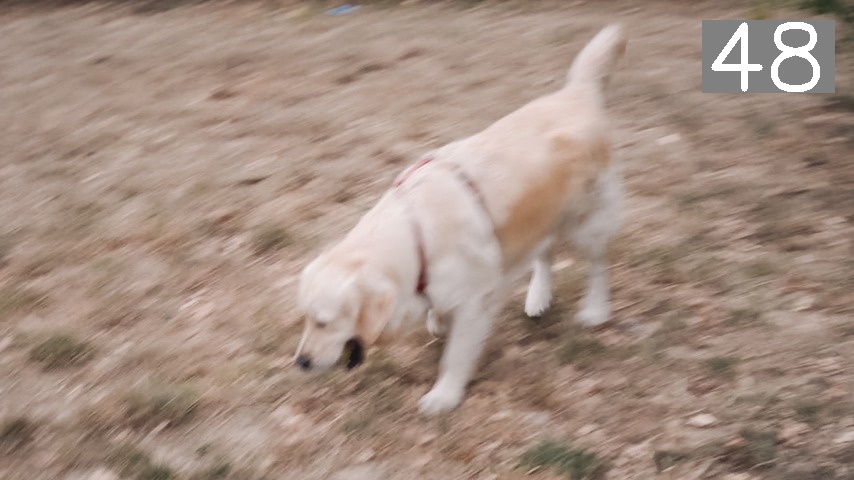}
         \caption{Original image}
         \label{fig:image}
     \end{subfigure}
     \hfill
     \begin{subfigure}[b]{0.16\textwidth}
         \centering
         \includegraphics[width=\textwidth]{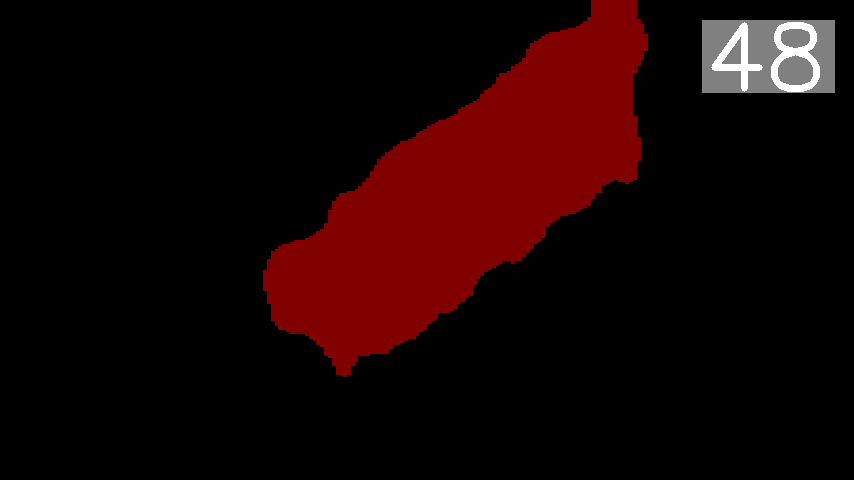}
         \caption{Bilinear}
         \label{fig:bilinear-openvoc}
     \end{subfigure}
     \hfill
     \begin{subfigure}[b]{0.16\textwidth}
         \centering
         \includegraphics[width=\textwidth]{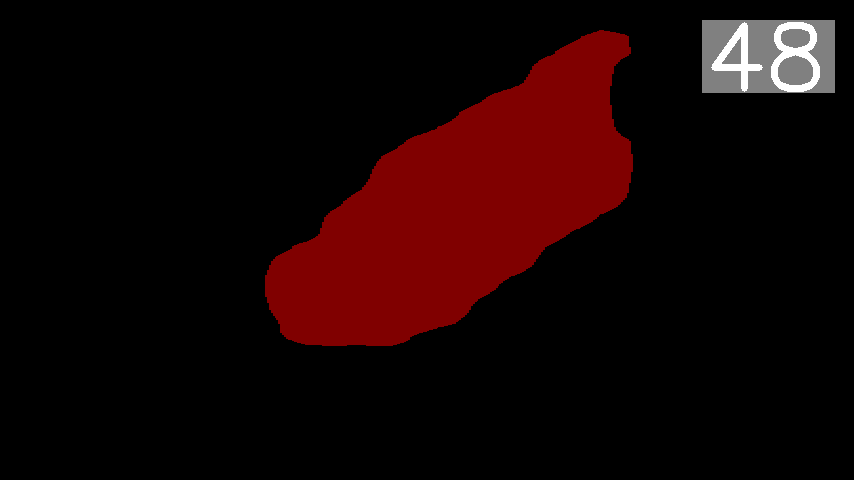}
         \caption{LiFT}
         \label{fig:five over x}
     \end{subfigure}
     \hfill
     \begin{subfigure}[b]{0.16\textwidth}
         \centering
         \includegraphics[width=\textwidth]{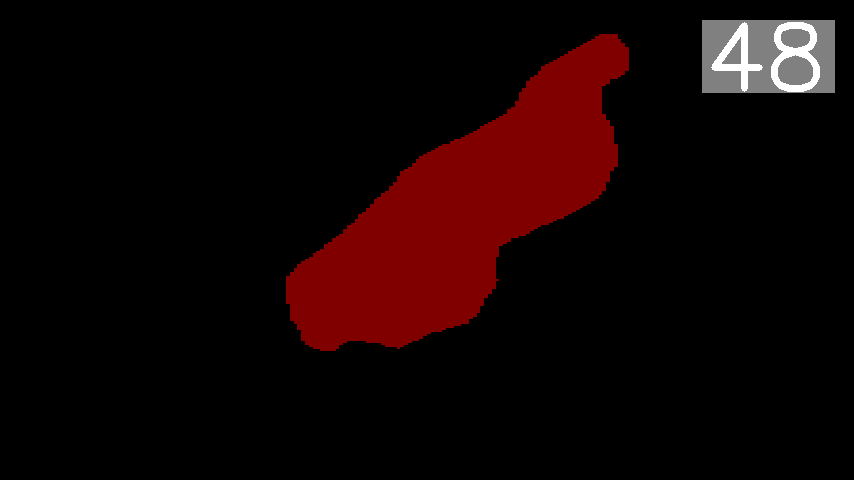}
         \caption{FeatUp}
         \label{fig:three sin x}
     \end{subfigure}
     \hfill
     \begin{subfigure}[b]{0.16\textwidth}
         \centering
         \includegraphics[width=\textwidth]{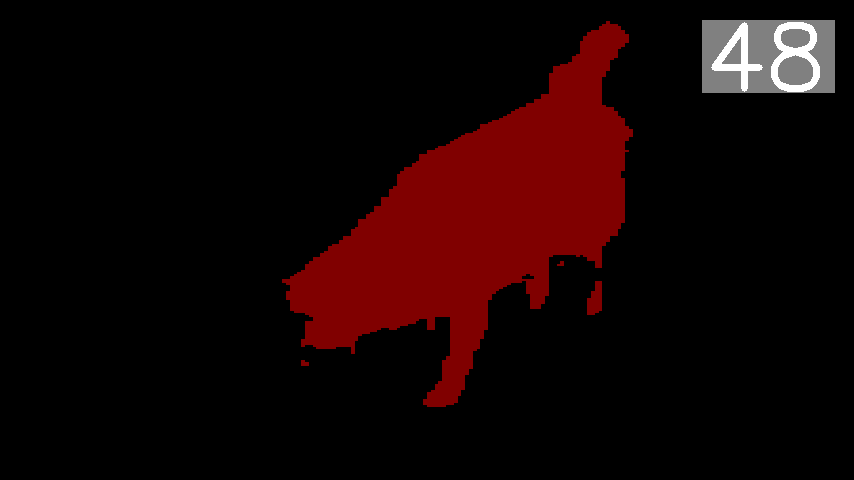}
         \caption{\textbf{LoftUp}}
         \label{fig:five over x}
     \end{subfigure}
     \hfill
     \begin{subfigure}[b]{0.16\textwidth}
         \centering
         \includegraphics[width=\textwidth]{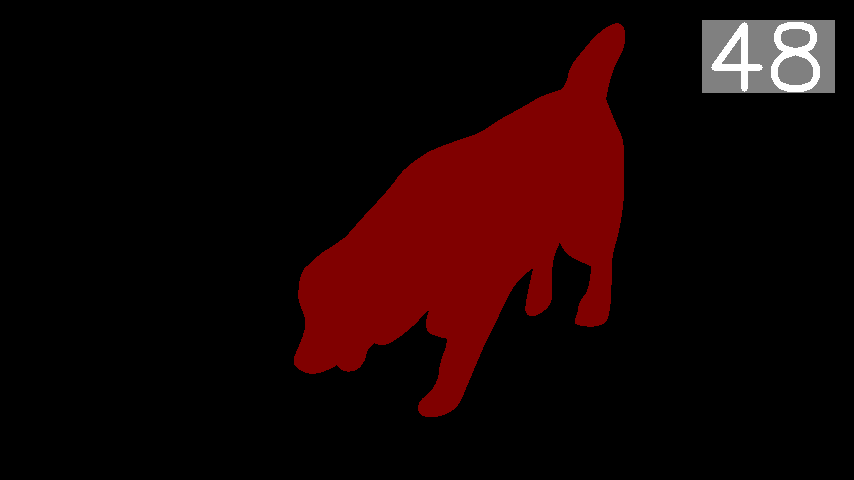}
         \caption{Groundtruth}
         \label{fig:five over x}
     \end{subfigure}
    \caption{\textbf{Visualization of prediction examples} of video object segmentation on the DAVIS 2017 dataset~\cite{Pont-Tuset_arXiv_2017_davis}. Each image displays its corresponding frame number in the top right corner. The groundtruth segmentation for the 0-th frame is provided, and dense feature affinity maps are employed to propagate its segmentation labels to subsequent frames. We can see that LoftUp outperforms all the other baselines in accurately tracking the objects across the frames.}
    \label{fig:vis-preds-video}
\end{figure*}

\begin{figure*}[t]
     \centering
     \begin{subfigure}[b]{0.19\textwidth}
         \centering
         \includegraphics[width=\textwidth]{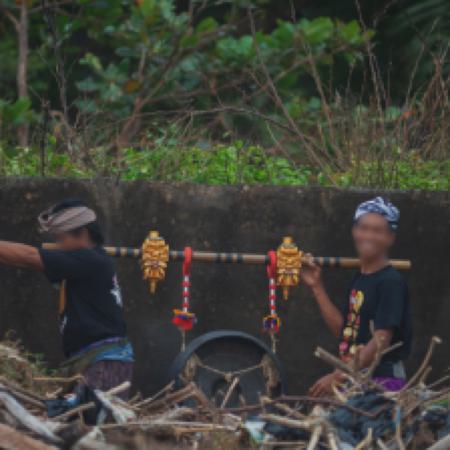}
         \label{fig:original}
     \end{subfigure}
     \hfill
     \begin{subfigure}[b]{0.19\textwidth}
         \centering
         \includegraphics[width=\textwidth]{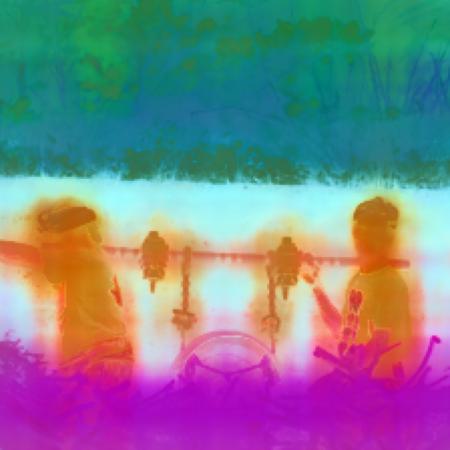}
         \label{fig:pseudo-gt-1}
     \end{subfigure}
     \hfill
     \begin{subfigure}[b]{0.19\textwidth}
         \centering
         \includegraphics[width=\textwidth]{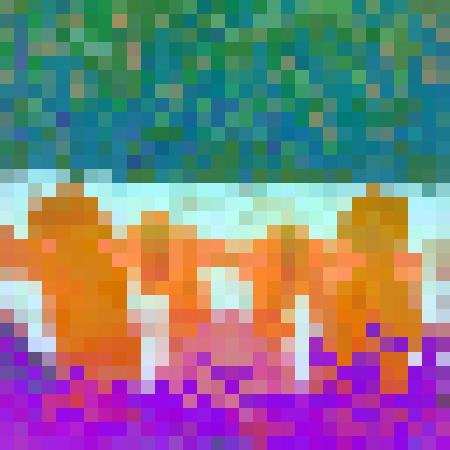}
         \label{fig:three sin x}
     \end{subfigure}
     \hfill
     \begin{subfigure}[b]{0.19\textwidth}
         \centering
         \includegraphics[width=\textwidth]{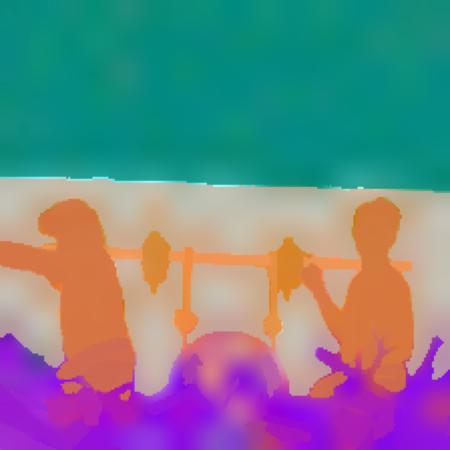}
         \label{fig:five over x}
     \end{subfigure}
     \begin{subfigure}[b]{0.19\textwidth}
         \centering
         \includegraphics[width=\textwidth]{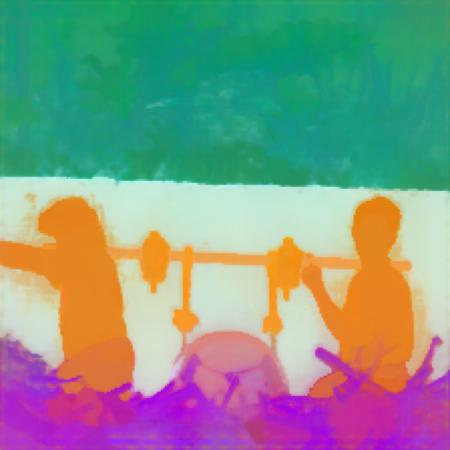}
         \label{fig:five over x}
     \end{subfigure}
     \\
     \begin{subfigure}[b]{0.19\textwidth}
         \centering
         \includegraphics[width=\textwidth]{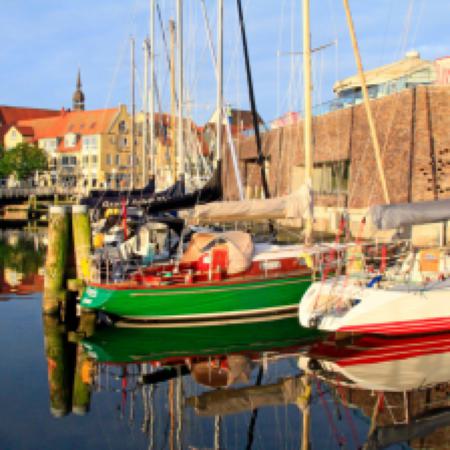}
         \label{fig:original}
     \end{subfigure}
     \hfill
     \begin{subfigure}[b]{0.19\textwidth}
         \centering
         \includegraphics[width=\textwidth]{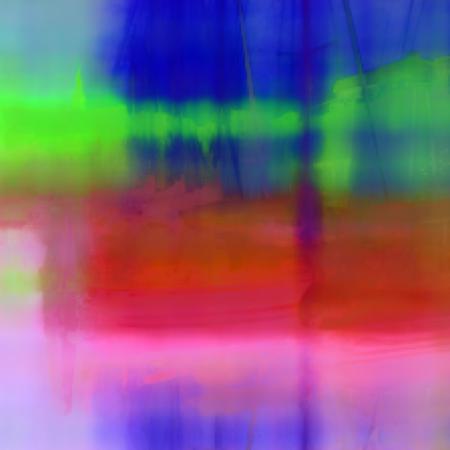}
         \label{fig:pseudo-gt-1}
     \end{subfigure}
     \hfill
     \begin{subfigure}[b]{0.19\textwidth}
         \centering
         \includegraphics[width=\textwidth]{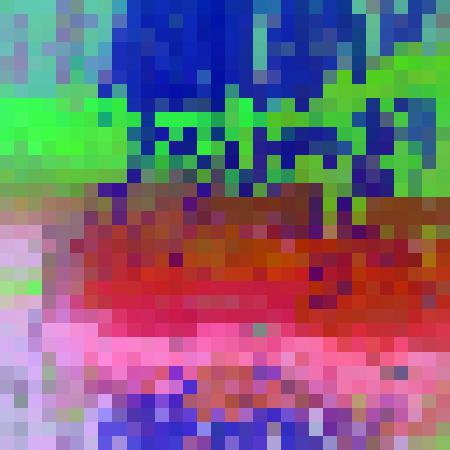}
         \label{fig:three sin x}
     \end{subfigure}
     \hfill
     \begin{subfigure}[b]{0.19\textwidth}
         \centering
         \includegraphics[width=\textwidth]{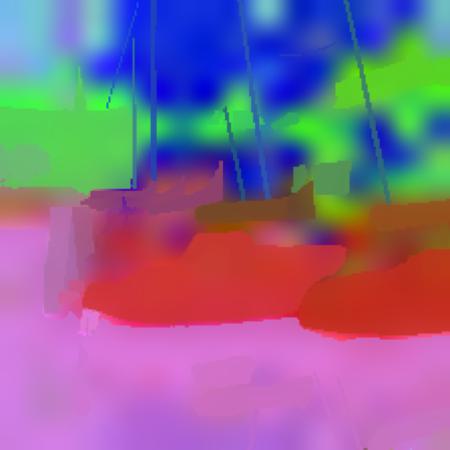}
         \label{fig:five over x}
     \end{subfigure}
     \begin{subfigure}[b]{0.19\textwidth}
         \centering
         \includegraphics[width=\textwidth]{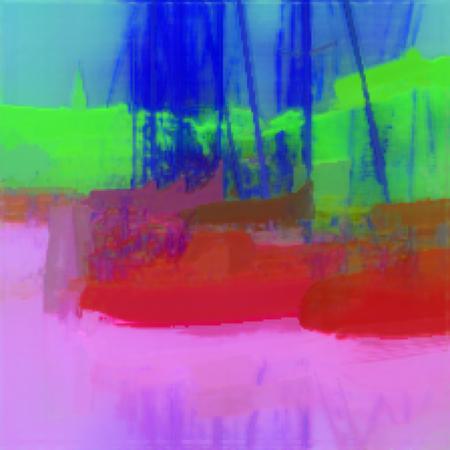}
         \label{fig:five over x}
     \end{subfigure}
     \\
     \begin{subfigure}[b]{0.19\textwidth}
         \centering
         \includegraphics[width=\textwidth]{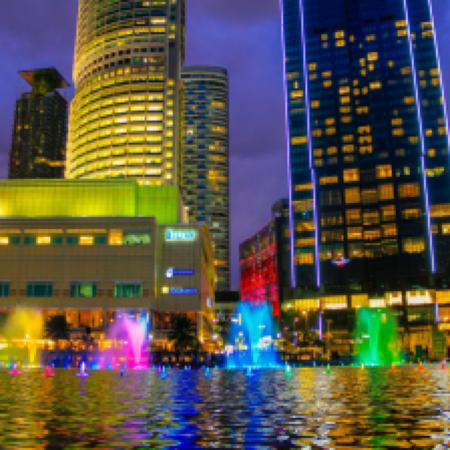}
         \label{fig:original}
     \end{subfigure}
     \hfill
     \begin{subfigure}[b]{0.19\textwidth}
         \centering
         \includegraphics[width=\textwidth]{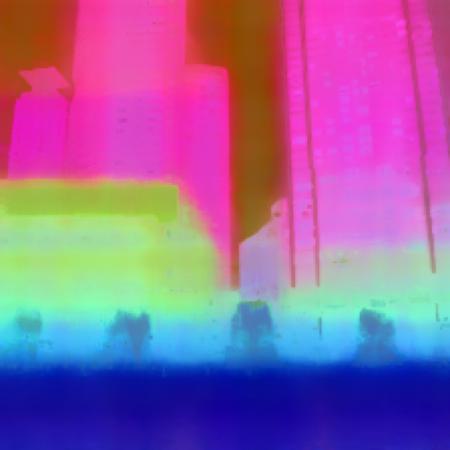}
         \label{fig:pseudo-gt-1}
     \end{subfigure}
     \hfill
     \begin{subfigure}[b]{0.19\textwidth}
         \centering
         \includegraphics[width=\textwidth]{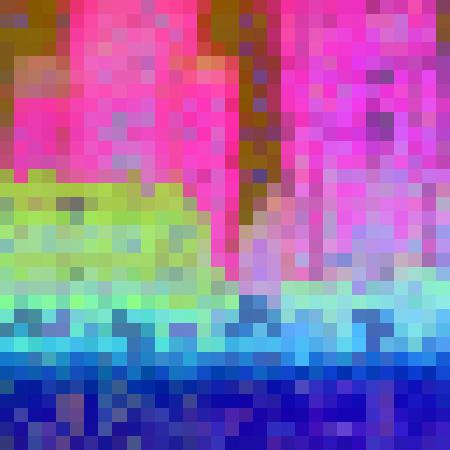}
         \label{fig:three sin x}
     \end{subfigure}
     \hfill
     \begin{subfigure}[b]{0.19\textwidth}
         \centering
         \includegraphics[width=\textwidth]{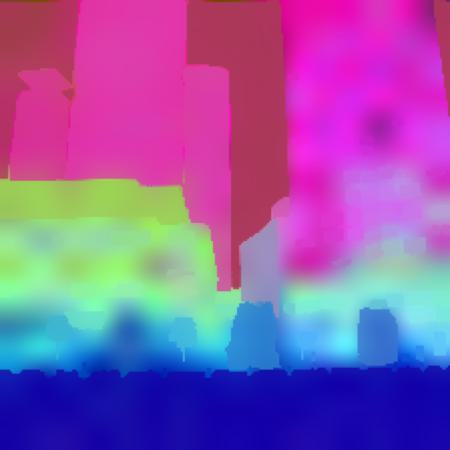}
         \label{fig:five over x}
     \end{subfigure}
     \begin{subfigure}[b]{0.19\textwidth}
         \centering
         \includegraphics[width=\textwidth]{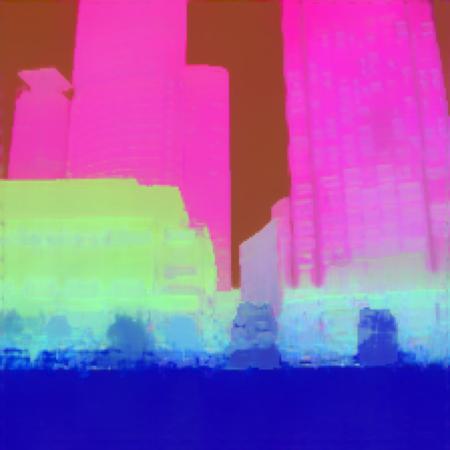}
         \label{fig:five over x}
     \end{subfigure}
     \\
      \begin{subfigure}[b]{0.19\textwidth}
         \centering
         \includegraphics[width=\textwidth]{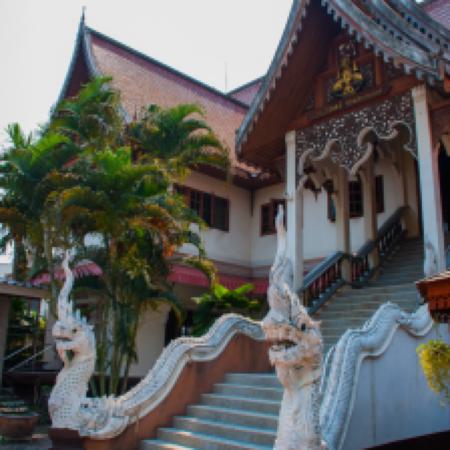}
         \label{fig:original}
     \end{subfigure}
     \hfill
     \begin{subfigure}[b]{0.19\textwidth}
         \centering
         \includegraphics[width=\textwidth]{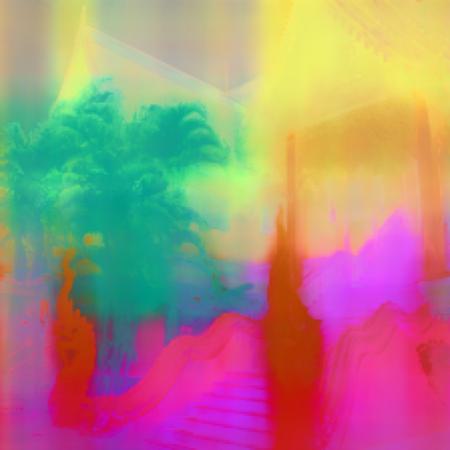}
         \label{fig:pseudo-gt-1}
     \end{subfigure}
     \hfill
     \begin{subfigure}[b]{0.19\textwidth}
         \centering
         \includegraphics[width=\textwidth]{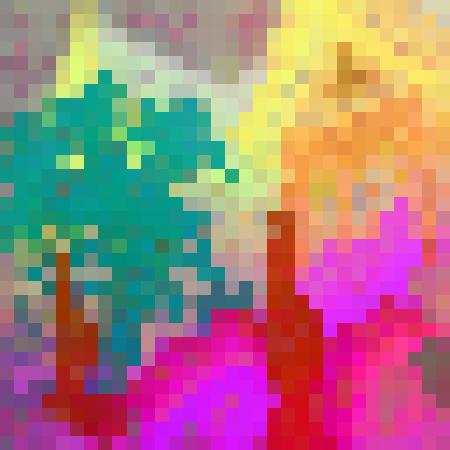}
         \label{fig:three sin x}
     \end{subfigure}
     \hfill
     \begin{subfigure}[b]{0.19\textwidth}
         \centering
         \includegraphics[width=\textwidth]{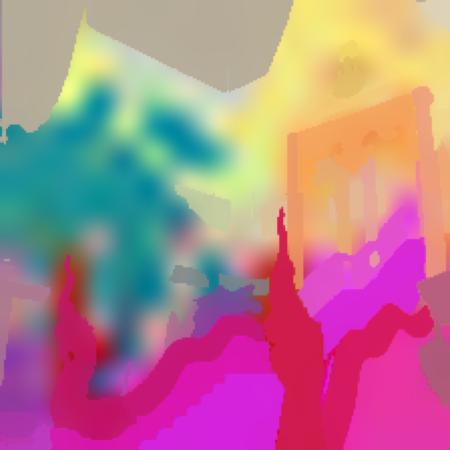}
         \label{fig:five over x}
     \end{subfigure}
     \begin{subfigure}[b]{0.19\textwidth}
         \centering
         \includegraphics[width=\textwidth]{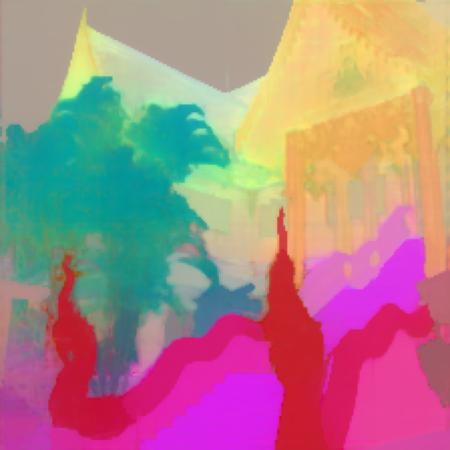}
         \label{fig:five over x}
     \end{subfigure}
     \\
     \begin{subfigure}[b]{0.19\textwidth}
         \centering
         \includegraphics[width=\textwidth]{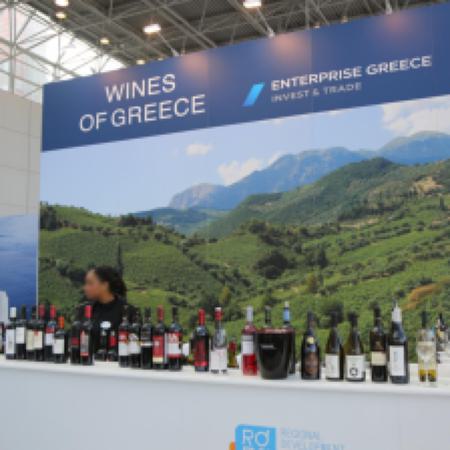}
         \label{fig:original}
     \end{subfigure}
     \hfill
     \begin{subfigure}[b]{0.19\textwidth}
         \centering
         \includegraphics[width=\textwidth]{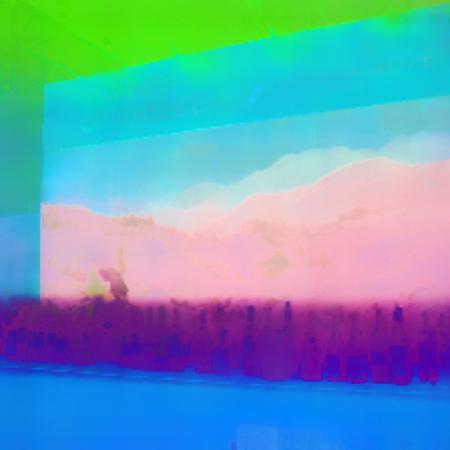}
         \label{fig:pseudo-gt-1}
     \end{subfigure}
     \hfill
     \begin{subfigure}[b]{0.19\textwidth}
         \centering
         \includegraphics[width=\textwidth]{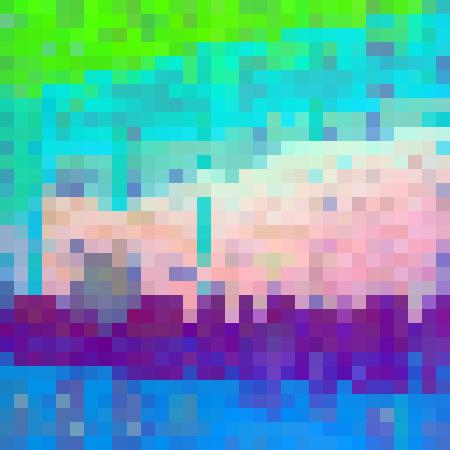}
         \label{fig:three sin x}
     \end{subfigure}
     \hfill
     \begin{subfigure}[b]{0.19\textwidth}
         \centering
         \includegraphics[width=\textwidth]{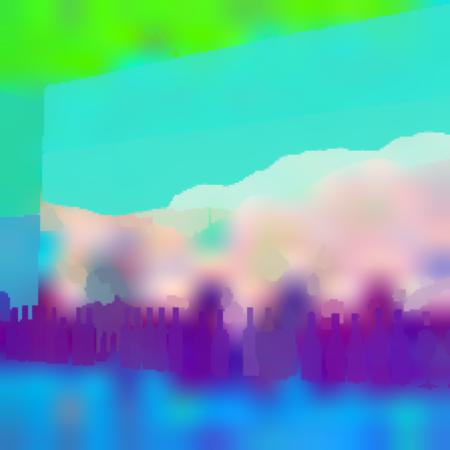}
         \label{fig:five over x}
     \end{subfigure}
     \begin{subfigure}[b]{0.19\textwidth}
         \centering
         \includegraphics[width=\textwidth]{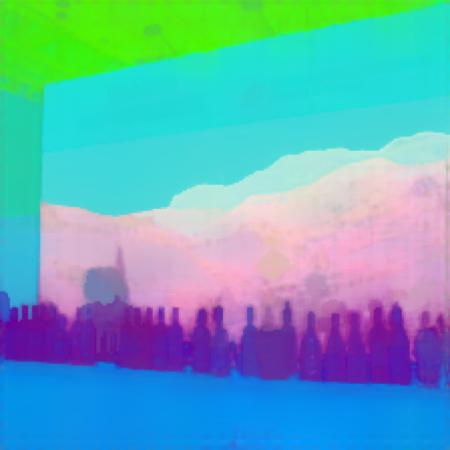}
         \label{fig:five over x}
     \end{subfigure}
     \\
     \begin{subfigure}[b]{0.19\textwidth}
         \centering
         \includegraphics[width=\textwidth]{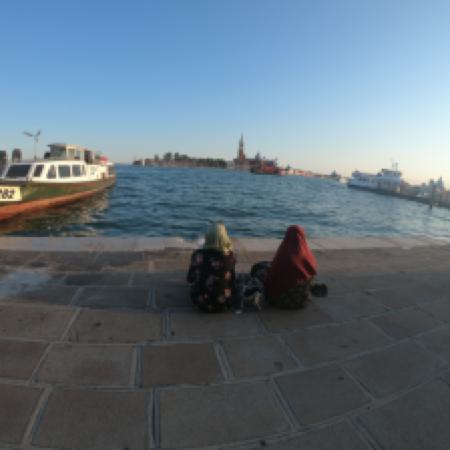}
         \caption{Original image}
         \label{fig:original}
     \end{subfigure}
     \hfill
     \begin{subfigure}[b]{0.19\textwidth}
         \centering
         \includegraphics[width=\textwidth]{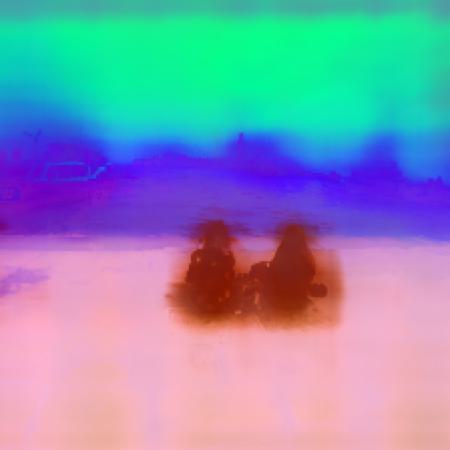}
         \caption{Per-image optimized}
         \label{fig:pseudo-gt-1}
     \end{subfigure}
     \hfill
     \begin{subfigure}[b]{0.19\textwidth}
         \centering
         \includegraphics[width=\textwidth]{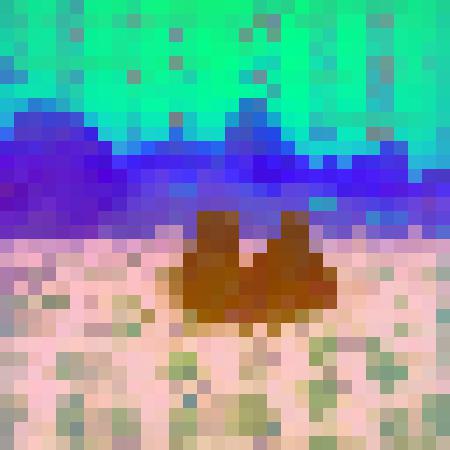}
         \caption{2x features (LiFT)}
         \label{fig:three sin x}
     \end{subfigure}
     \hfill
     \begin{subfigure}[b]{0.19\textwidth}
         \centering
         \includegraphics[width=\textwidth]{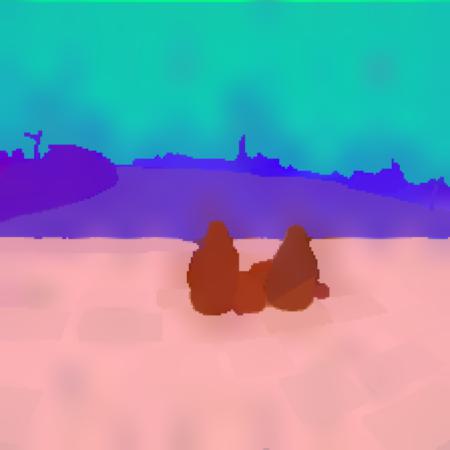}
         \caption{Mask-Bicubic}
         \label{fig:five over x}
     \end{subfigure}
     \begin{subfigure}[b]{0.19\textwidth}
         \centering
         \includegraphics[width=\textwidth]{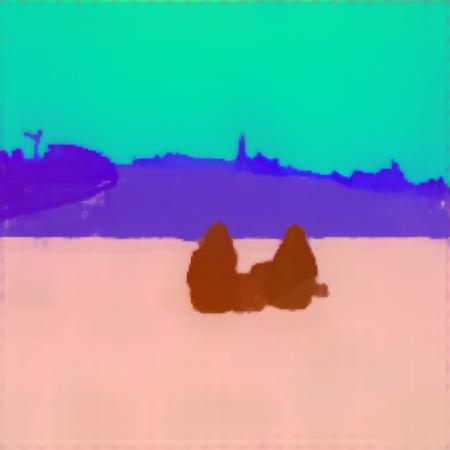}
         \caption{Self-Distilled}
         \label{fig:five over x}
     \end{subfigure}
        \caption{\textbf{More visualization of different pseudo-GT.} }
        \label{fig:more-pseudo-gt}
\end{figure*}

\begin{figure*}[t]
     \centering
     \begin{subfigure}[b]{0.24\linewidth}
         \centering
         \includegraphics[width=\linewidth]{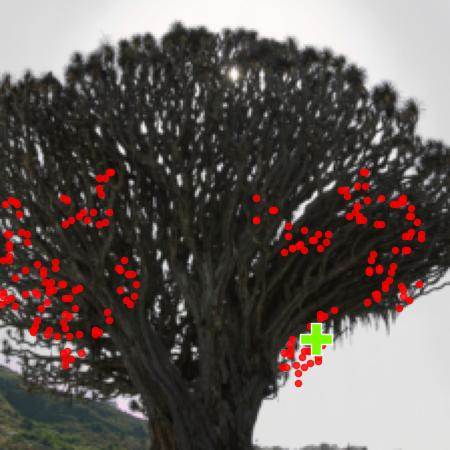}
         \label{fig:original}
     \end{subfigure}
     \begin{subfigure}[b]{0.24\linewidth}
         \centering
         \includegraphics[width=\linewidth]{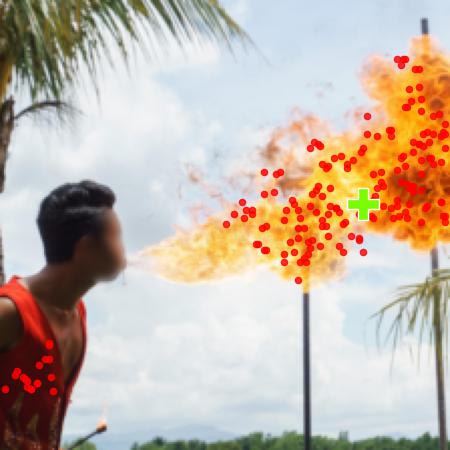}
         \label{fig:original}
     \end{subfigure} 
     \vspace{-0.3cm}
     \begin{subfigure}[b]{0.24\linewidth}
         \centering
         \includegraphics[width=\linewidth]{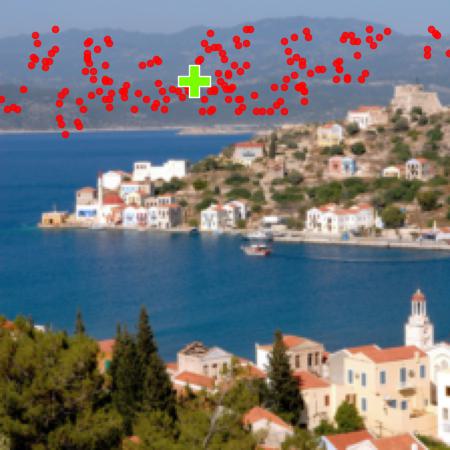}
         \label{fig:original}
     \end{subfigure}
     \begin{subfigure}[b]{0.24\linewidth}
         \centering
         \includegraphics[width=\linewidth]{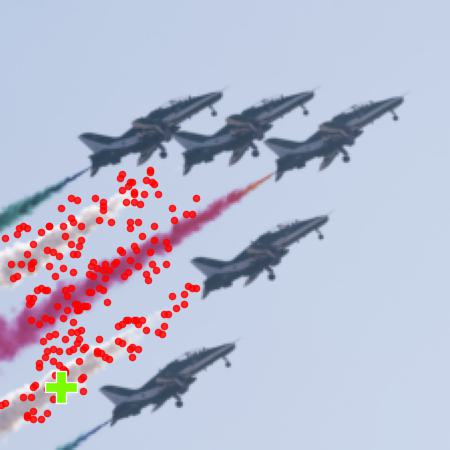}
         \label{fig:original}
     \end{subfigure}
     \\
     \begin{subfigure}[b]{0.24\linewidth}
         \centering
         \includegraphics[width=\linewidth]{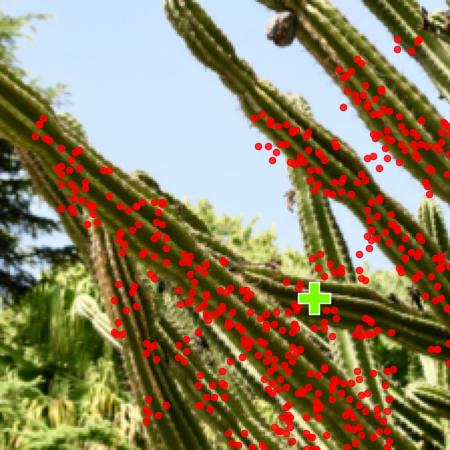}
         \label{fig:original}
     \end{subfigure}
     \begin{subfigure}[b]{0.24\linewidth}
         \centering
         \includegraphics[width=\linewidth]{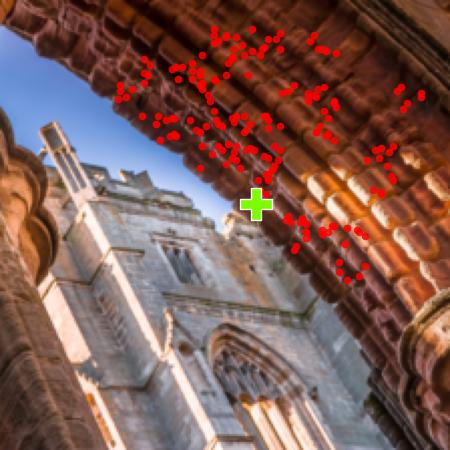}
         \label{fig:original}
     \end{subfigure} 
     \vspace{-0.3cm}
     \begin{subfigure}[b]{0.24\linewidth}
         \centering
         \includegraphics[width=\linewidth]{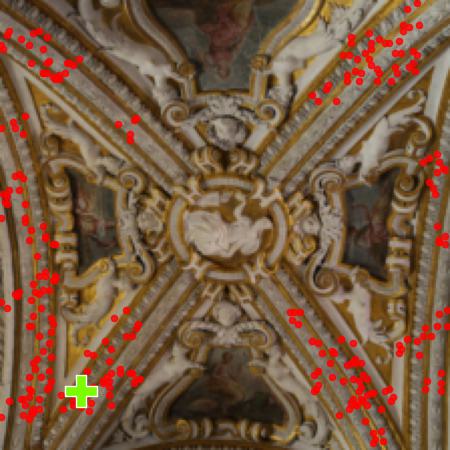}
         \label{fig:original}
     \end{subfigure}
     \begin{subfigure}[b]{0.24\linewidth}
         \centering
         \includegraphics[width=\linewidth]{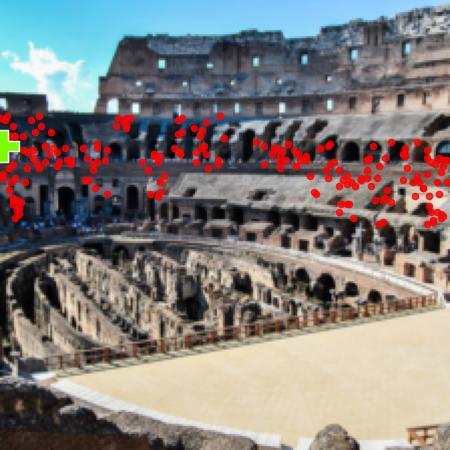}
         \label{fig:original}
     \end{subfigure}
     \\
      \begin{subfigure}[b]{0.24\linewidth}
         \centering
         \includegraphics[width=\linewidth]{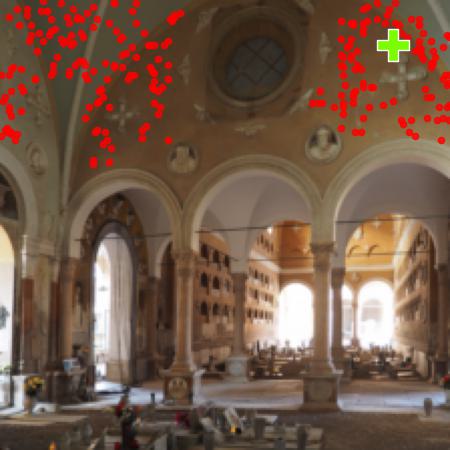}
         \label{fig:original}
     \end{subfigure}
     \begin{subfigure}[b]{0.24\linewidth}
         \centering
         \includegraphics[width=\linewidth]{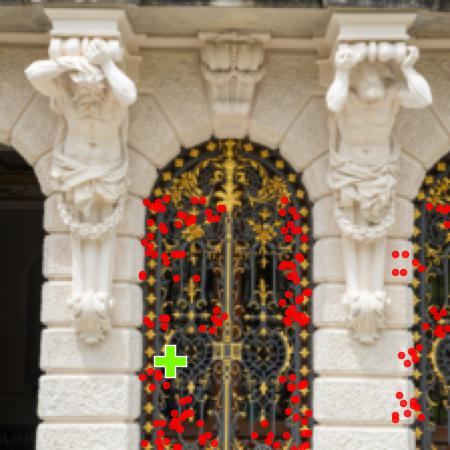}
         \label{fig:original}
     \end{subfigure} 
     \vspace{-0.3cm}
     \begin{subfigure}[b]{0.24\linewidth}
         \centering
         \includegraphics[width=\linewidth]{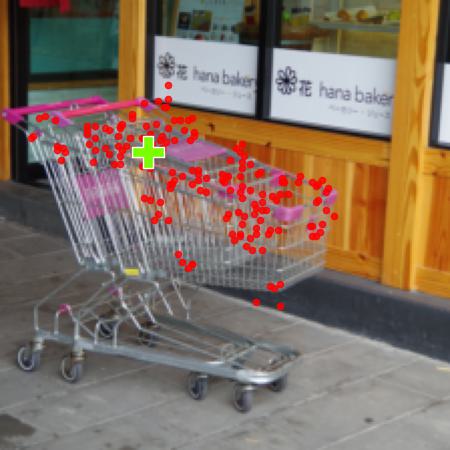}
         \label{fig:original}
     \end{subfigure}
     \begin{subfigure}[b]{0.24\linewidth}
         \centering
         \includegraphics[width=\linewidth]{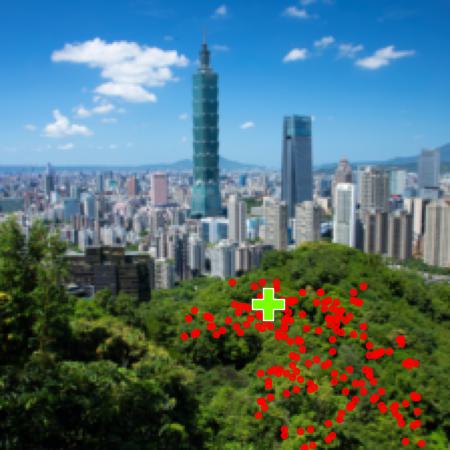}
         \label{fig:original}
     \end{subfigure}
     \\
      \begin{subfigure}[b]{0.24\linewidth}
         \centering
         \includegraphics[width=\linewidth]{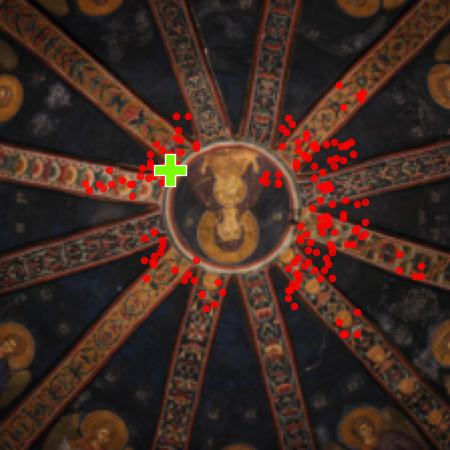}
         \label{fig:original}
     \end{subfigure}
     \begin{subfigure}[b]{0.24\linewidth}
         \centering
         \includegraphics[width=\linewidth]{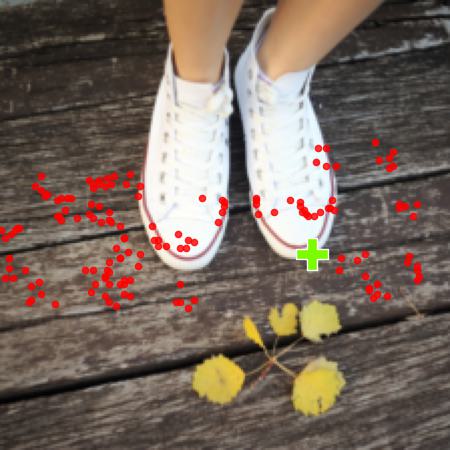}
         \label{fig:original}
     \end{subfigure} 
     \vspace{-0.3cm}
     \begin{subfigure}[b]{0.24\linewidth}
         \centering
         \includegraphics[width=\linewidth]{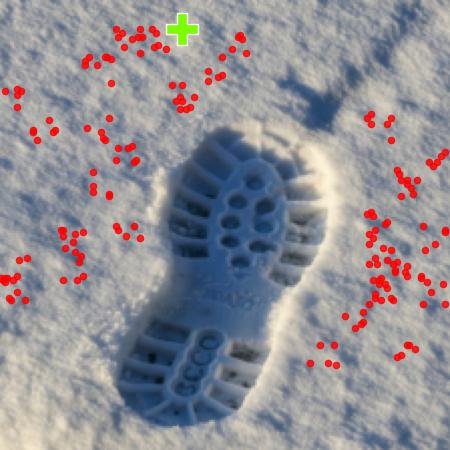}
         \label{fig:original}
     \end{subfigure}
     \begin{subfigure}[b]{0.24\linewidth}
         \centering
         \includegraphics[width=\linewidth]{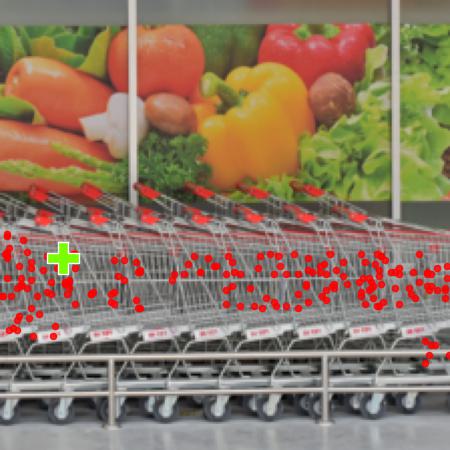}
         \label{fig:original}
     \end{subfigure}
     \\

\caption{\textbf{Visualization of attended region} (in \textcolor{red}{dots}) in the low-res features of a high-res pixel (in \textcolor{green}{cross}). The density of \textcolor{red}{dots} reflects the value of the attention map. LoftUp is able to use relevant information across the global feature map for upsampling features at each pixel.}
        \label{fig:attn-map-supp}

\end{figure*}

We further provide more visualization examples of upsampled features of various methods in~\cref{fig:more-vis-feats}, more prediction examples in semantic segmentation, depth estimation, and video object segmentation in~\cref{fig:more-vis-preds} and~\cref{fig:vis-preds-video}, more examples of different pseduo-GT in~\cref{fig:more-pseudo-gt}, and more examples of the attended regions of a high-resolution pixel in~\cref{fig:attn-map-supp}.


\end{document}